\documentclass[letterpaper]{article}
\usepackage{aaai18}
\usepackage{times}
\usepackage{helvet}
\usepackage{courier}
\usepackage{url}
\usepackage{graphicx}
\usepackage{graphicx} 
\usepackage{rotating}
\usepackage{epstopdf}  
\usepackage[footnotesize]{subfigure}
\usepackage{amsmath}  
\usepackage{amsfonts}
\newcommand{\e}[1]{{\mathbb E}\left[ #1 \right]}
\usepackage{afterpage}
\usepackage{enumitem}
\usepackage{balance}
\usepackage{algorithm}
\usepackage{algorithmic}
\usepackage{mathtools}
\mathtoolsset{showonlyrefs=true}

\newcounter{jzlNoteCounter}
\newcounter{mlNoteCounter}
\newcounter{thNoteCounter}
\newcounter{tgNoteCounter}

\newcommand{\argmax}{\operatornamewithlimits{argmax}}

\newcommand{\cD}{\mathcal{D}}

\newcommand{\abs}[1]{\left|#1\right|}

\author{Todd Hester\\Google DeepMind\\toddhester@google.com\\
\And Matej Vecerik\\Google DeepMind\\matejvecerik@google.com\\
\And Olivier Pietquin\\Google DeepMind\\pietquin@google.com\\
\And Marc Lanctot\\Google DeepMind\\lanctot@google.com\\
\And Tom Schaul\\Google DeepMind\\schaul@google.com\\
\AND Bilal Piot\\Google DeepMind\\piot@google.com\\
\And Dan Horgan\\Google DeepMind\\horgan@google.com\\
\And John Quan\\Google DeepMind\\johnquan@google.com\\
\And Andrew Sendonaris\\Google DeepMind\\sendos@yahoo.com\\
\And Ian Osband\\Google DeepMind\\iosband@google.com\\
\AND Gabriel Dulac-Arnold\\Google DeepMind\\gabe@squirrelsoup.net\\
\And John Agapiou\\Google DeepMind\\jagapiou@google.com\\
\And Joel Z. Leibo\\Google DeepMind\\jzl@google.com\\
\And Audrunas Gruslys\\Google DeepMind\\audrunas@google.com
}

\title{Deep Q-learning from Demonstrations}

\begin{document} 
\maketitle

\begin{abstract}
Deep reinforcement learning (RL) has achieved several high profile
successes in difficult decision-making problems.  However, these
algorithms typically require a huge amount of data before they reach
reasonable performance. In fact, their performance during learning can
be extremely poor. This may be acceptable for a simulator, but it
severely limits the applicability of deep RL to many real-world tasks,
where the agent must learn in the real environment. In this paper we
study a setting where the agent may access data from previous control
of the system. We present an algorithm, \textit{Deep Q-learning from
  Demonstrations} (DQfD), that leverages small sets of demonstration
data to massively accelerate the learning process even from relatively
small amounts of demonstration data and is able to automatically
assess the necessary ratio of demonstration data while learning thanks
to a prioritized replay mechanism. DQfD works by combining temporal
difference updates with supervised classification of the
demonstrator's actions. We show that DQfD has better initial
performance than Prioritized Dueling Double Deep Q-Networks (PDD DQN)
as it starts with better scores on the first million steps on 41 of 42
games and on average it takes PDD DQN 83 million steps to catch up to
DQfD's performance.  DQfD learns to out-perform the best demonstration
given in 14 of 42 games. In addition, DQfD leverages human
demonstrations to achieve state-of-the-art results for 11 games.
Finally, we show that DQfD performs better than three related
algorithms for incorporating demonstration data into DQN.
\end{abstract}

\section{Introduction}
\label{sec:introduction}
Over the past few years, there have been a number of successes in
learning policies for sequential decision-making problems and
control. Notable examples include deep model-free Q-learning for
general Atari game-playing~\cite{Mnih:2015}, end-to-end policy search
for control of robot motors~\cite{levine2015end}, model predictive
control with embeddings~\cite{Watter:2015}, and strategic policies
that combined with search led to defeating a top human expert at the
game of Go~\cite{Silver16Go}. An important part of the success of
these approaches has been to leverage the recent contributions to
scalability and performance of deep learning~\cite{LeCun:2015}. The
approach taken in~\cite{Mnih:2015} builds a data set of previous
experience using batch RL to train large convolutional neural networks
in a supervised fashion from this data. By sampling from this data set
rather than from current experience, the correlation
in values from state distribution bias is mitigated, leading
to good (in many cases, super-human) control policies.

It still remains difficult to apply these algorithms to real world
settings such as data centers, autonomous
vehicles~\cite{MLJ12-hester},
helicopters~\cite{Abbeel07anapplication}, or recommendation
systems~\cite{Shani:2005}. Typically these algorithms learn good
control policies only after many millions of steps of very poor
performance in simulation. This situation is acceptable when there is
a perfectly accurate simulator; however, many real world problems do
not come with such a simulator. Instead, in these situations, the
agent must learn in the real domain with real consequences for its
actions, which requires that the agent have good on-line performance
from the start of learning. While accurate simulators are difficult to
find, most of these problems have data of the system operating under a
previous controller (either human or machine) that performs reasonably
well. In this work, we make use of this demonstration data to
pre-train the agent so that it can perform well in the task from the
start of learning, and then continue improving from its own
self-generated data. Enabling learning in this framework opens up the
possibility of applying RL to many real world problems where
demonstration data is common but accurate simulators do not exist.

We propose a new deep reinforcement learning algorithm, \textit{Deep
  Q-learning from Demonstrations} (DQfD), which leverages even very
small amounts of demonstration data to massively accelerate
learning. DQfD initially pre-trains solely on the demonstration data
using a combination of temporal difference (TD) and supervised
losses. The supervised loss enables the algorithm to learn to imitate
the demonstrator while the TD loss enables it to learn a
self-consistent value function from which it can continue learning
with RL. After pre-training, the agent starts interacting with the
domain with its learned policy. The agent updates its network with a
mix of demonstration and self-generated data. In practice, choosing
the ratio between demonstration and self-generated data while learning
is critical to improve the performance of the algorithm. One of our
contributions is to use a prioritized replay
mechanism~\cite{SchaulQAS16} to automatically control this ratio. DQfD
out-performs pure reinforcement learning using Prioritized Dueling
Double DQN (PDD DQN)~\cite{SchaulQAS16,vanHasselt:2016,Wang16dueling}
in 41 of 42 games on the first million steps, and on average it takes
83 million steps for PDD DQN to catch up to DQfD. In addition, DQfD
out-performs pure imitation learning in mean score on 39 of 42 games
and out-performs the best demonstration given in 14 of 42 games. DQfD
leverages the human demonstrations to learn state-of-the-art policies
on 11 of 42 games.  Finally, we show that DQfD performs better than
three related algorithms for incorporating demonstration data into
DQN.

\section{Background}
\label{sec:background}
We adopt the standard Markov Decision Process (MDP) formalism for this
work~\cite{SuttonBarto:1998}. An MDP is defined by a tuple $\left<S,
A, R, T, \gamma\right>$, which consists of a set of states $S$, a set
of actions $A$, a reward function $R(s,a)$, a transition function
$T(s, a, s') = P(s'|s,a)$, and a discount factor $\gamma$. In each
state $s \in S$, the agent takes an action $a \in A$. Upon taking this
action, the agent receives a reward $R(s,a)$ and reaches a new state
$s'$, determined from the probability distribution $P(s'|s,a)$. A
policy $\pi$ specifies for each state which action the agent will
take. The goal of the agent is to find the policy $\pi$ mapping states
to actions that maximizes the expected discounted total reward over
the agent's lifetime. The value $Q^\pi(s,a)$ of a given state-action
pair $(s,a)$ is an estimate of the expected future reward that can be
obtained from $(s,a)$ when following policy $\pi$. The optimal value
function $Q^*(s,a)$ provides maximal values in all states and is
determined by solving the Bellman equation:
\begin{equation}
Q^*(s,a) = \e{R(s,a) + \gamma \sum_{s'} P(s'|s,a) \max_{a'} Q^*(s',a')}.
\end{equation}
The optimal policy $\pi$ is then $\pi(s) = \argmax_{a\in A}Q^*(s,a).$
DQN~\cite{Mnih:2015} approximates the value function $Q(s,a)$ with a
deep neural network that outputs a set of action values
$Q(s,\cdot;\theta)$ for a given state input $s$, where $\theta$ are
the parameters of the network. There are two key components of DQN
that make this work. First, it uses a separate target network that is
copied every $\tau$ steps from the regular network so that the target
Q-values are more stable. Second, the agent adds all of its
experiences to a replay buffer $\cD^{replay}$, which is then sampled
uniformly to perform updates on the network.

The double Q-learning update~\cite{vanHasselt:2016} uses the current
network to calculate the argmax over next state values and the target
network for the value of that action. The double DQN loss is
    $J_{DQ}(Q) = \left( R(s,a) + \gamma Q(s_{t+1}, a^{\max}_{t+1}; \theta') - Q(s, a; \theta) \right)^2$,   
where $\theta'$ are the parameters of the target network, and
$a^{\max}_{t+1} = \argmax_{a}Q(s_{t+1}, a; \theta)$. Separating the
value functions used for these two variables reduces the upward bias
that is created with regular Q-learning updates.

Prioritized experience replay~\cite{SchaulQAS16} modifies the DQN
agent to sample more important transitions from its replay buffer more
frequently. The probability of sampling a particular transition $i$
is proportional to its priority, $P(i) = \frac{p_i^\alpha}{\sum_k p_k^\alpha}$,
where the priority $p_i = \abs{\delta_i} + \epsilon$, and
$\delta_i$ is the last TD error calculated for this transition
and $\epsilon$ is a small positive constant to ensure all transitions
are sampled with some probability.  To account for the change in the
distribution, updates to the network are weighted with importance
sampling weights, $w_i = (\frac{1}{N} \cdot \frac{1}{P(i)})^\beta$,
where $N$ is the size of the replay buffer and
$\beta$ controls the amount of importance sampling with no
importance sampling when $\beta = 0$ and full importance sampling when
$\beta = 1$. $\beta$ is annealed linearly from $\beta_0$ to $1$.

\section{Related Work}
\label{sec:relatedwork}

{\it Imitation learning} is primarily concerned with matching the
performance of the demonstrator.  One popular algorithm,
DAGGER~\cite{Ross11Dagger}, iteratively produces new policies based on
polling the expert policy outside its original state space, showing
that this leads to no-regret over validation data in the online
learning sense. DAGGER requires the expert to be available during
training to provide additional feedback to the agent. In addition,
it does not combine imitation with reinforcement learning, meaning
it can never learn to improve beyond the expert as DQfD can. 

Deeply AggreVaTeD~\cite{SunVGBB17} extends DAGGER to work with deep
neural networks and continuous action spaces. Not only does it require
an always available expert like DAGGER does, the expert must provide a
value function in addition to actions. Similar to DAGGER, Deeply
AggreVaTeD only does imitation learning and cannot learn to improve
upon the expert.

Another popular
paradigm is to setup a zero-sum game where the learner chooses a
policy and the adversary chooses a reward
function~\cite{Syed07Game,Syed08Apprenticeship,Ho16Generative}.
Demonstrations have also been used for inverse optimal control in
high-dimensional, continuous robotic control
problems~\cite{Finn16Guided}.  However, these approaches only do
imitation learning and do not allow for learning from task rewards.


Recently, demonstration data has been shown to help in difficult
exploration problems in RL~\cite{Subramanian16}.  There has also been
recent interest in this combined imitation and RL problem. For
example, the HAT algorithm transfers knowledge directly from human
policies~\cite{Taylor11}. Follow-ups to this work showed how expert
advice or demonstrations can be used to shape rewards in the RL
problem~\cite{Brys15,SuayBTC16}.  A different approach is to shape the
policy that is used to sample experience~\cite{Cederborg15}, or to use
policy iteration from demonstrations~\cite{Kim13,Chemali15}.

Our algorithm works in a scenario where rewards are given by the
environment used by the demonstrator. This framework was appropriately
called Reinforcement Learning with Expert Demonstrations (RLED)
in~\cite{Piot14Boosted} and is also evaluated
in~\cite{Kim13,Chemali15}. Our setup is similar
to~\cite{Piot14Boosted} in that we combine TD and classification
losses in a batch algorithm in a model-free setting; ours differs in
that our agent is pre-trained on the demonstration data initially and
the batch of self-generated data grows over time and is used as
experience replay to train deep Q-networks.  In addition, a
prioritized replay mechanism is used to balance the amount of
demonstration data in each mini-batch.
\cite{RCAL2014}
present interesting results showing that adding a TD loss to the
supervised classification loss improves imitation learning even when
there are no rewards.


Another work that is similarly motivated to ours is
\cite{Schaal96}. This work is focused on real world learning on
robots, and thus is also concerned with on-line performance. Similar
to our work, they pre-train the agent with demonstration data before
letting it interact with the task. However, they do not use supervised
learning to pre-train their algorithm, and are only able to find one
case where pre-training helps learning on Cart-Pole.

In one-shot imitation learning~\cite{Duan17}, the agent is provided
with an entire demonstration as input in addition to the current
state. The demonstration specifies the goal state that is wanted, but
from different initial conditions. The agent is trained with target
actions from more demonstrations. This setup also uses demonstrations,
but requires a distribution of tasks with different initial conditions
and goal states, and the agent can never learn to improve upon the
demonstrations.

AlphaGo~\cite{Silver16Go} takes a similar approach to our work in
pre-training from demonstration data before interacting with the real
task. AlphaGo first trains a policy network from a dataset of 30
million expert actions, using supervised learning to predict the
actions taken by experts. It then uses this as a starting point to
apply policy gradient updates during self-play, combined with planning
rollouts. Here, we do not have a model available for planning, so we
focus on the model-free Q-learning case.

Human Experience Replay (HER)~\cite{Hosu2016Playing} is an algorithm
in which the agent samples from a replay buffer that is mixed between
agent and demonstration data, similar to our approach. Gains were only
slightly better than a random agent, and were surpassed by their
alternative approach, Human Checkpoint Replay, which requires the
ability to set the state of the environment. While their algorithm is
similar in that it samples from both datasets, it does not pre-train the
agent or use a supervised loss. Our results show higher scores over a
larger variety of games, without requiring full access to the
environment.
Replay Buffer Spiking (RBS)~\cite{LiptonGLLAD16} is another similar
approach where the DQN agent's replay buffer is initialized with
demonstration data, but they do not pre-train the agent for good
initial performance or keep the demonstration data permanently.

The work that most closely relates to ours is a workshop paper
presenting Accelerated DQN with Expert
Trajectories (ADET)~\cite{Lakshminarayanan:2016}. They are also combining TD
and classification losses in a deep Q-learning setup. They use a
trained DQN agent to generate their demonstration data, which on most
games is better than human data. It also guarantees that the policy
used by the demonstrator can be represented by the apprenticeship
agent as they are both using the same state input and network
architecture. They use a cross-entropy classification loss rather than
the large margin loss DQfD uses and they do not pre-train the agent to
perform well from its first interactions with the environment.

\section{Deep Q-Learning from Demonstrations}
\label{sec:algorithm}

In many real-world settings of reinforcement learning, we have access
to data of the system being operated by its previous controller, but
we do not have access to an accurate simulator of the
system. Therefore, we want the agent to learn as much as possible from
the demonstration data before running on the real system. The goal of
the pre-training phase is to learn to imitate the demonstrator with a
value function that satisfies the Bellman equation so that it can be
updated with TD updates once the agent starts interacting with the
environment. During this pre-training phase, the agent samples
mini-batches from the demonstration data and updates the network by
applying four losses: the 1-step double Q-learning loss, an n-step
double Q-learning loss, a supervised large margin classification loss,
and an L2 regularization loss on the network weights and biases. The
supervised loss is used for classification of the demonstrator's
actions, while the Q-learning loss ensures that the network satisfies
the Bellman equation and can be used as a starting point for TD
learning.

The supervised loss is critical for the pre-training to have any
effect. Since the demonstration data is necessarily covering a narrow
part of the state space and not taking all possible actions, many
state-actions have never been taken and have no data to ground them to
realistic values. If we were to pre-train the network with only
Q-learning updates towards the max value of the next state, the
network would update towards the highest of these ungrounded variables
and the network would propagate these values throughout the Q
function.
We add a large margin classification loss~\cite{Piot14Boosted}:
\begin{equation}
  J_E(Q) = \max_{a \in A}[Q(s, a) + l(a_E, a)] - Q(s, a_E)
  \label{eq:sup}
  \end{equation}
where $a_E$ is the action the expert demonstrator took in state $s$
and $l(a_E, a)$ is a margin function that is $0$ when $a = a_E$ and
positive otherwise. This loss forces the values of the other actions
to be at least a margin lower than the value of the demonstrator's
action. Adding this loss grounds the values of the unseen actions to
reasonable values, and makes the greedy policy induced by the value
function imitate the demonstrator. If the algorithm pre-trained with
only this supervised loss, there would be nothing constraining the
values between consecutive states and the Q-network would not satisfy
the Bellman equation, which is required to improve the policy on-line
with TD learning.

Adding n-step returns (with $n=10$) helps propagate the values of the expert's
trajectory to all the earlier states, leading to better
pre-training. The n-step return is:
\begin{equation}
  r_t + {\gamma}r_{t+1} + ... + \gamma^{n-1}r_{t+n-1} + max_{a}\gamma^nQ(s_{t+n}, a),
\end{equation}
which we calculate using the forward view, similar to
A3C~\cite{mnih2016asynchronous}.

We also add an L2 regularization loss applied to the weights and biases
of the
network to help prevent it from over-fitting on the relatively small
demonstration dataset. The overall loss used to update the network is a
combination of all four losses:
\begin{equation}
  J(Q) = J_{DQ}(Q) + {\lambda_1}J_n(Q) + {\lambda_2}J_E(Q) + {\lambda_3}J_{L2}(Q).
  \label{eq:losses}
\end{equation}
The $\lambda$ parameters control the weighting between the losses.
We examine removing some of these losses in Section~\ref{sec:results}.

Once the pre-training phase is complete, the agent starts acting on
the system, collecting self-generated data, and adding it to its
replay buffer $\cD^{replay}$. Data is added to the replay buffer until
it is full, and then the agent starts over-writing old data in that
buffer. However, the agent never over-writes the demonstration data.
For proportional prioritized sampling, different small positive
constants, $\epsilon_a$ and $\epsilon_d$, are added to the priorities
of the agent and demonstration transitions to control the relative
sampling of demonstration versus agent data.
All the losses are applied to the demonstration data in both phases, while the
supervised loss is not applied to self-generated data ($\lambda_2 = 0$).

Overall, Deep Q-learning from Demonstration (DQfD) differs from
PDD DQN in six key ways:
\begin{itemize}[noitemsep,nolistsep]
\item Demonstration data: DQfD is given a set of demonstration data,
  which it retains in its replay buffer permanently.
\item Pre-training: DQfD initially trains solely on the demonstration data
  before starting any interaction with the environment.
\item Supervised losses: In addition to TD losses, a large margin
  supervised loss is applied that pushes the value of the demonstrator's
  actions above the other action values~\cite{Piot14Boosted}.
\item L2 Regularization losses: The algorithm also adds L2
  regularization losses on the network weights to prevent over-fitting
  on the demonstration data.
\item N-step TD losses: The agent updates its Q-network
  with targets from a mix of 1-step and n-step returns.
\item Demonstration priority bonus: The priorities of demonstration
  transitions are given a bonus of $\epsilon_d$, to boost the frequency
  that they are sampled.
\end{itemize}

Pseudo-code is sketched in Algorithm~\ref{alg:dqfd}. The behavior
policy $\pi^{\epsilon Q_\theta}$ is $\epsilon$-greedy with respect to $Q_\theta$.

\begin{algorithm}[h!]
  \caption{Deep Q-learning from Demonstrations. \label{alg:dqfd}}
  \begin{algorithmic}[1]
\STATE Inputs:
$\cD^{replay}$: initialized with demonstration data set,
$\theta$: weights for initial behavior network (random),
$\theta'$: weights for target network (random),
$\tau$: frequency at which to update target net,
$k$: number of pre-training gradient updates
\FOR{steps $t \in \{ 1, 2, \ldots k\}$} 
  \STATE Sample a mini-batch of $n$ transitions from $\cD^{replay}$ with prioritization
  \STATE Calculate loss $J(Q)$ using target network
  \STATE Perform a gradient descent step to update $\theta$
  \STATE {\bf if} $t~\mathbf{mod}~\tau = 0$ {\bf then} $\theta' \leftarrow \theta$ {\bf end if}
\ENDFOR
\FOR{steps $t \in \{ 1, 2, \ldots \}$} 
  \STATE Sample action from behavior policy $a \sim \pi^{\epsilon Q_\theta}$
  \STATE Play action $a$ and observe $(s', r)$.
  \STATE Store $(s,a,r,s')$ into $\cD^{replay}$, overwriting oldest self-generated transition if over capacity
  \STATE Sample a mini-batch of $n$ transitions from $\cD^{replay}$ with prioritization
  \STATE Calculate loss $J(Q)$ using target network
  \STATE Perform a gradient descent step to update $\theta$
  \STATE {\bf if} $t~\mathbf{mod}~\tau = 0$ {\bf then} $\theta' \leftarrow \theta$ {\bf end if}
  \STATE $s \leftarrow s'$
\ENDFOR
\end{algorithmic}
\end{algorithm}

\section{Experimental Setup}
\label{sec:experiments}

We evaluated DQfD on the Arcade Learning Environment
(ALE)~\cite{Bellemare:2013}. ALE is a set of Atari games that are a
standard benchmark for DQN and contains many games on which humans
still perform better than the best learning agents. The agent plays
the Atari games from a down-sampled 84x84 image of the game screen
that has been converted to greyscale, and the agent stacks four of
these frames together as its state. The agent must output one of 18
possible actions for each game. The agent applies a discount factor of
0.99 and all of its actions are repeated for four Atari frames. Each
episode is initialized with up to 30 no-op actions to
provide random starting positions. The scores reported are the scores
in the Atari game, regardless of how the agent is representing
reward internally.

For all of our experiments, we evaluated three different algorithms,
each averaged across four trials:
\begin{itemize}[noitemsep,nolistsep]
\item Full DQfD algorithm with human demonstrations
\item PDD DQN learning without any demonstration data
\item Supervised imitation from demonstration data without any environment
  interaction
\end{itemize}

We performed informal parameter tuning for all the algorithms on six
Atari games and then used the same parameters for the entire set of
games. The parameters used for the algorithms are shown in the
appendix.
Our coarse search over prioritization and n-step return parameters
led to the same best parameters for DQfD and PDD DQN.
PDD DQN differs
from DQfD because it does not have demonstration data, pre-training,
supervised losses, or regularization losses. We included n-step
returns in PDD DQN to provide a better baseline for comparison between
DQfD and PDD DQN. All three algorithms use the dueling
state-advantage convolutional
network architecture~\cite{Wang16dueling}.
 
For the supervised imitation comparison, we performed supervised
classification of the demonstrator's actions using a cross-entropy
loss, with the same network architecture and L2 regularization used by
DQfD. The imitation algorithm did not use any TD loss.
Imitation learning only learns from the pre-training and not from any additional interactions.

We ran experiments on a randomly selected subset of 42 Atari games.
We had a human player play each game between three and twelve times.
Each episode was played either until the game terminated or for 20
minutes. During game play, we logged the agent's state, actions,
rewards, and terminations. The human demonstrations range from 5,574
to 75,472 transitions per game. DQfD learns from a very small dataset
compared to other similar work, as AlphaGo~\cite{Silver16Go} learns
from 30 million human transitions, and DQN~\cite{Mnih:2015} learns
from over 200 million frames. DQfD's smaller demonstration dataset
makes it more difficult to learn a good representation without
over-fitting. The demonstration scores for each game are shown in a
table in the Appendix. Our human demonstrator is much better than PDD
DQN on some games (e.g. Private Eye, Pitfall), but much worse than PDD
DQN on many games (e.g. Breakout, Pong).


We found that in many of the games where the human player is better
than DQN, it was due to DQN being trained with all rewards clipped to
1. For example, in Private Eye, DQN has no reason
to select actions that reward 25,000 versus actions that reward
10. To make the reward function used by the human demonstrator and
the agent more consistent, we used unclipped rewards and converted the
rewards using a log scale: $r_{agent} = sign(r) \cdot log(1 + \abs{r})$.
This transformation keeps the rewards over a reasonable scale for the
neural network to learn, while conveying important information about
the relative scale of individual rewards.
These adapted rewards are used internally by the all the algorithms in
our experiments. Results are still reported using actual game scores
as is typically done in the Atari literature~\cite{Mnih:2015}.

\section{Results}
\label{sec:results}

\begin{figure*}[tb]
  \centering
  \includegraphics[width=0.52\columnwidth]{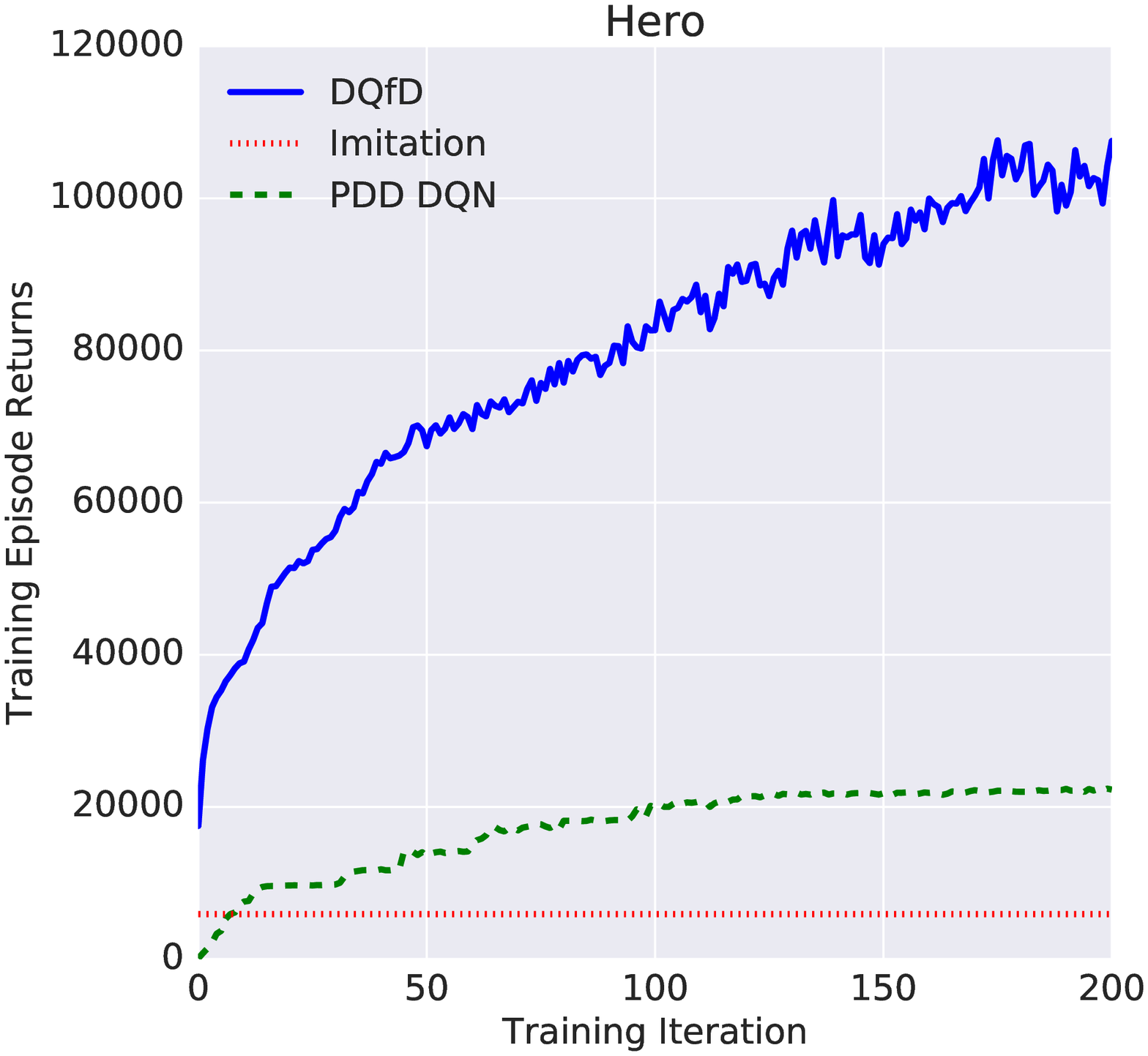}
  \includegraphics[width=0.52\columnwidth]{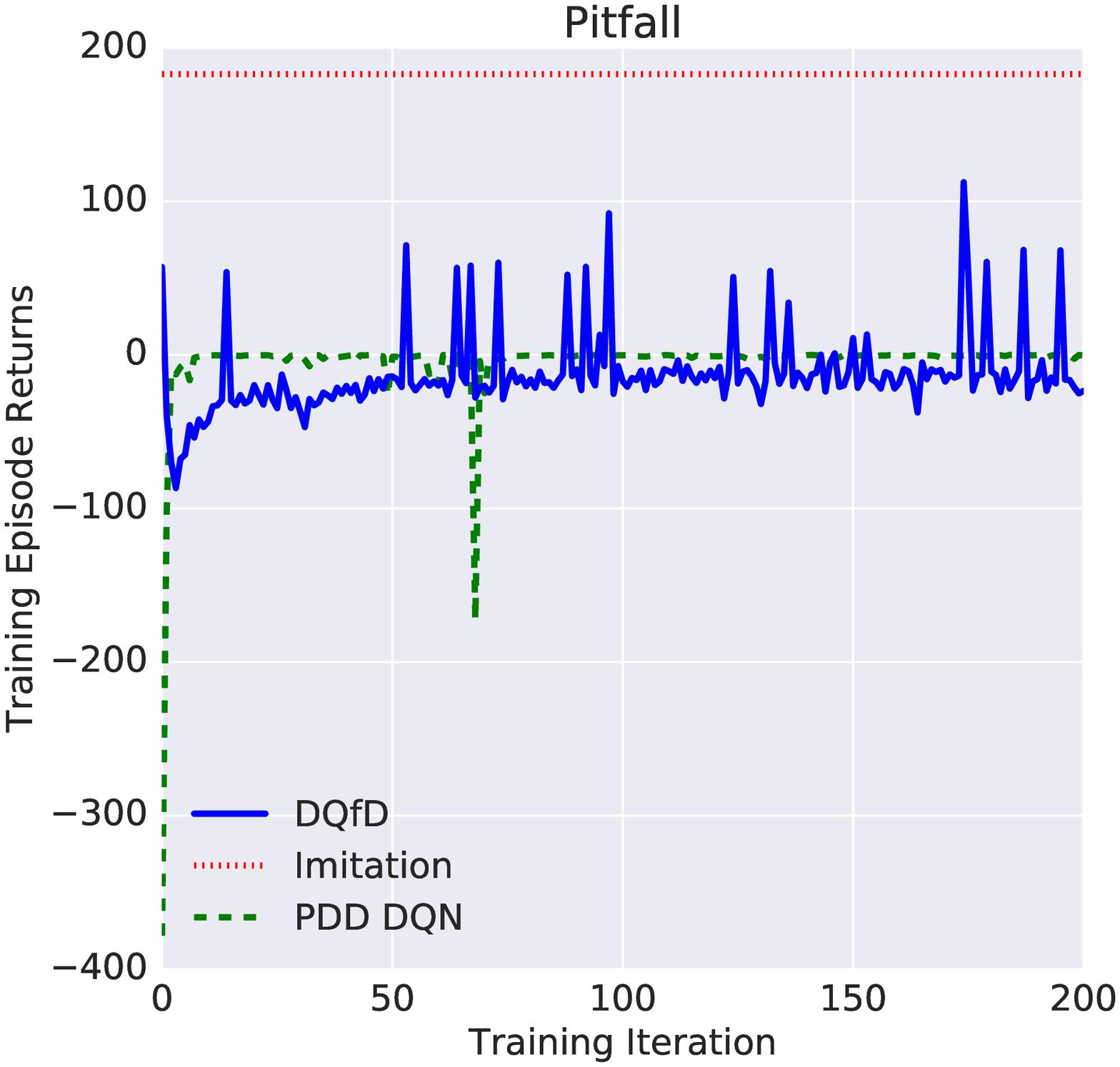}
  \includegraphics[width=0.52\columnwidth]{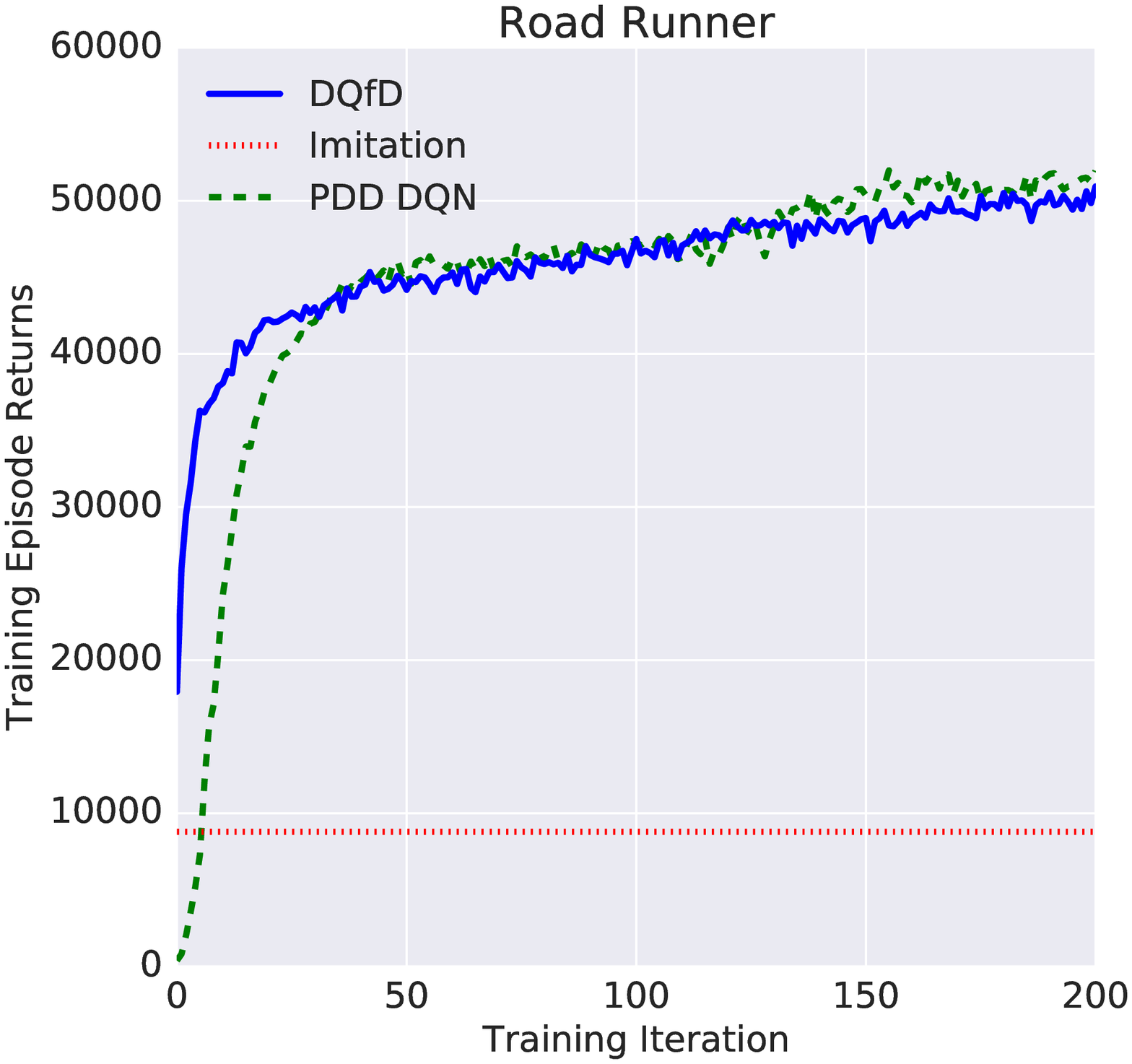}
  \includegraphics[width=0.52\columnwidth]{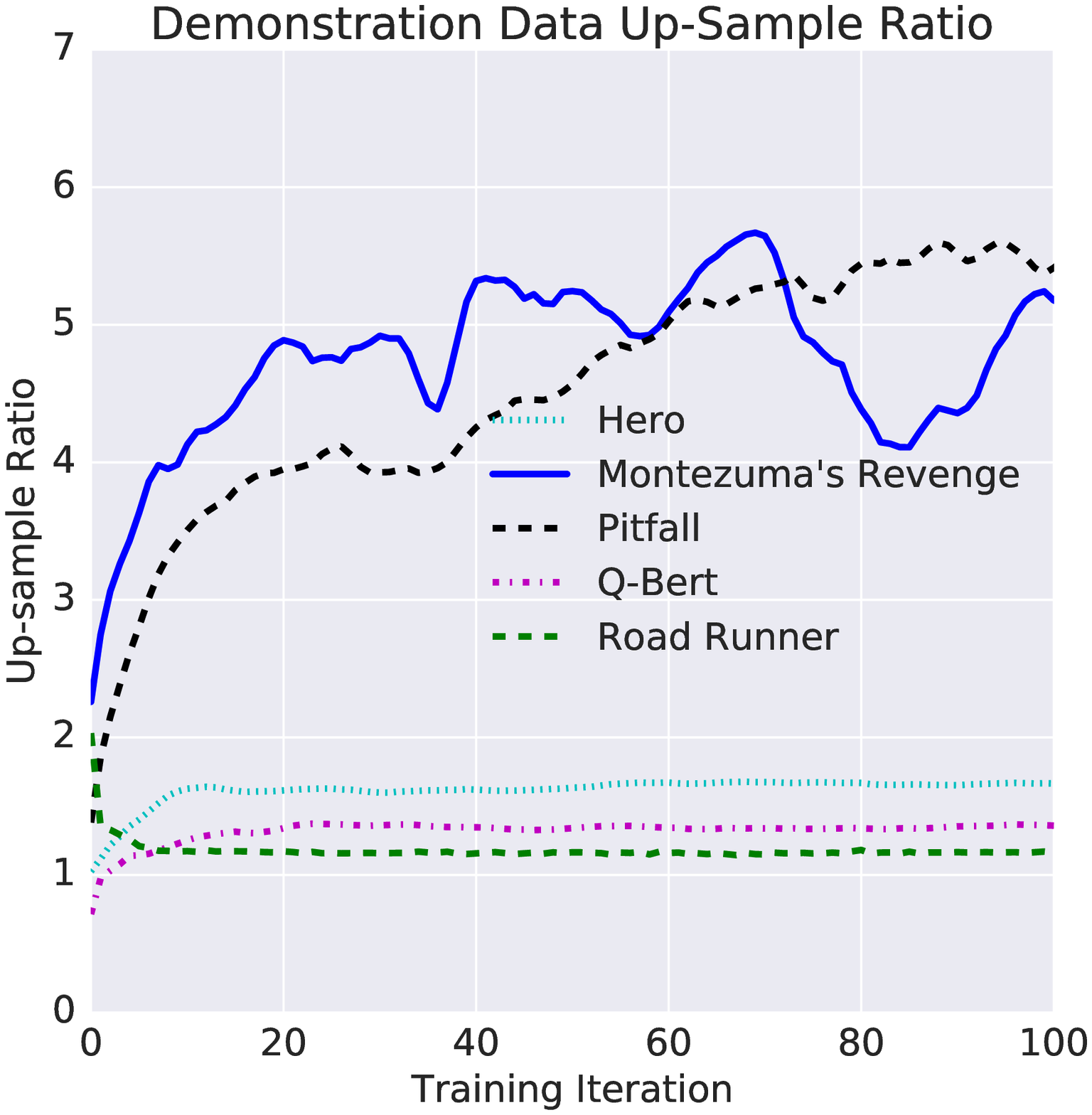}
  \vspace{-0.7cm}
  \caption{On-line scores of the algorithms on the games of Hero,
    Pitfall, and Road Runner. On Hero and Pitfall, DQfD leverages the
    human demonstrations to achieve a higher score than any previously
    published result. The last plot shows how much more frequently the
    demonstration data was sampled than if data were sampled uniformly,
    for five different games.}
  \label{fig:hero}
\vspace{-0.2cm}
\end{figure*}

First, we show learning curves in Figure~\ref{fig:hero} for three
games: Hero, Pitfall, and Road Runner. On Hero and Pitfall, the human
demonstrations enable DQfD to achieve a score higher than any
previously published result. Videos for both games are available
at \url{https://www.youtube.com/watch?v=JR6wmLaYuu4}.
On Hero, DQfD achieves a higher score than any of the human demonstrations
as well as any previously published result.
Pitfall may be the most difficult Atari game,
as it has very sparse positive rewards and dense negative rewards. No
previous approach achieved any positive rewards on this game, while
DQfD's best score on this game averaged over a 3 million step period
is $394.0$.

On Road Runner, agents typically learn super-human policies with a
score exploit that differs greatly from human play. Our
demonstrations are only human and have a maximum score of 20,200.
Road Runner is the game with the smallest set of human
demonstrations (only 5,574 transitions). Despite these factors, DQfD
still achieves a higher score than PDD DQN for the first 36 million
steps and matches PDD DQN's performance after that.

The right subplot in Figure~\ref{fig:hero} shows the ratio of how
often the demonstration data was sampled versus how much it would be
sampled with uniform sampling. For the most difficult games like
Pitfall and Montezuma's Revenge, the demonstration data is sampled
more frequently over time. For most other games, the ratio converges
to a near constant level, which differs for each game.

In real world tasks, the agent must perform well from its very first
action and must learn quickly. DQfD performed better than PDD DQN
on the first million steps on 41 of 42 games. In addition, on
31 games, DQfD starts out with higher performance than pure imitation
learning, as the addition of the TD loss helps the agent generalize
the demonstration data better.
On average, PDD DQN does not surpass the performance of DQfD until 83
million steps into the task and never surpasses it in mean scores.
 

In addition to boosting initial performance, DQfD is able to leverage
the human demonstrations to learn better policies on the most
difficult Atari games. We compared DQfD's scores over 200 million
steps with that of other deep reinforcement learning approaches: DQN,
Double DQN, Prioritized DQN, Dueling DQN, PopArt, DQN+CTS, and
DQN+PixelCNN~\cite{Mnih:2015,vanHasselt:2016,SchaulQAS16,Wang16dueling,popart,Ostrovski17}.
We took the best 3 million step window averaged over 4 seeds for the
DQfD scores.  DQfD achieves better scores than these algorithms on 11
of 42 games, shown in Table~\ref{tab:sota}. Note that we do not
compare with A3C~\cite{mnih2016asynchronous} or Reactor~\cite{reactor}
as the only published
results are for human starts, and we do not compare with
UNREAL~\cite{Jaderberg16} as they select the best hyper-parameters per
game. Despite this fact, DQfD still out-performs the best UNREAL
results on 10 games. DQN with count-based
exploration~\cite{Ostrovski17} is designed for and achieves the best
results on the most difficult exploration games. On the six sparse
reward, hard exploration games both algorithms were run on, DQfD
learns better policies on four of six games.

\begin{table}[tbp]
\scriptsize \tabcolsep 3pt \centering
\begin{tabular}{||l||l|l|l|}
  \hline \hline
  Game & DQfD & Prev. Best & Algorithm \\
  \hline
Alien & \textbf{4745.9} & 4461.4 & Dueling DQN~\cite{Wang16dueling}\\
Asteroids & \textbf{3796.4} & 2869.3 & PopArt~\cite{popart}\\
Atlantis & \textbf{920213.9} & 395762.0 & Prior. Dueling DQN~\cite{Wang16dueling}\\
Battle Zone & \textbf{41971.7} & 37150.0 & Dueling DQN~\cite{Wang16dueling}\\
Gravitar & \textbf{1693.2} & 859.1 & DQN+PixelCNN~\cite{Ostrovski17}\\
Hero & \textbf{105929.4} & 23037.7 & Prioritized DQN~\cite{SchaulQAS16}\\
Montezuma Revenge & \textbf{4739.6} & 3705.5 & DQN+CTS~\cite{Ostrovski17}\\
Pitfall & \textbf{50.8} & 0.0 & Prior. Dueling DQN~\cite{Wang16dueling}\\
Private Eye & \textbf{40908.2} & 15806.5 & DQN+PixelCNN~\cite{Ostrovski17}\\
Q-Bert & \textbf{21792.7} & 19220.3 & Dueling DQN~\cite{Wang16dueling}\\
Up N Down & \textbf{82555.0} & 44939.6 & Dueling DQN~\cite{Wang16dueling}\\

  \hline
  \hline
\end{tabular}
\caption{Scores for the 11 games where DQfD achieves higher scores
  than any previously published deep RL result using random no-op starts.
  Previous results take
  the best agent at its best iteration and evaluate it for 100
  episodes. DQfD scores are the best 3 million step window averaged
  over four seeds, which is 508 episodes on average.}
\label{tab:sota}
\vspace{-0.4cm}
\end{table}

DQfD out-performs the worst demonstration episode it was given
on in 29 of 42 games and it learns to play better than the best
demonstration episode in 14 of the games: Amidar, Atlantis, Boxing,
Breakout, Crazy Climber, Defender, Enduro, Fishing Derby, Hero, James Bond,
Kung Fu Master, Pong, Road Runner, and Up N Down. In comparison, pure
imitation learning is worse than the demonstrator's performance in
every game.

\begin{figure*}[tb]
  \centering
  \includegraphics[width=0.52\columnwidth]{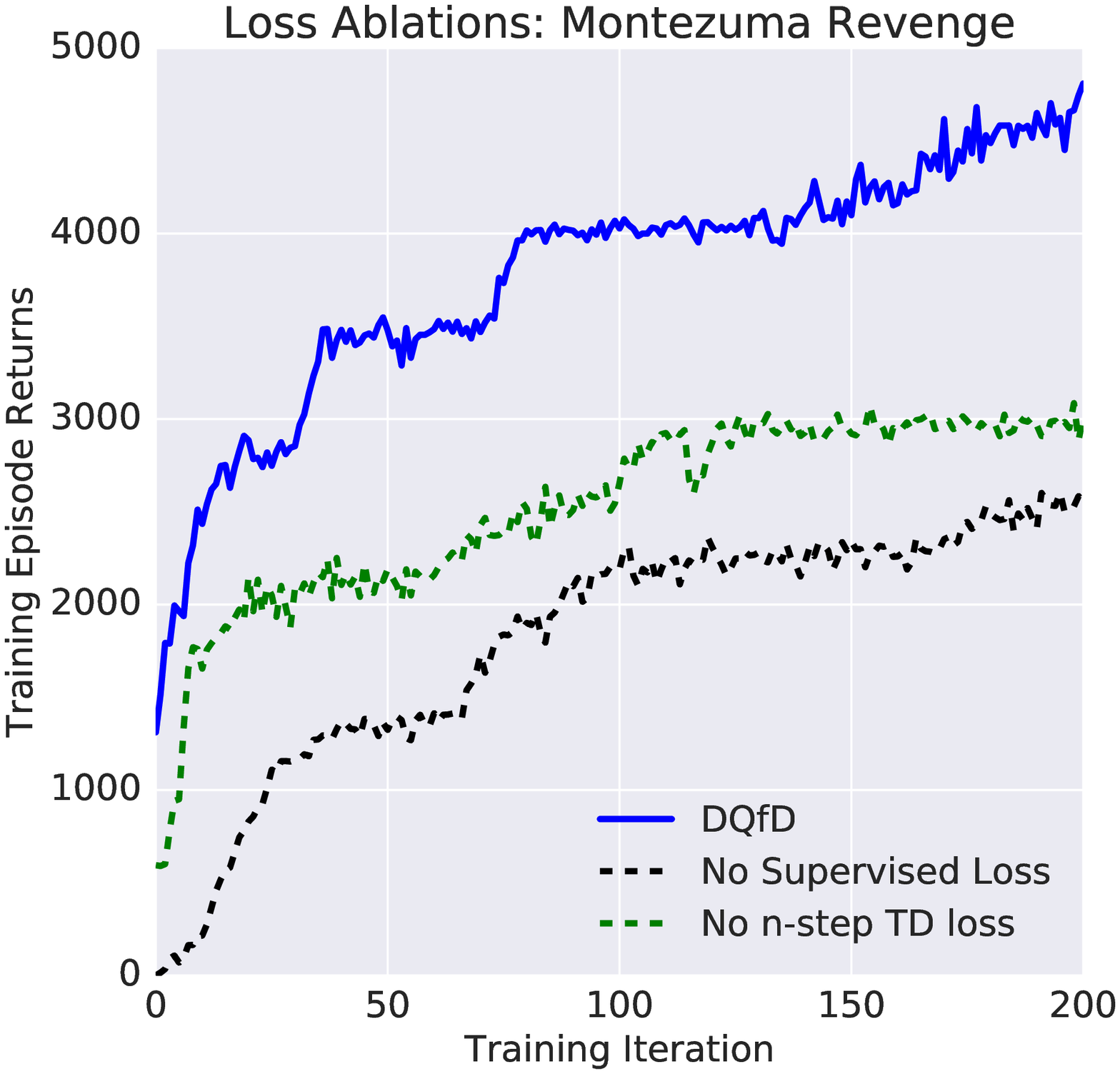}
  \includegraphics[width=0.52\columnwidth]{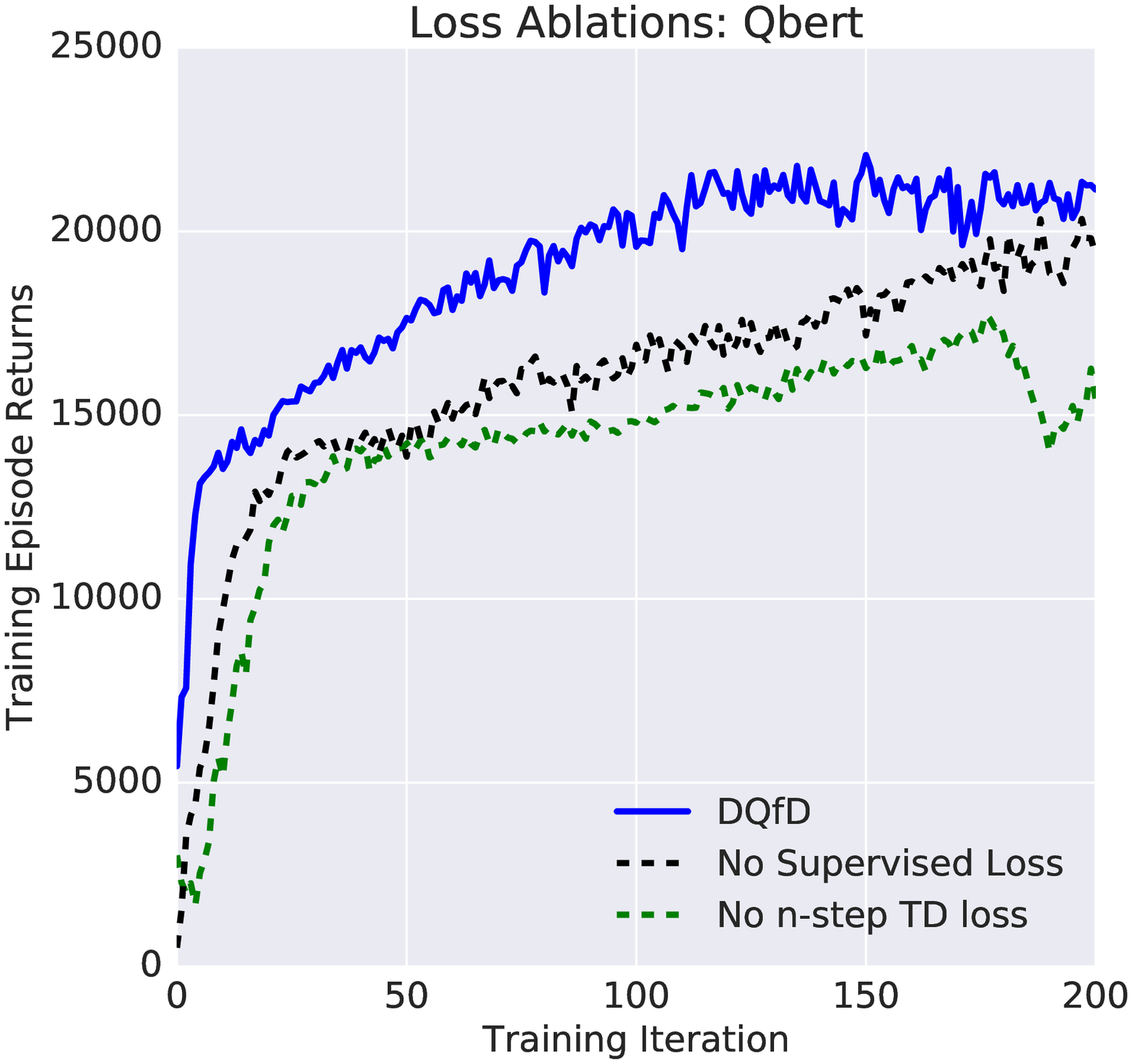}
  \includegraphics[width=0.52\columnwidth]{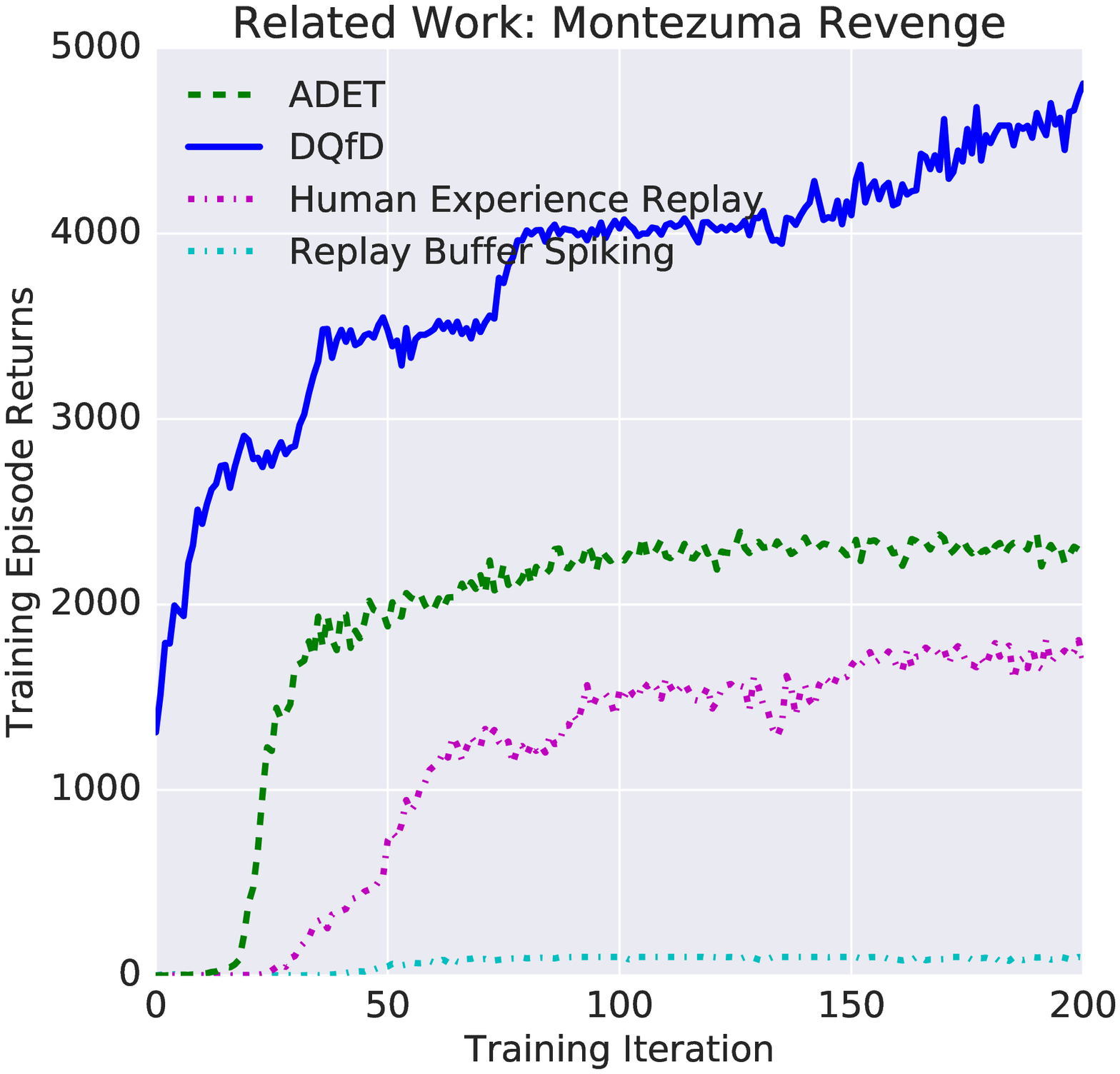}
  \includegraphics[width=0.52\columnwidth]{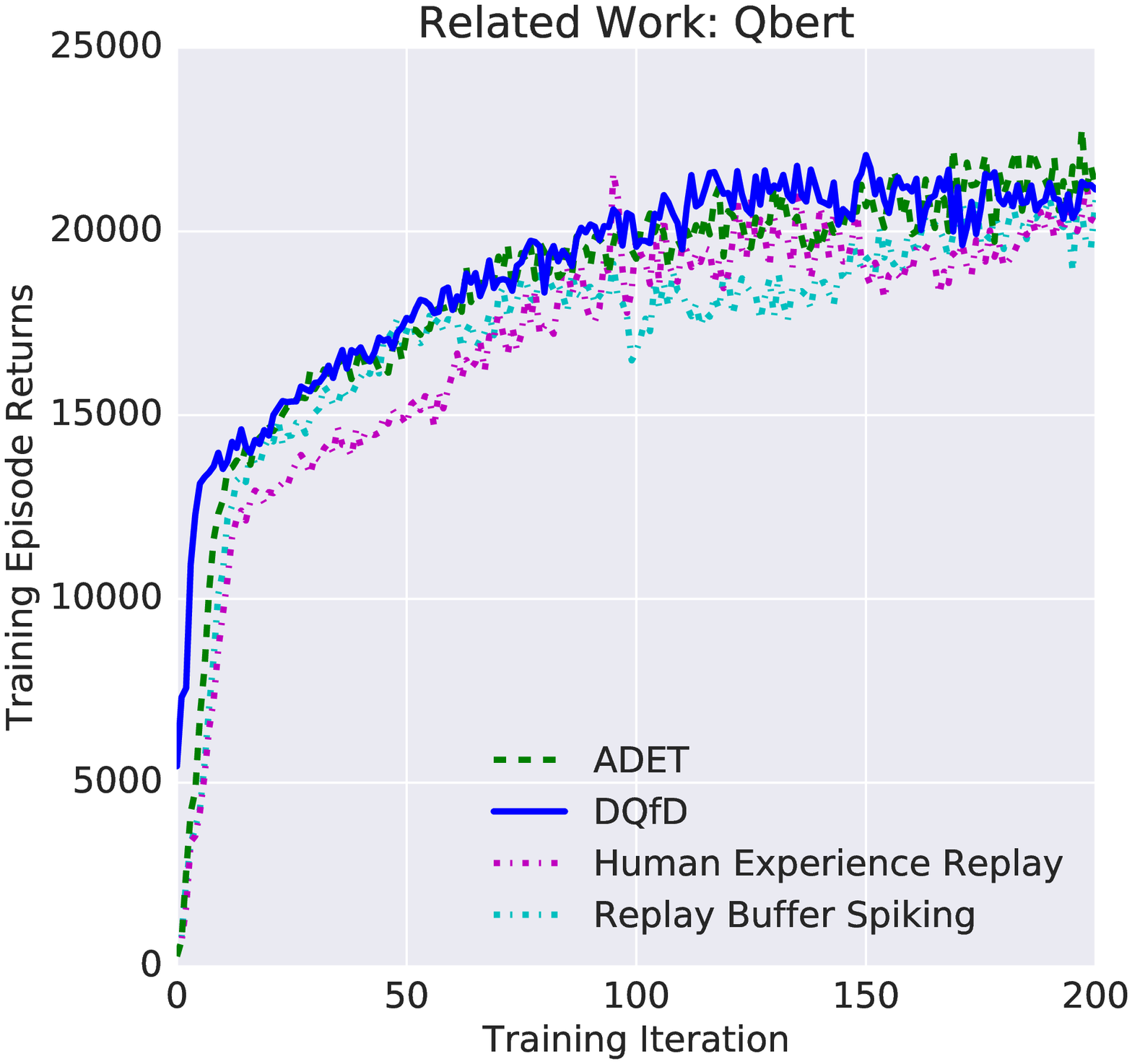}
  \vspace{-0.7cm}
  \caption{The left plots show on-line rewards of DQfD with some
    losses removed on the games of Montezuma's Revenge and
    Q-Bert. Removing either loss degrades the performance of the
    algorithm. The right plots compare DQfD with three algorithms from
    the related work section. The other approaches do not perform as
    well as DQfD, particularly on Montezuma's Revenge.}
  \label{fig:losses}
\vspace{-0.2cm}
\end{figure*}

Figure~\ref{fig:losses} shows comparisons of DQfD with $\lambda_1$ and $\lambda_2$ set to 0,
on two games where DQfD achieved state-of-the-art
results: Montezuma's Revenge and Q-Bert. As expected, pre-training
without any supervised loss results in a network trained towards
ungrounded Q-learning targets and the agent starts with much lower
performance and is slower to improve. Removing the n-step TD loss has
nearly as large an impact on initial performance, as the n-step TD
loss greatly helps in learning from the limited demonstration
dataset. 


The right subplots in Figure~\ref{fig:losses} compare DQfD with three
related algorithms for leveraging demonstration data in DQN:
\begin{itemize}[noitemsep,nolistsep]
\item Replay Buffer Spiking (RBS)~\cite{LiptonGLLAD16}
\item Human Experience Replay (HER)~\cite{Hosu2016Playing}
\item Accelerated DQN with Expert Trajectories
  (ADET)~\cite{Lakshminarayanan:2016}
\end{itemize}
RBS is simply PDD DQN with the replay buffer initially full of
demonstration data. HER keeps the demonstration data and mixes
demonstration and agent data in each mini-batch. ADET is essentially
DQfD with the large margin supervised loss replaced with a
cross-entropy loss. The results show that all three of these
approaches are worse than DQfD in both games. Having a supervised loss
is critical to good performance, as both DQfD and ADET perform much
better than the other two algorithms. All the algorithms use the
exact same demonstration data used for DQfD. We included the prioritized
replay mechanism and the n-step returns in all of these algorithms
to make them as strong a comparison as possible.

\section{Discussion}
\label{sec:discussion}

The learning framework that we have presented in this paper is one
that is very common in real world problems such as controlling data
centers, autonomous vehicles~\cite{MLJ12-hester}, or recommendation
systems~\cite{Shani:2005}. In these problems, typically there is no
accurate simulator available, and learning must be performed on the
real system with real consequences. However, there is often data
available of the system being operated by a previous controller.  We
have presented a new algorithm called DQfD that takes advantage of
this data to accelerate learning on the real system.  It first
pre-trains solely on demonstration data, using a combination of 1-step
TD, n-step TD, supervised, and regularization losses so that it has a
reasonable policy that is a good starting point for learning in the
task. Once it starts interacting with the task, it continues learning
by sampling from both its self-generated data as well as the
demonstration data.  The ratio of both types of data in each
mini-batch is automatically controlled by a prioritized-replay
mechanism.

We have shown that DQfD gets a large boost in initial performance
compared to PDD DQN. DQfD has better performance on the first million
steps than PDD DQN on 41 of 42 Atari games, and on average it takes
DQN 82 million steps to match DQfD's performance.  On most real world
tasks, an agent may never get hundreds of millions of steps from which
to learn. We also showed that DQfD out-performs three other algorithms
for leveraging demonstration data in RL.
The fact that DQfD out-performs all these algorithms
makes it clear that it is the better choice for any real-world
application of RL where this type of demonstration data is available.

In addition to its early performance boost, DQfD is able to leverage
the human demonstrations to achieve state-of-the-art results on 11
Atari games. Many of these games are the hardest exploration games
(i.e. Montezuma's Revenge, Pitfall, Private Eye) where the
demonstration data can be used in place of smarter exploration. This
result enables the deployment of RL to problems where more
intelligent exploration would otherwise be required.

DQfD achieves these results despite having a very small amount of
demonstration data (5,574 to 75,472 transitions per game) that can
be easily generated in just a few minutes of gameplay. DQN and
DQfD receive three orders of magnitude more interaction data for RL
than demonstration data.  DQfD demonstrates the gains that can be
achieved by adding just a small amount of demonstration data with the
right algorithm. As the related work comparison shows, naively adding
(e.g. only pre-training or filling the replay buffer) this small
amount of data to a pure deep RL algorithm does not provide similar
benefit and can sometimes be detrimental.

These results may seem obvious given that DQfD has access to
privileged data, but the rewards and demonstrations are mathematically
dissimilar training signals, and naive approaches to combining them
can have disastrous results. Simply doing supervised learning on the
human demonstrations is not successful, while DQfD learns to
out-perform the best demonstration in 14 of 42 games. DQfD also
out-performs three prior algorithms for incorporating demonstration
data into DQN. We argue that the combination of all four losses during
pre-training is critical for the agent to learn a coherent
representation that is not destroyed by the switch in training signals
after pre-training. Even after pre-training, the agent must continue
using the expert data. In particular, the right sub-figure of Figure 1
shows that the ratio of expert data needed (selected by prioritized
replay) grows during the interaction phase for the most difficult
exploration games, where the demonstration data becomes more useful as
the agent reaches new screens in the game. RBS shows an example where
just having the demonstration data initially is not enough to provide
good performance.

Learning from human demonstrations is particularly difficult. In most
games, imitation learning is unable to perfectly
classify the demonstrator's actions even on the demonstration
dataset. Humans may play the games in a way that differs greatly from
a policy that an agent would learn, and may be using information that
is not available in the agent's state representation. In future work,
we plan to measure these differences between demonstration and agent
data to inform approaches that derive more value from the
demonstrations. Another future direction is to apply these concepts to
domains with continuous actions, where the classification loss becomes
a regression loss.

\section*{Acknowledgments} 
The authors would like to thank Keith Anderson, Chris Apps, Ben
Coppin, Joe Fenton, Nando de Freitas, Chris Gamble, Thore Graepel,
Georg Ostrovski, Cosmin Paduraru, Jack Rae, Amir Sadik, Jon Scholz,
David Silver, Toby Pohlen, Tom Stepleton, Ziyu Wang, and many others
at DeepMind for insightful discussions, code contributions, and other
efforts.

\balance
\bibliography{paper}
\bibliographystyle{aaai}

\clearpage
\section*{Supplementary Material}

Here are the parameters used for the three algorithms. DQfD used all of these parameters, while the other two algorithms only used the applicable parameters.
\begin{itemize}[noitemsep,nolistsep]
\item Pre-training steps $k = 750,000$ mini-batch updates.
\item N-Step Return weight $\lambda_1 = 1.0$
\item Supervised loss weight $\lambda_2 = 1.0$.
\item L2 regularization weight $\lambda_3 = 10^{-5}$.
\item Expert margin $l(a_E, a) when a \neq a_E = 0.8$.
\item $\epsilon$-greedy exploration with $\epsilon = 0.01$, same used by Double DQN~\cite{vanHasselt:2016}.
\item Prioritized replay exponent $\alpha = 0.4$.
\item Prioritized replay constants $\epsilon_a = 0.001$, $\epsilon_d = 1.0$.
\item Prioritized replay importance sampling exponent $\beta_0 = 0.6$ as in \cite{SchaulQAS16}.
\item N-step returns with $n = 10$.
\item Target network update period $\tau = 10,000$ as in ~\cite{Mnih:2015}
\end{itemize}


\begin{figure*}[p]
  \centering
  \subfigure{\includegraphics[width=0.177\linewidth]{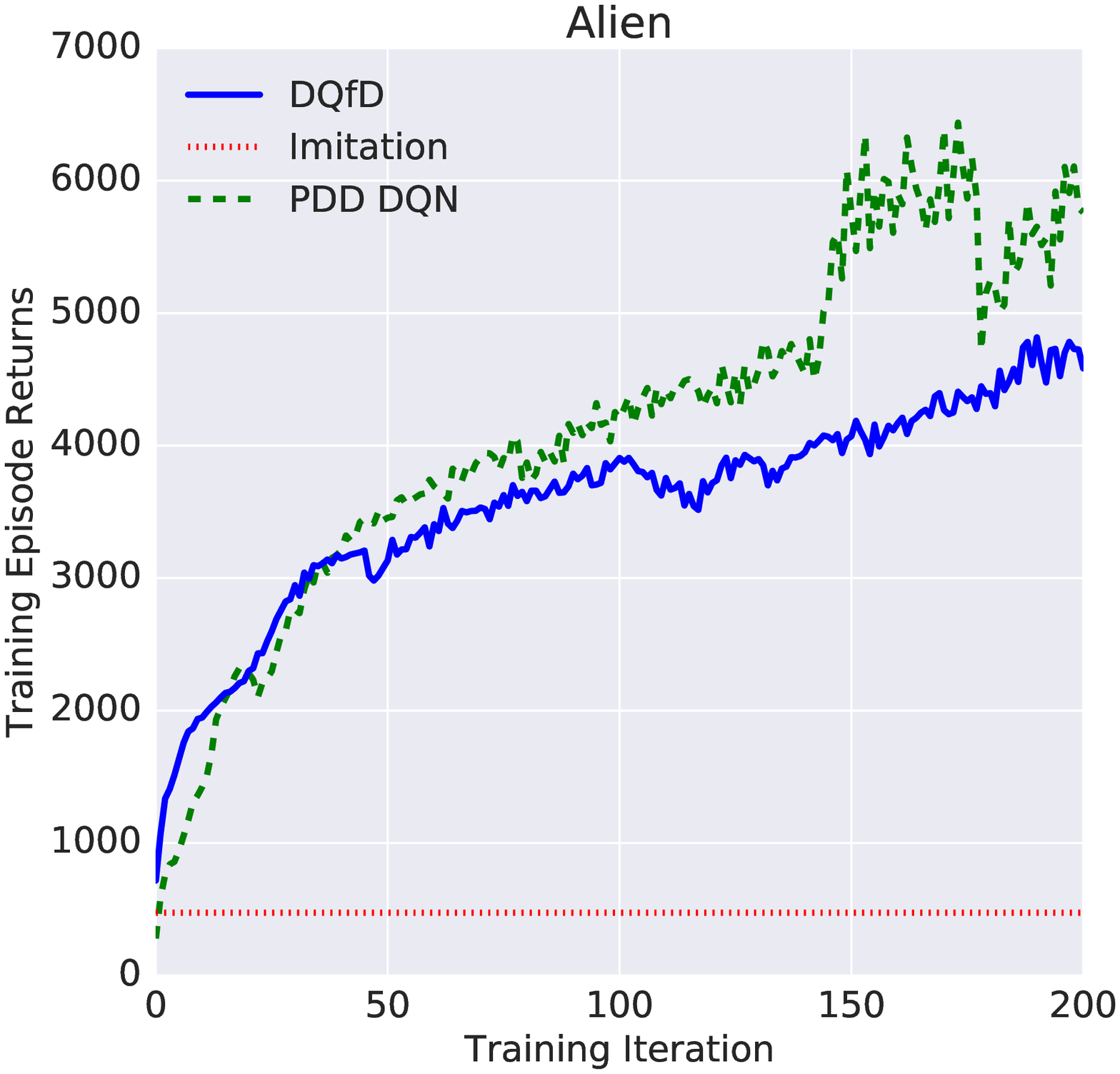}}
  \hspace{-0.02\linewidth}\subfigure{\includegraphics[width=0.177\linewidth]{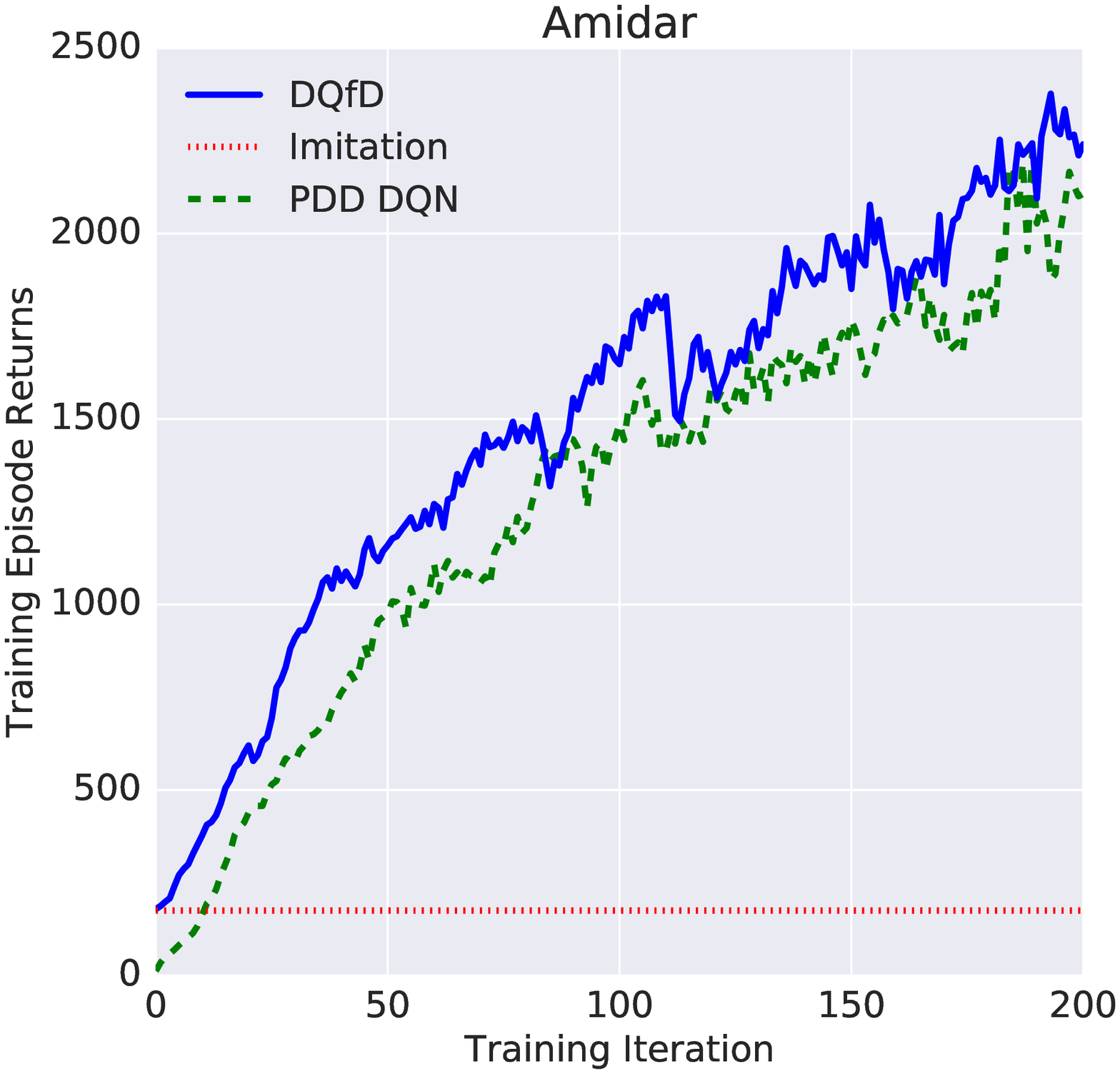}}
  \hspace{-0.02\linewidth}\subfigure{\includegraphics[width=0.177\linewidth]{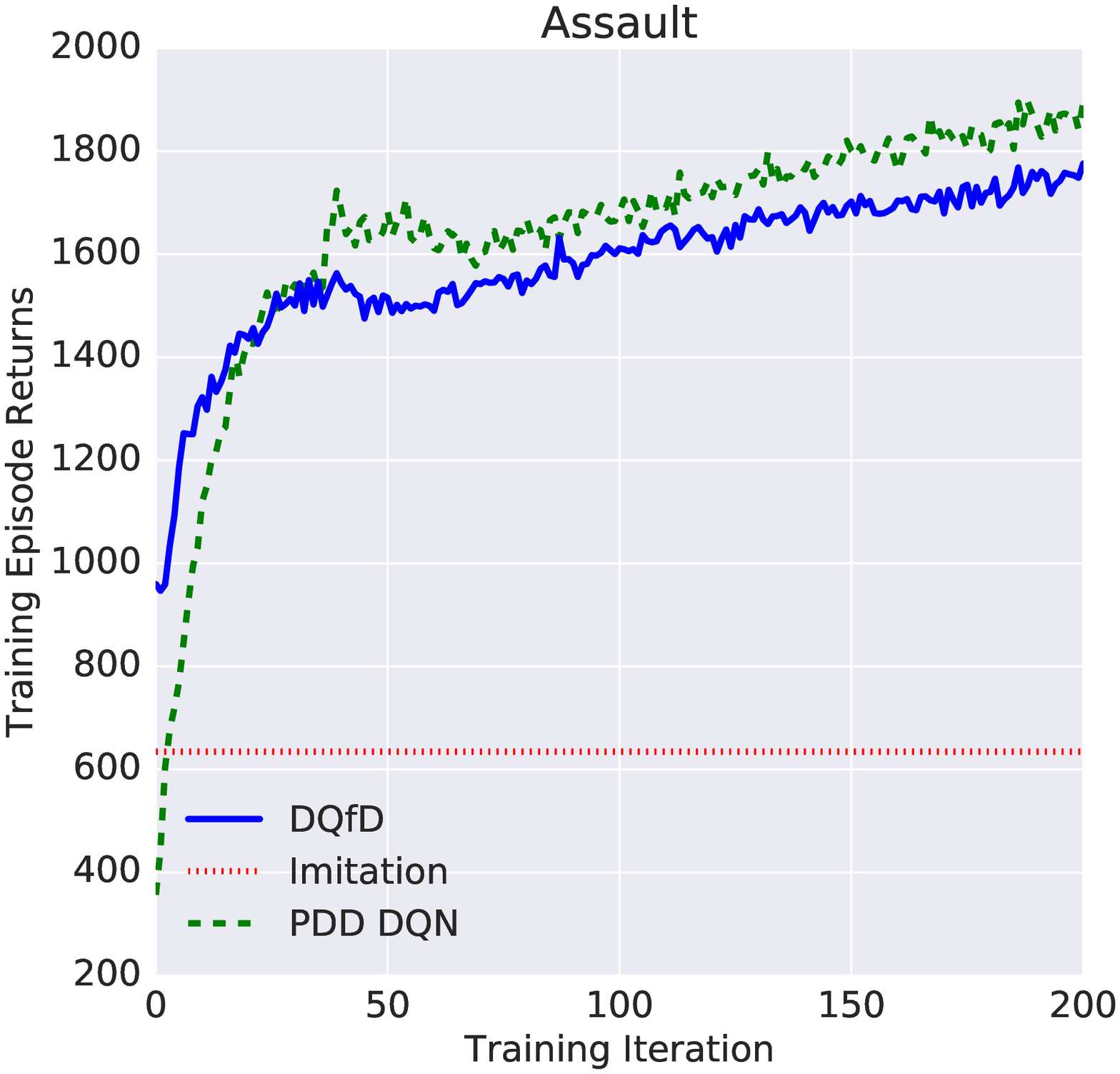}}
  \hspace{-0.02\linewidth}\subfigure{\includegraphics[width=0.177\linewidth]{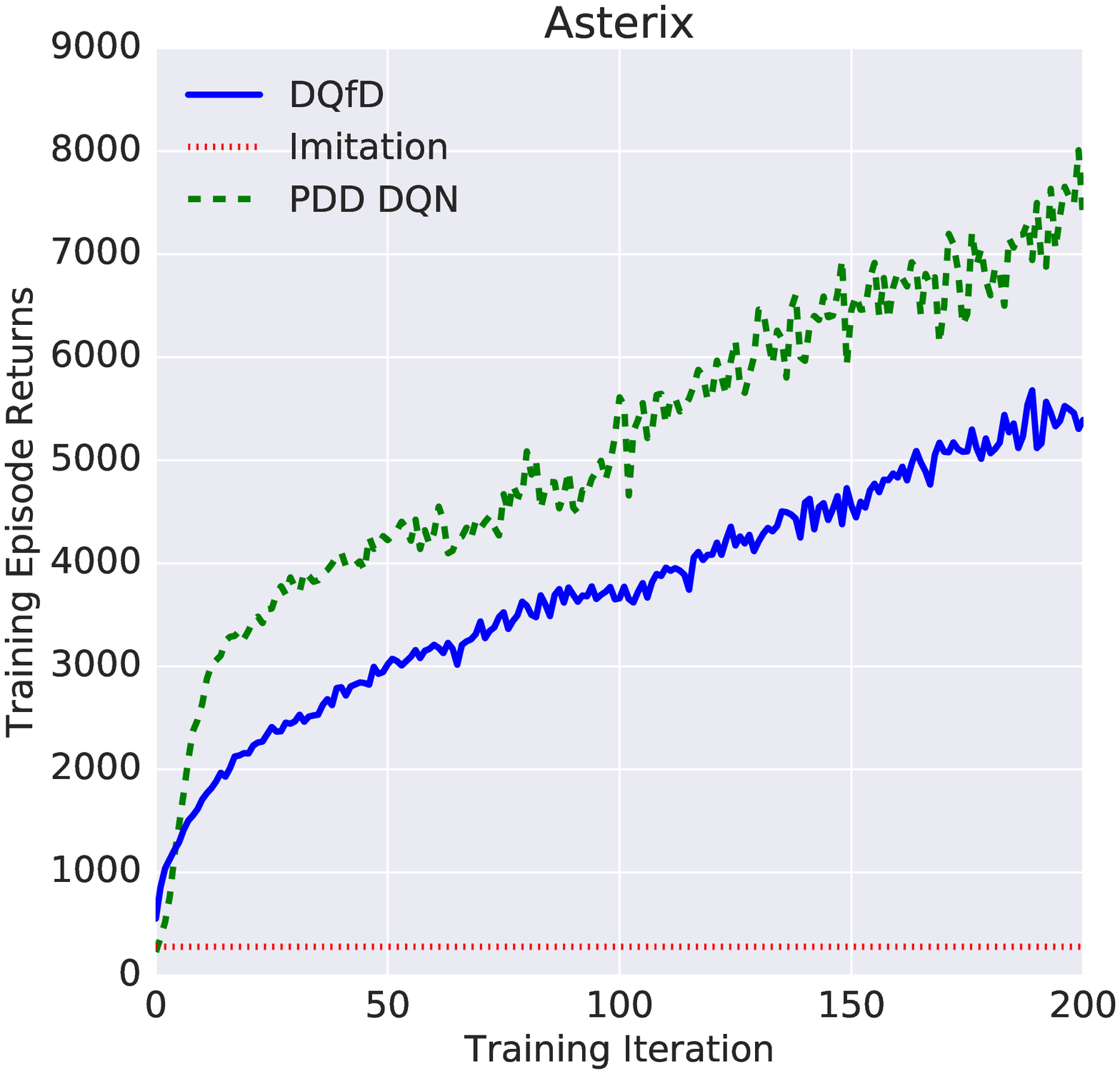}}
  \hspace{-0.02\linewidth}\subfigure{\includegraphics[width=0.177\linewidth]{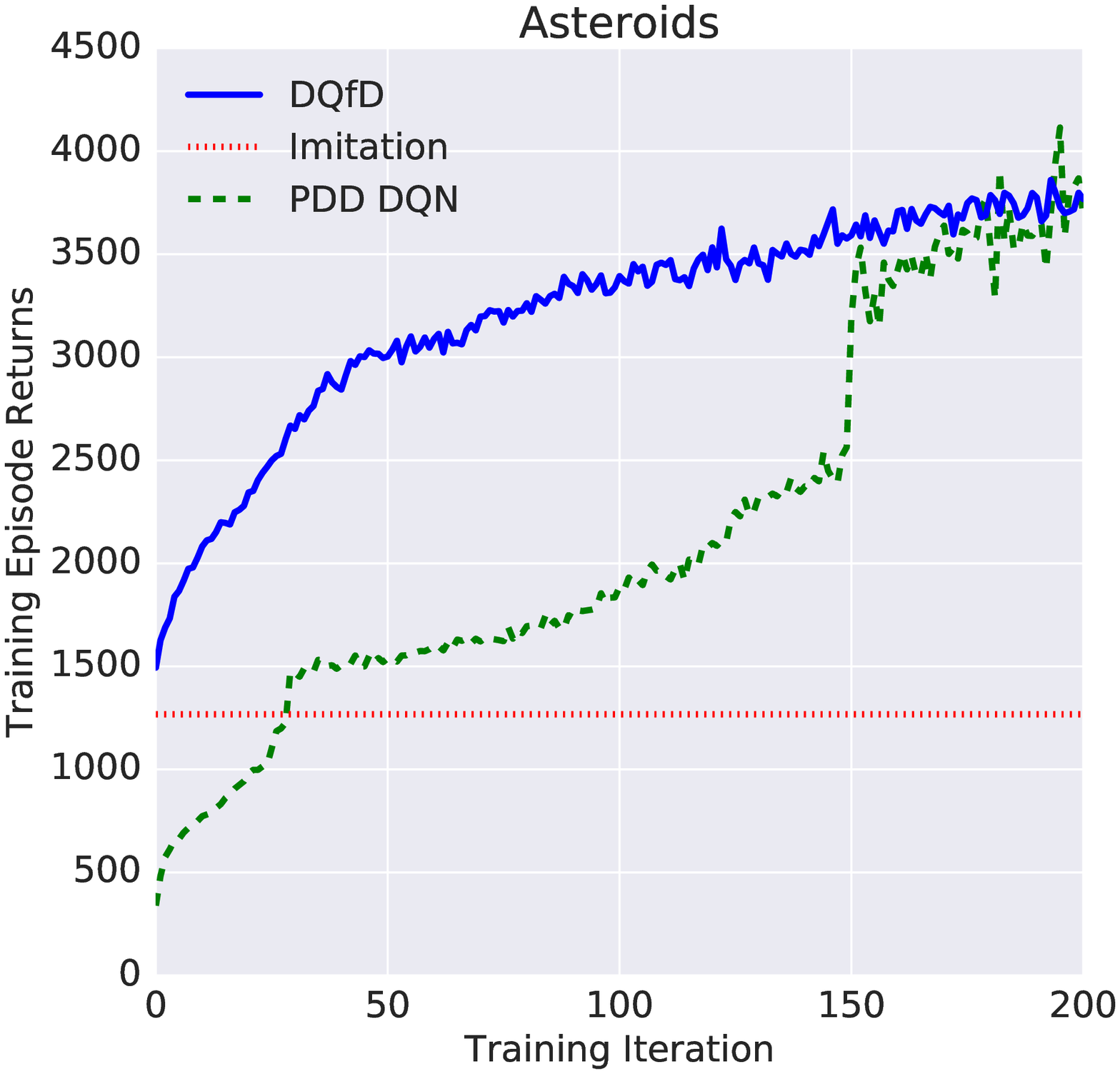}}
  \hspace{-0.02\linewidth}\subfigure{\includegraphics[width=0.177\linewidth]{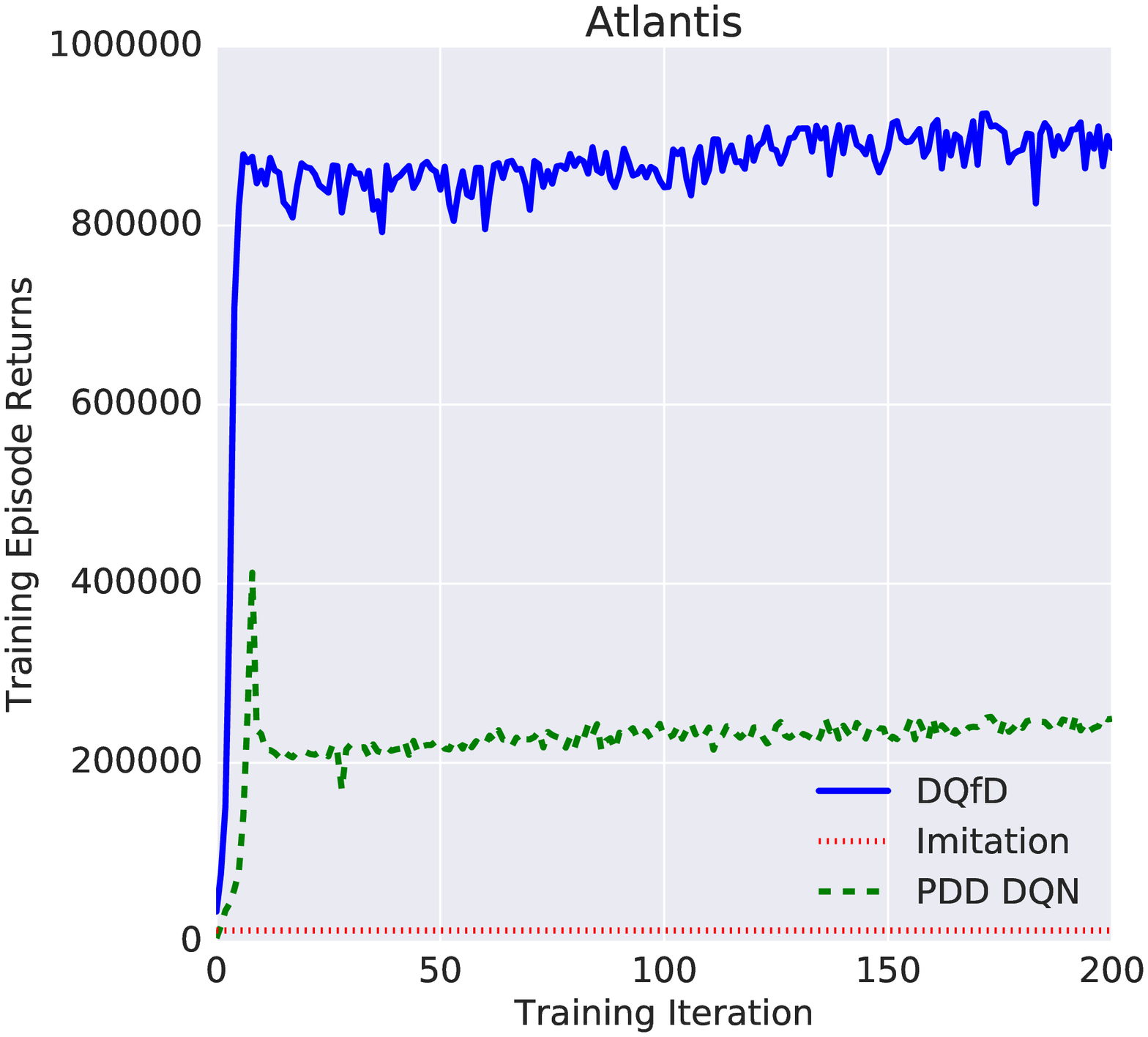}}
  \vspace{-0.25cm}
  \hspace{-0.02\linewidth}\subfigure{\includegraphics[width=0.177\linewidth]{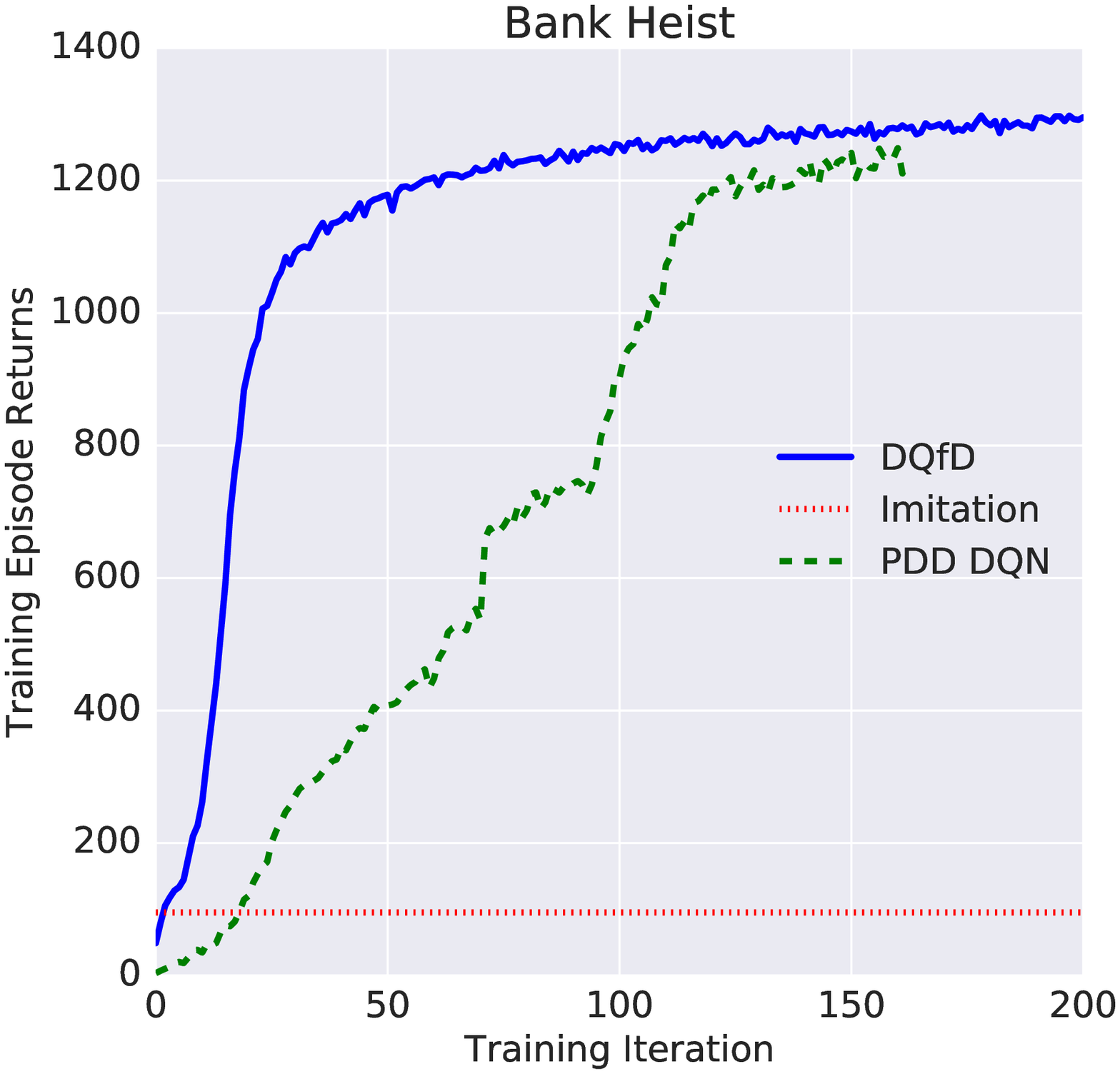}}
  \hspace{-0.02\linewidth}\subfigure{\includegraphics[width=0.177\linewidth]{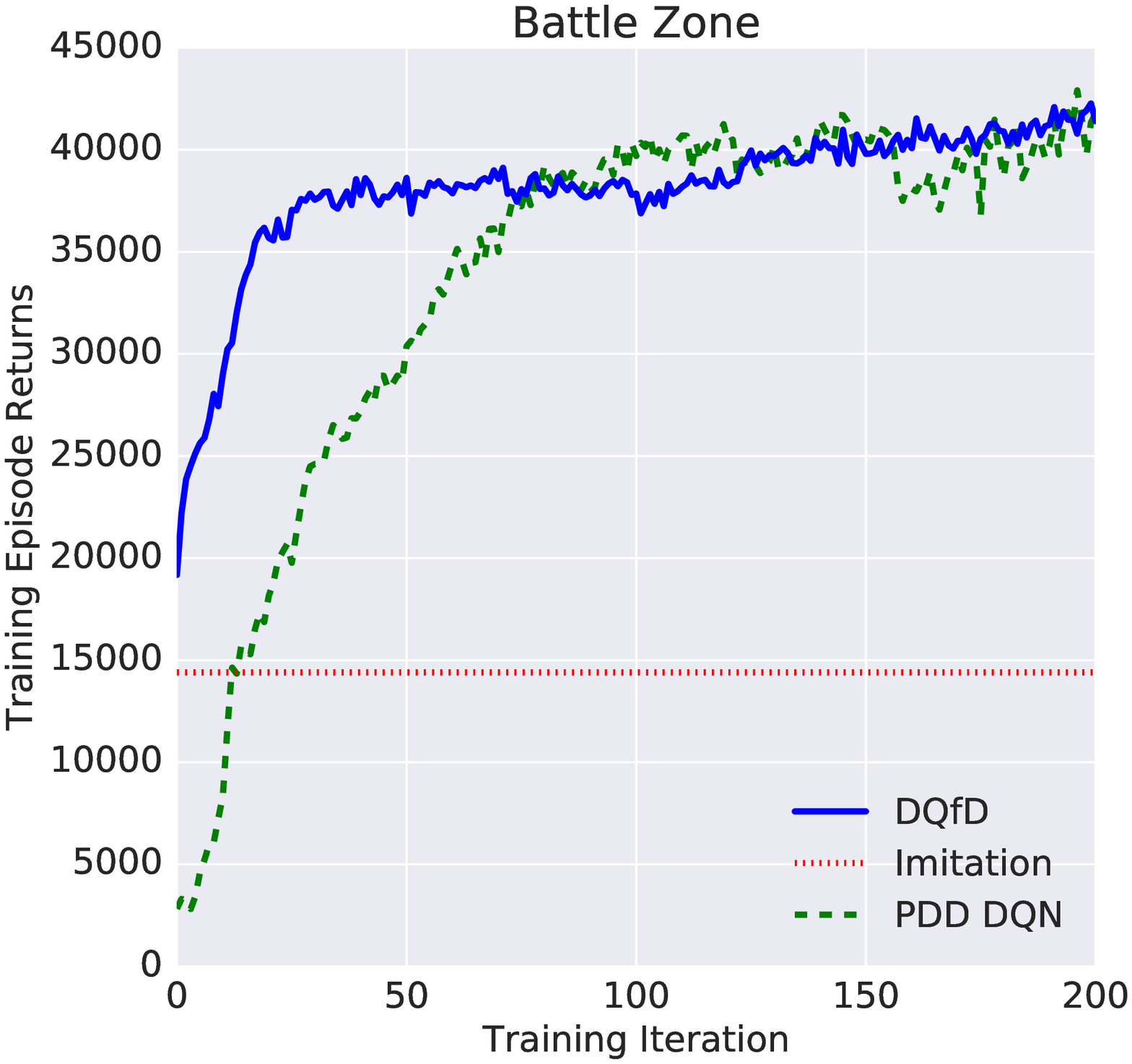}}
  \hspace{-0.02\linewidth}\subfigure{\includegraphics[width=0.177\linewidth]{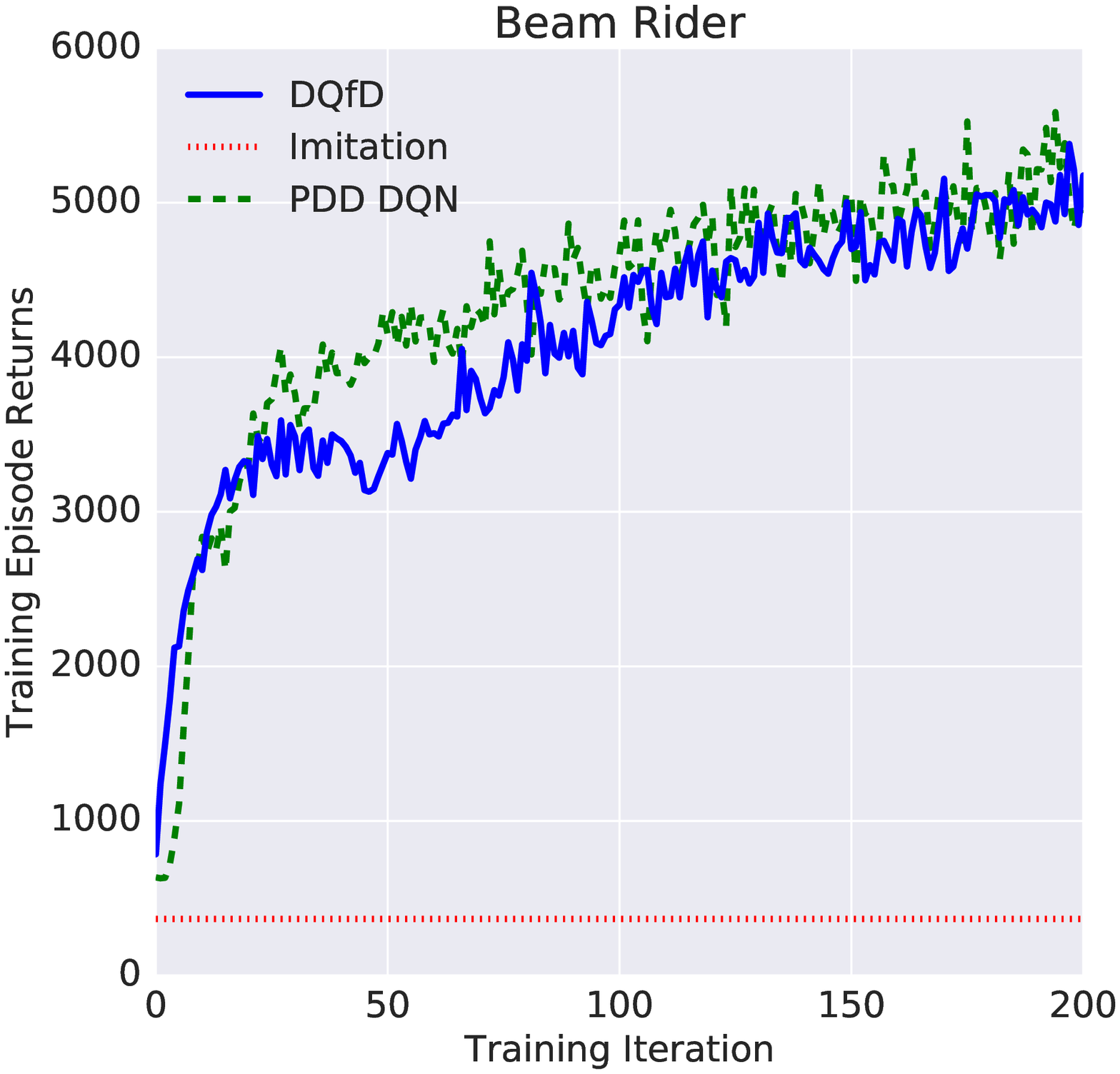}}
  \hspace{-0.02\linewidth}\subfigure{\includegraphics[width=0.177\linewidth]{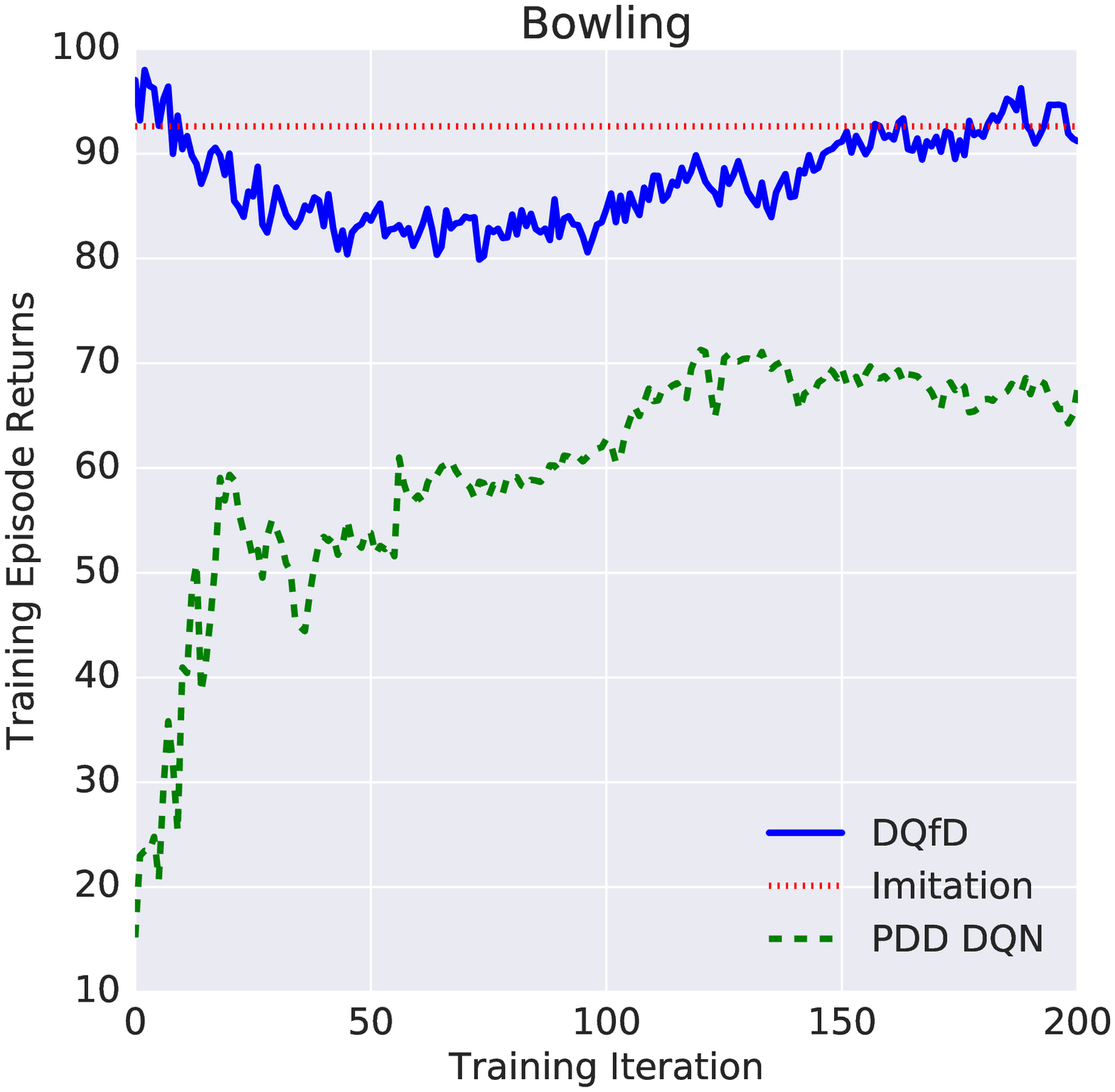}}
  \hspace{-0.02\linewidth}\subfigure{\includegraphics[width=0.177\linewidth]{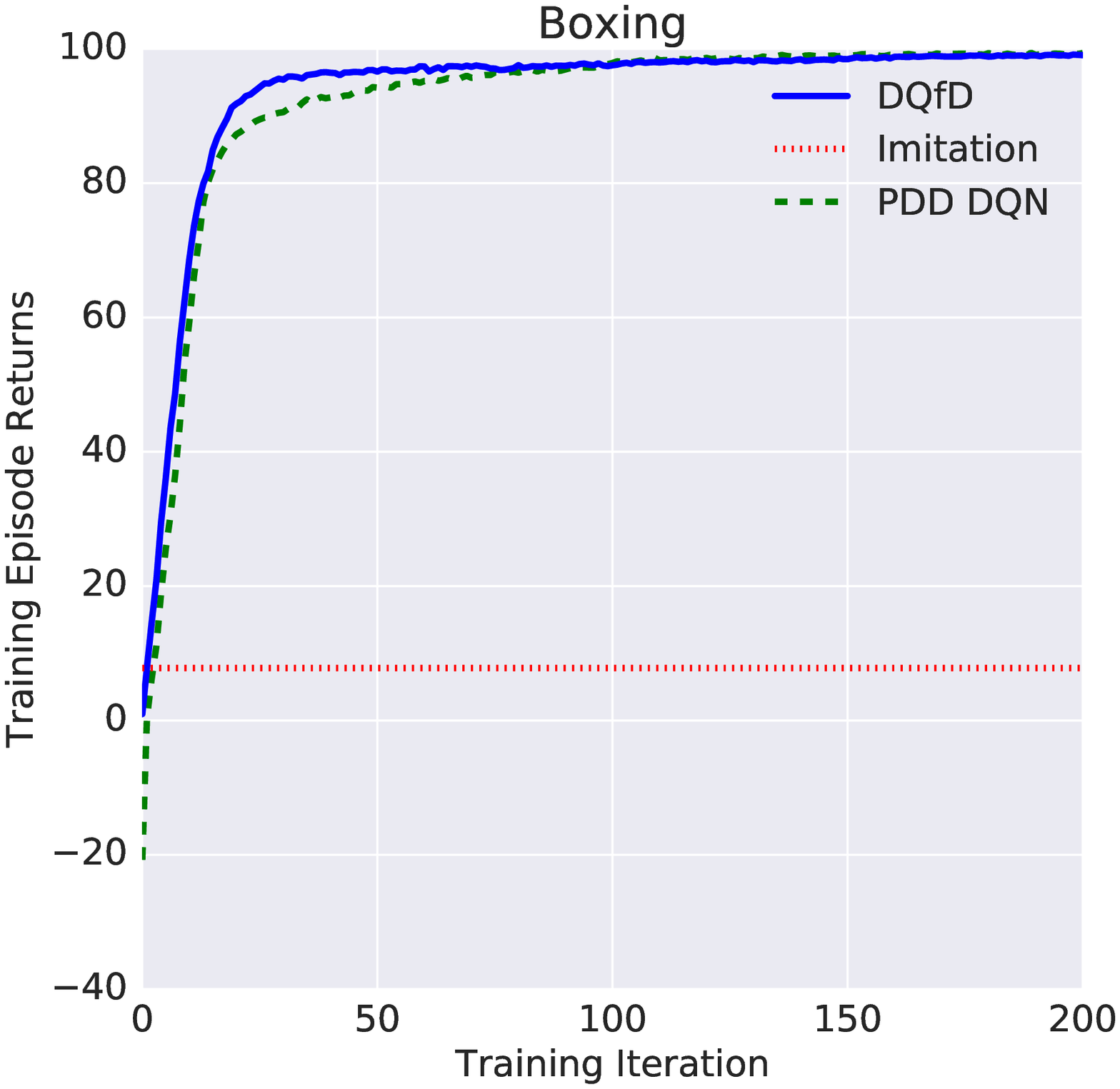}}
  \hspace{-0.02\linewidth}\subfigure{\includegraphics[width=0.177\linewidth]{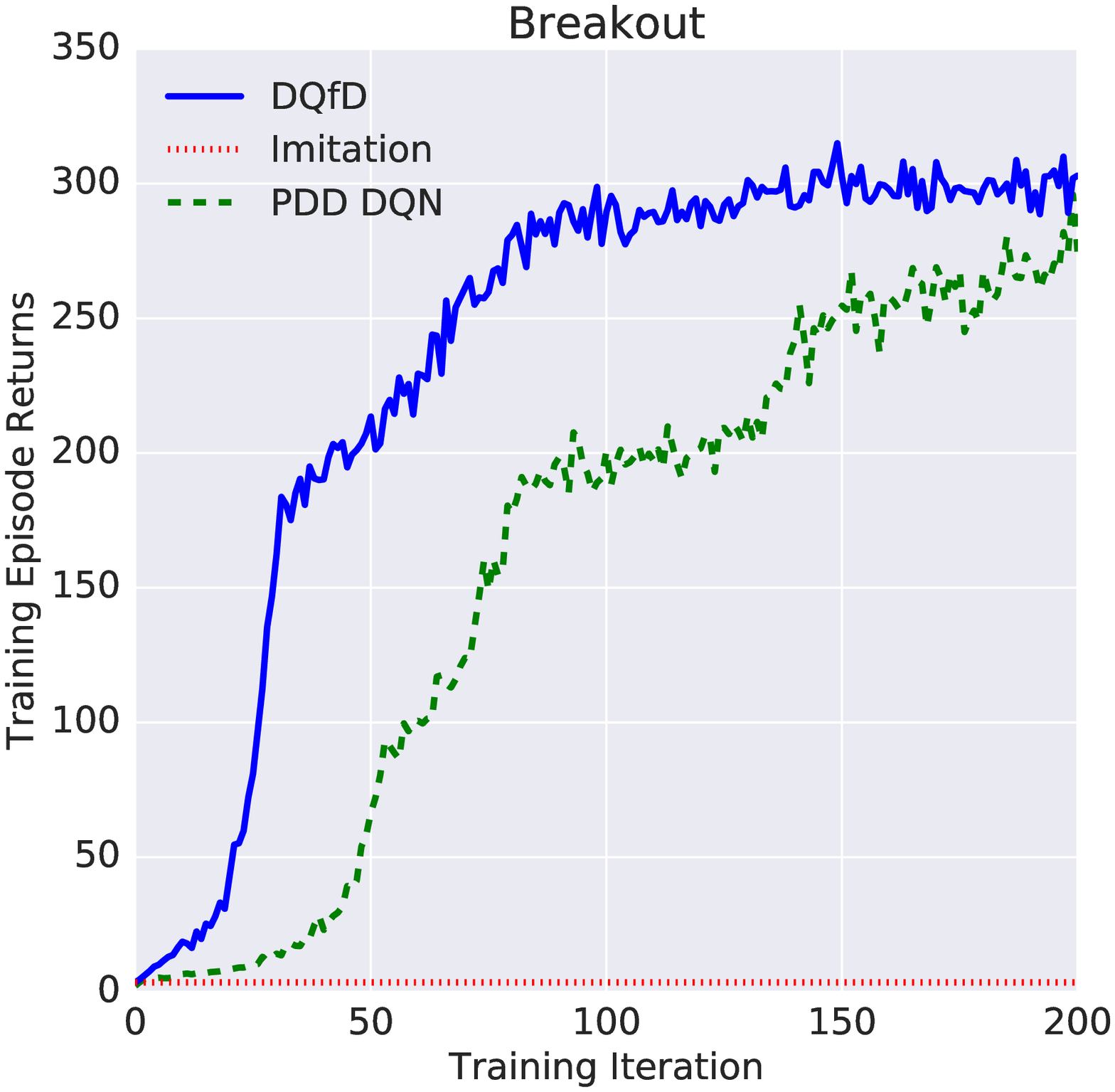}}
  \vspace{-0.25cm}
  \hspace{-0.02\linewidth}\subfigure{\includegraphics[width=0.177\linewidth]{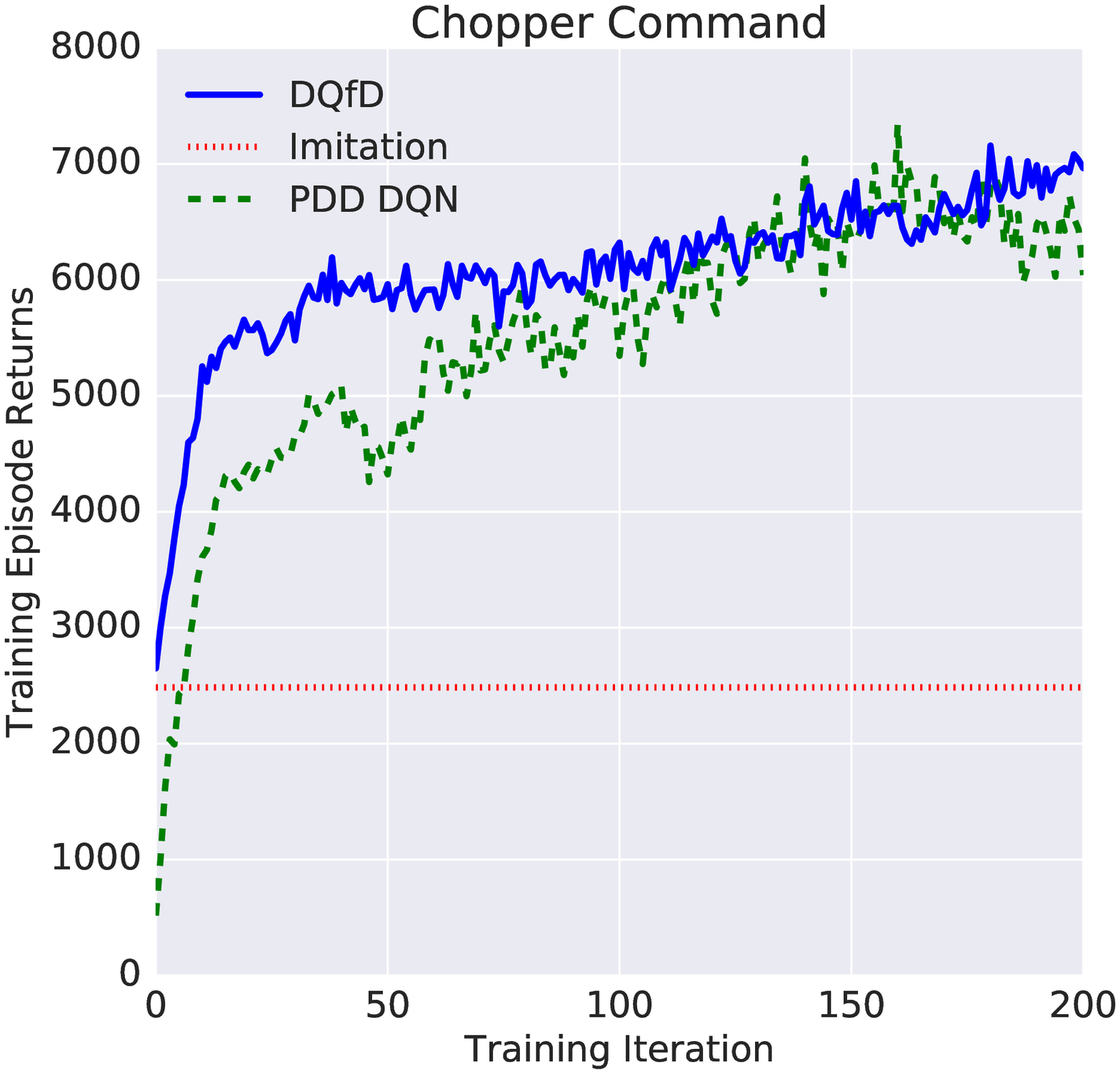}}
  \hspace{-0.02\linewidth}\subfigure{\includegraphics[width=0.177\linewidth]{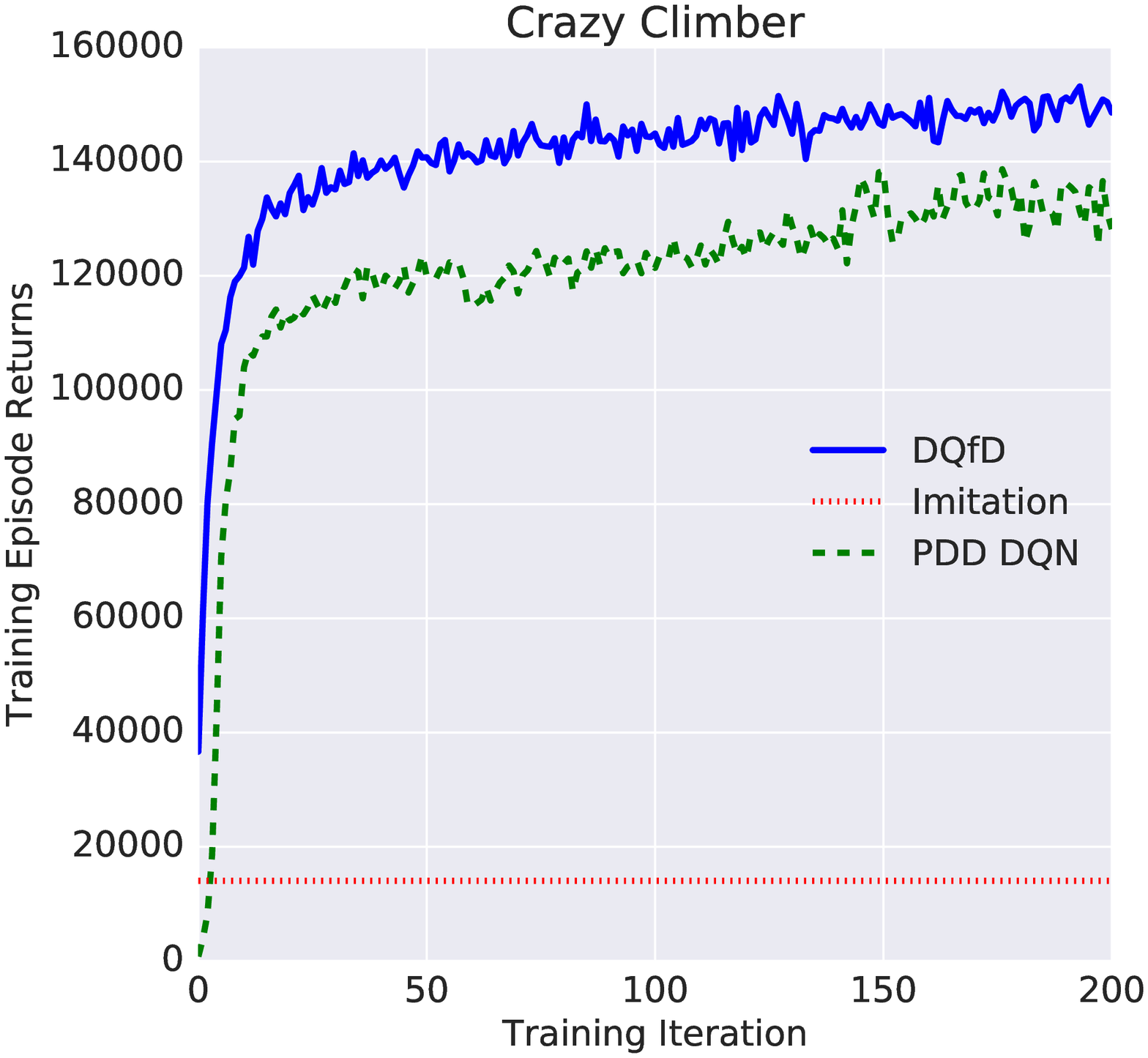}}
  \hspace{-0.02\linewidth}\subfigure{\includegraphics[width=0.177\linewidth]{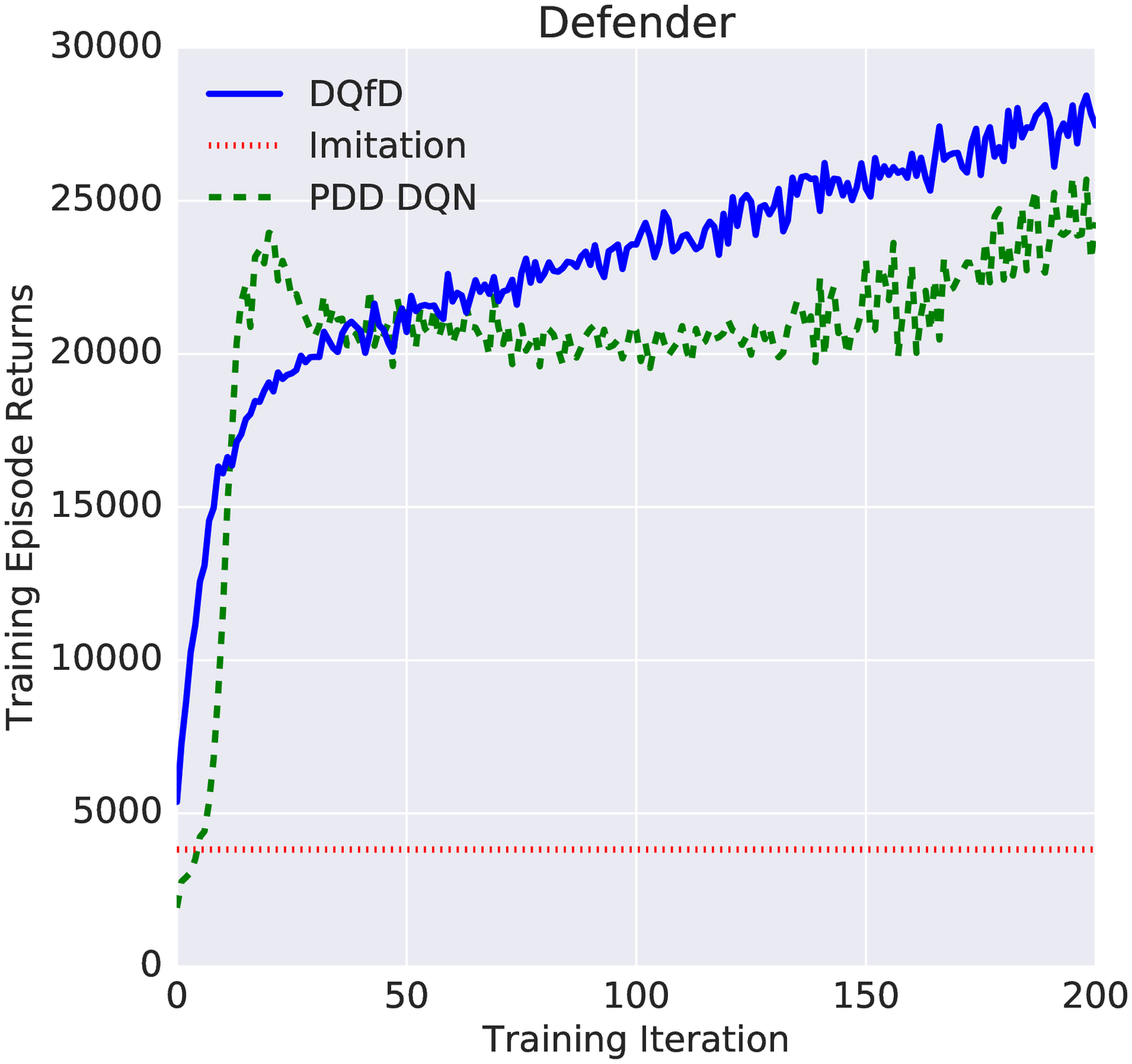}}
  \hspace{-0.02\linewidth}\subfigure{\includegraphics[width=0.177\linewidth]{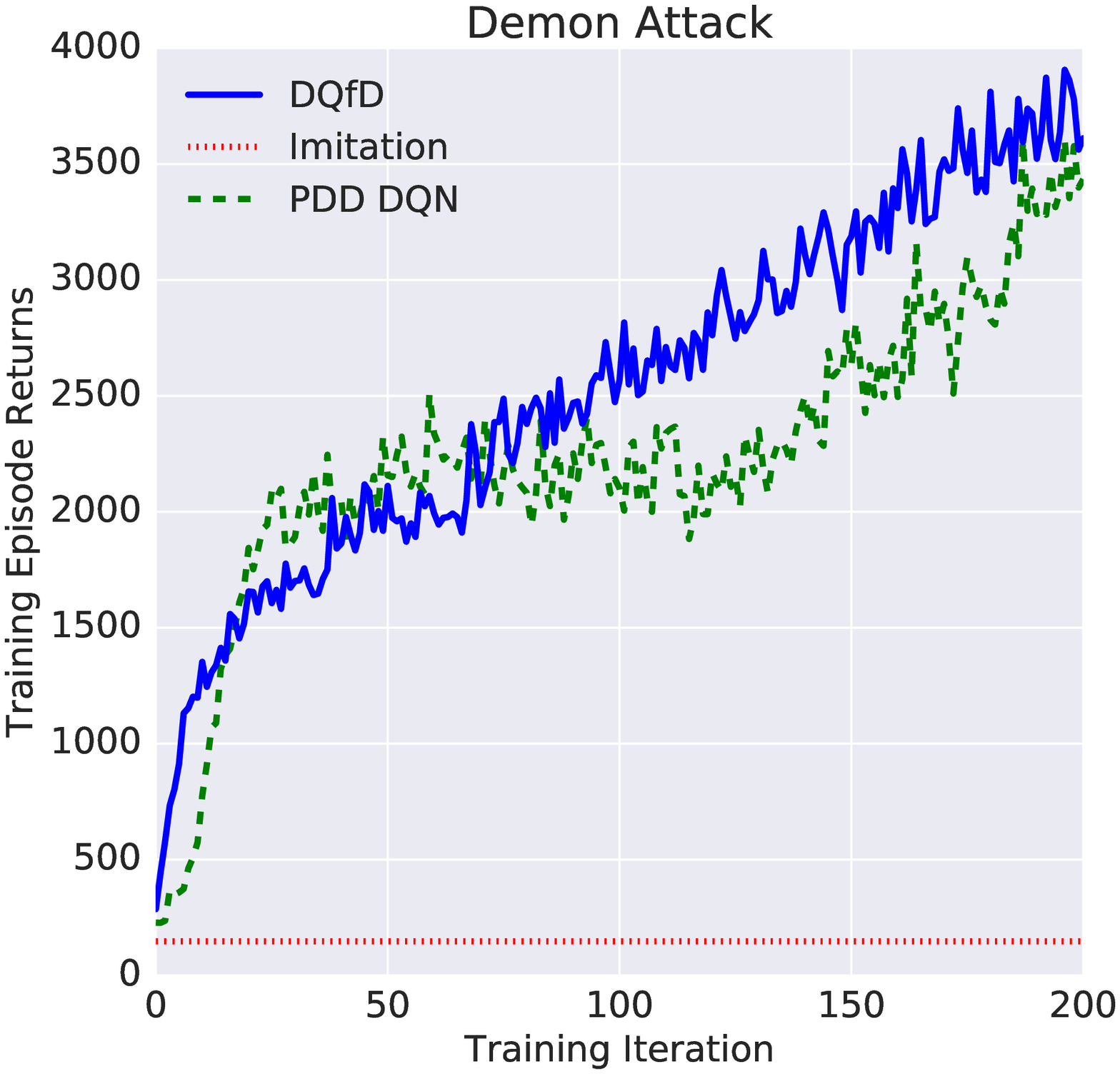}}
  \hspace{-0.02\linewidth}\subfigure{\includegraphics[width=0.177\linewidth]{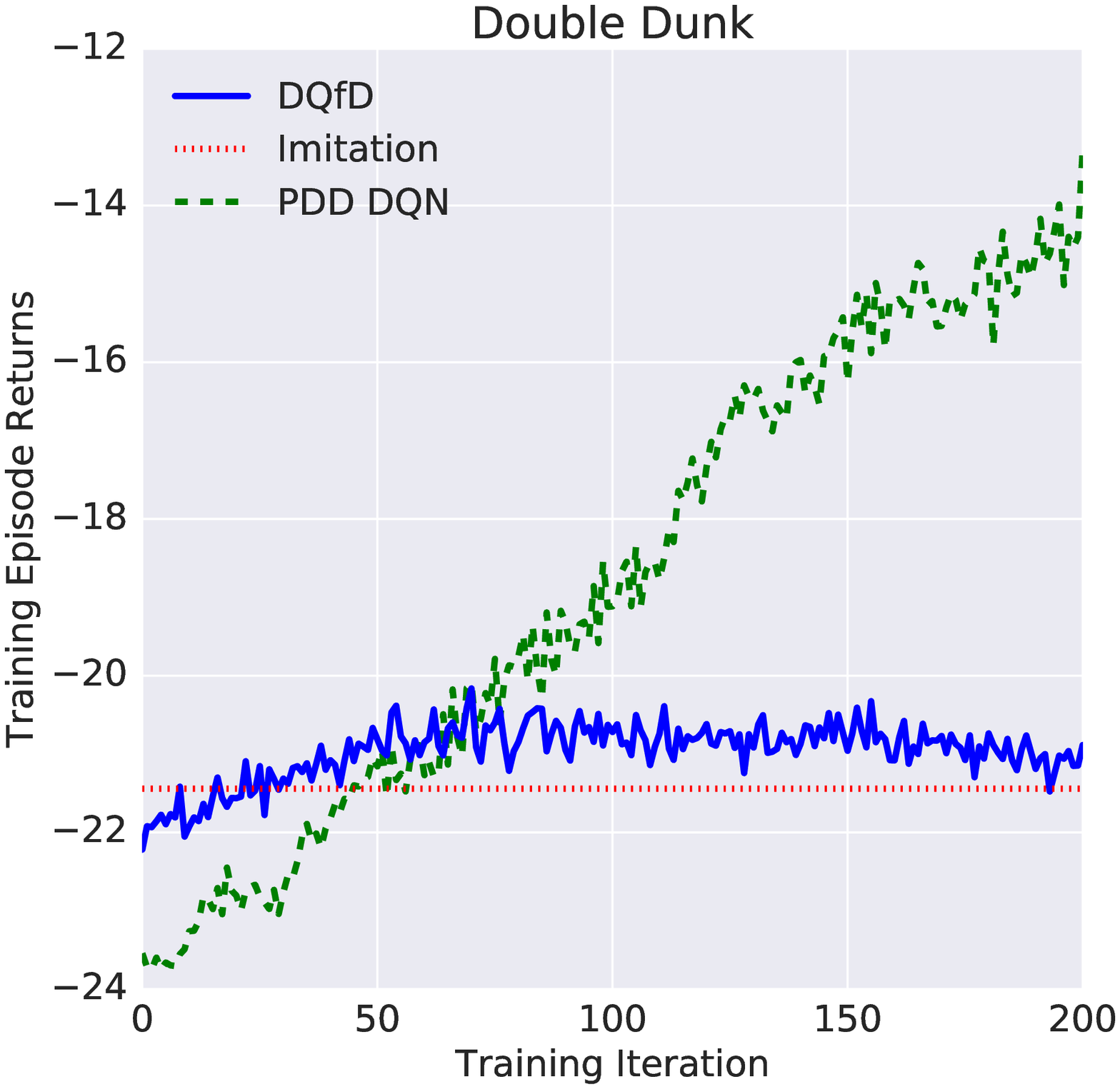}}
  \hspace{-0.02\linewidth}\subfigure{\includegraphics[width=0.177\linewidth]{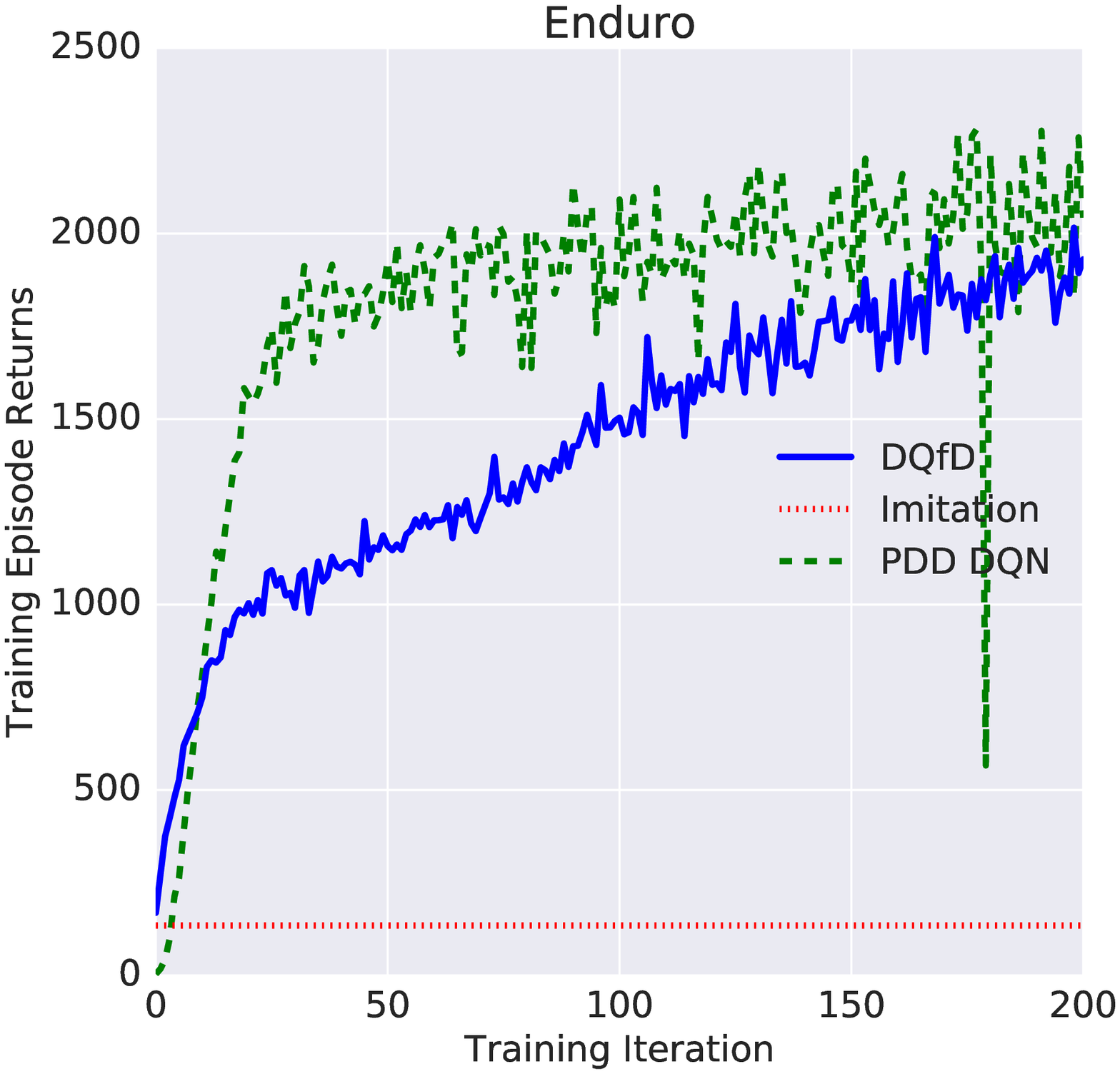}}
  \vspace{-0.25cm}
  \hspace{-0.02\linewidth}\subfigure{\includegraphics[width=0.177\linewidth]{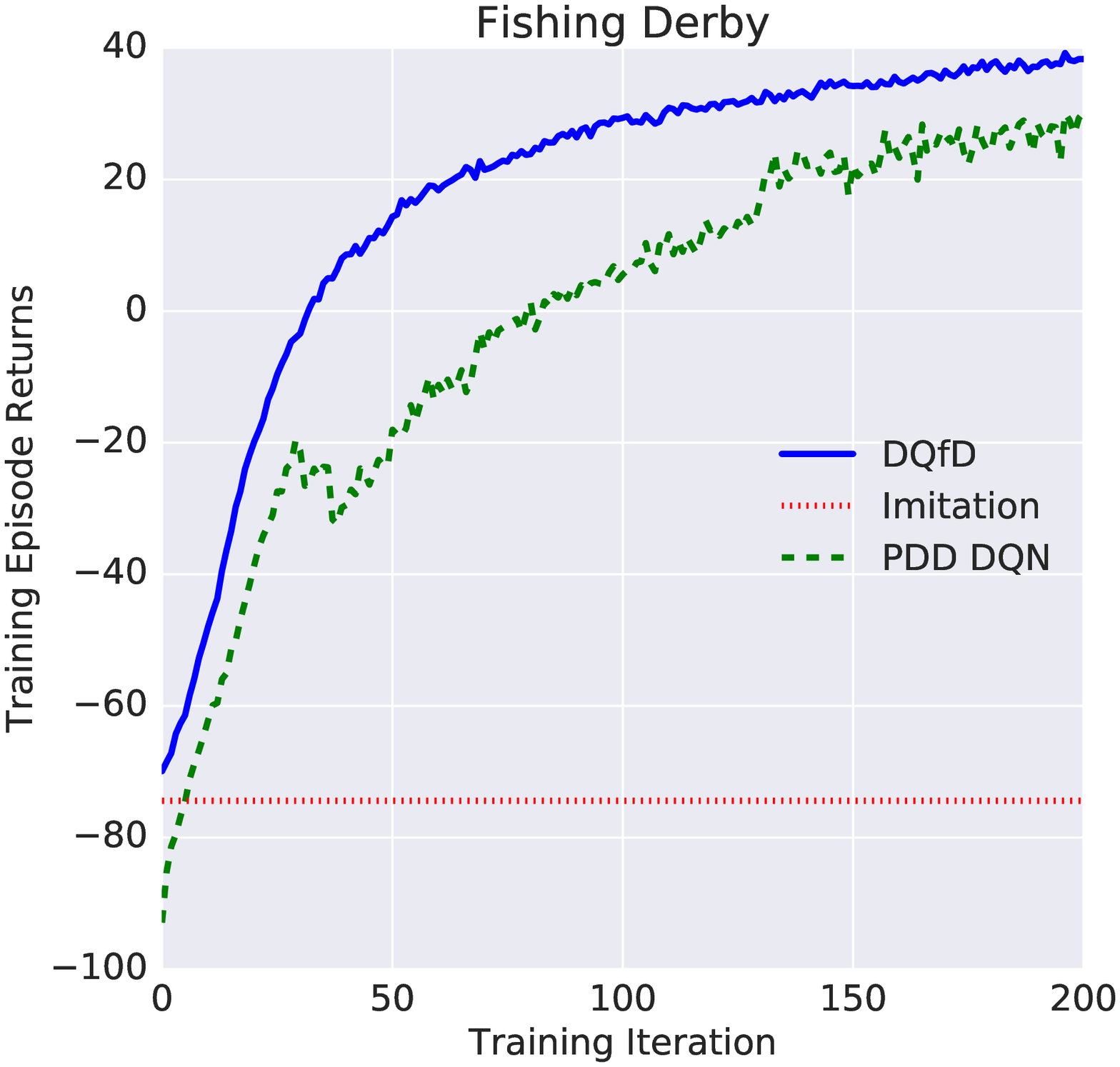}}
  \hspace{-0.02\linewidth}\subfigure{\includegraphics[width=0.177\linewidth]{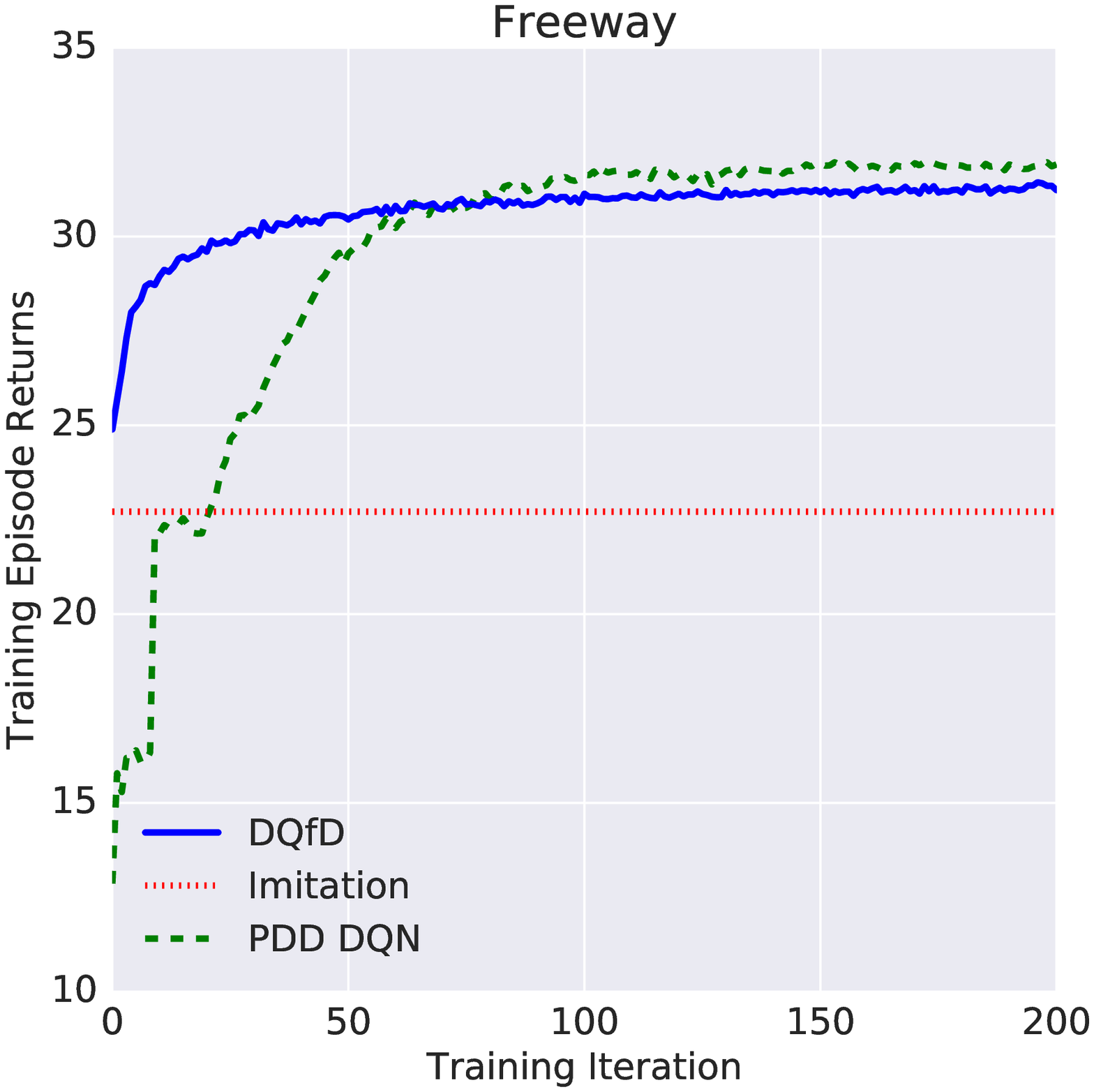}}
  \hspace{-0.02\linewidth}\subfigure{\includegraphics[width=0.177\linewidth]{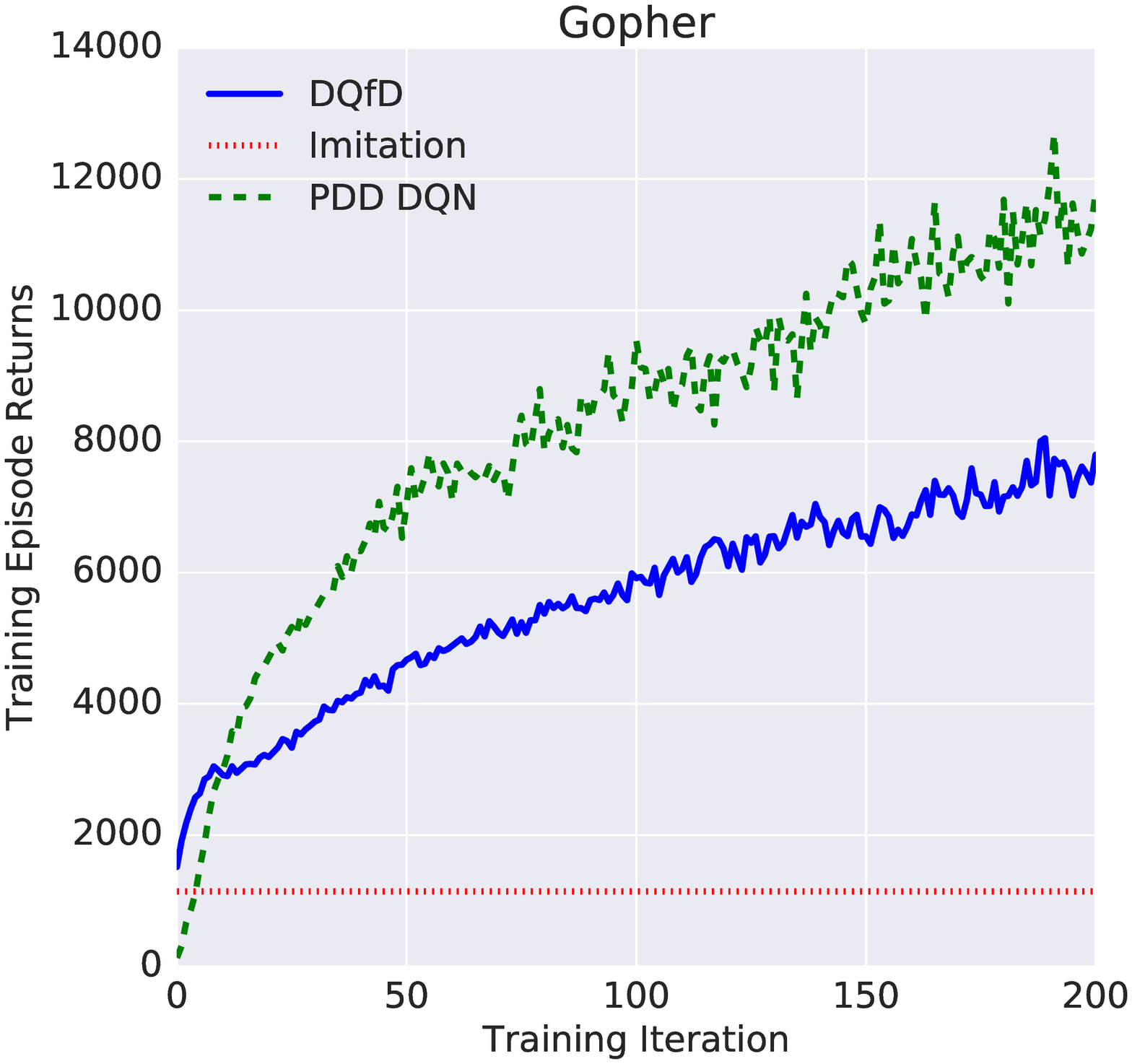}}
  \hspace{-0.02\linewidth}\subfigure{\includegraphics[width=0.177\linewidth]{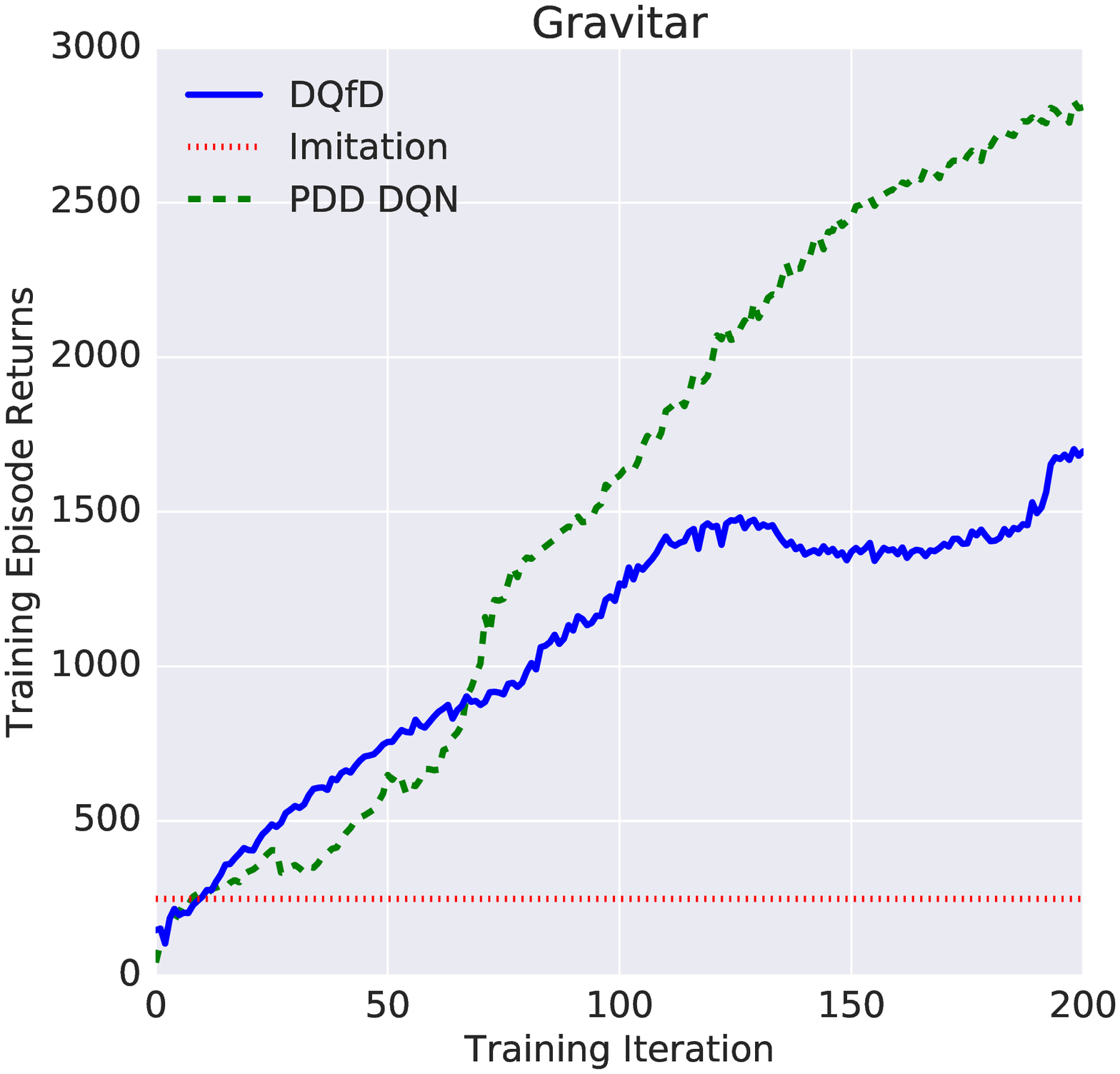}}
  \hspace{-0.02\linewidth}\subfigure{\includegraphics[width=0.177\linewidth]{figures/Hero_training_mean_rewards_new2.eps}}
  \hspace{-0.02\linewidth}\subfigure{\includegraphics[width=0.177\linewidth]{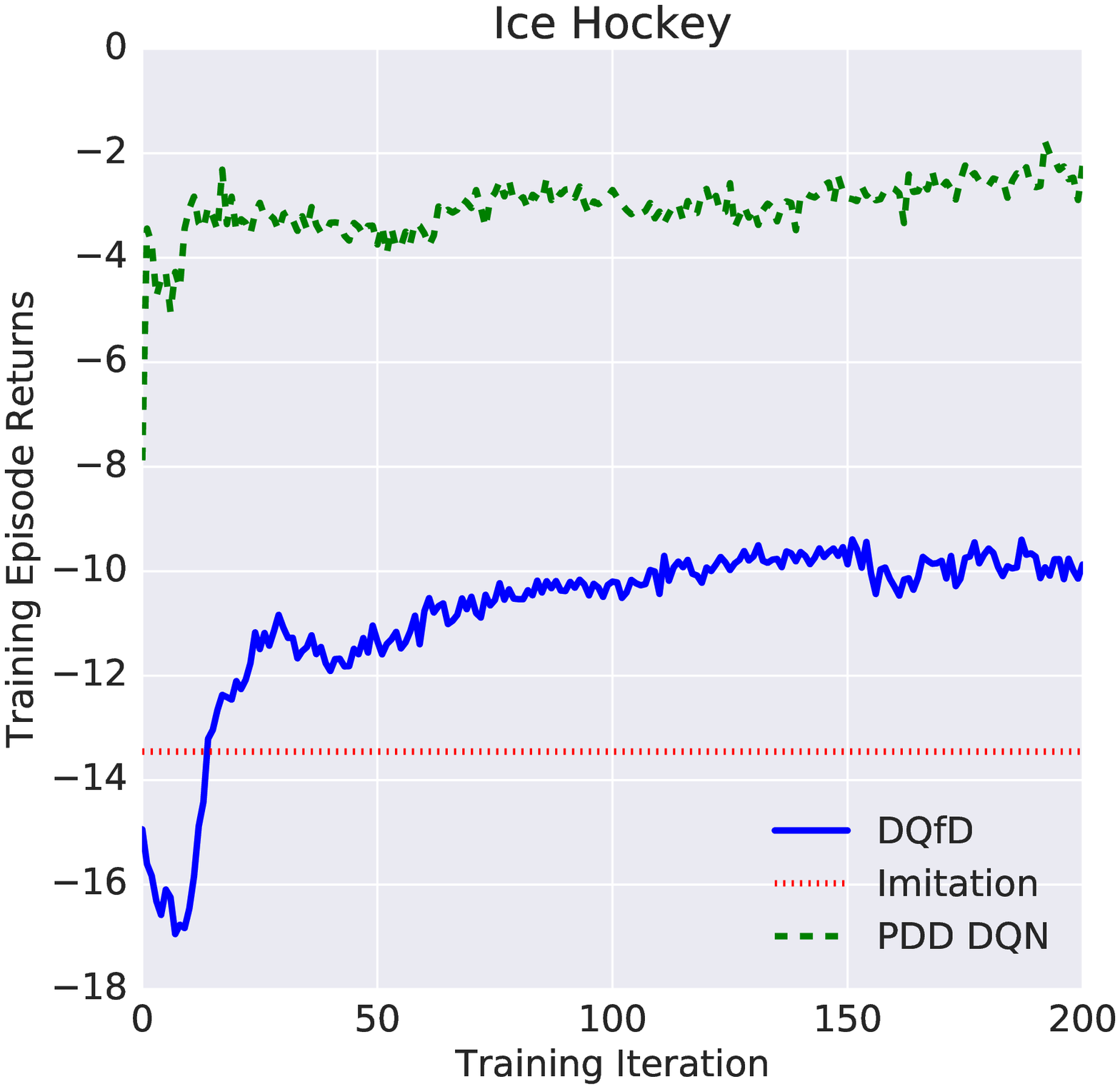}}
  \vspace{-0.25cm}
  \hspace{-0.02\linewidth}\subfigure{\includegraphics[width=0.177\linewidth]{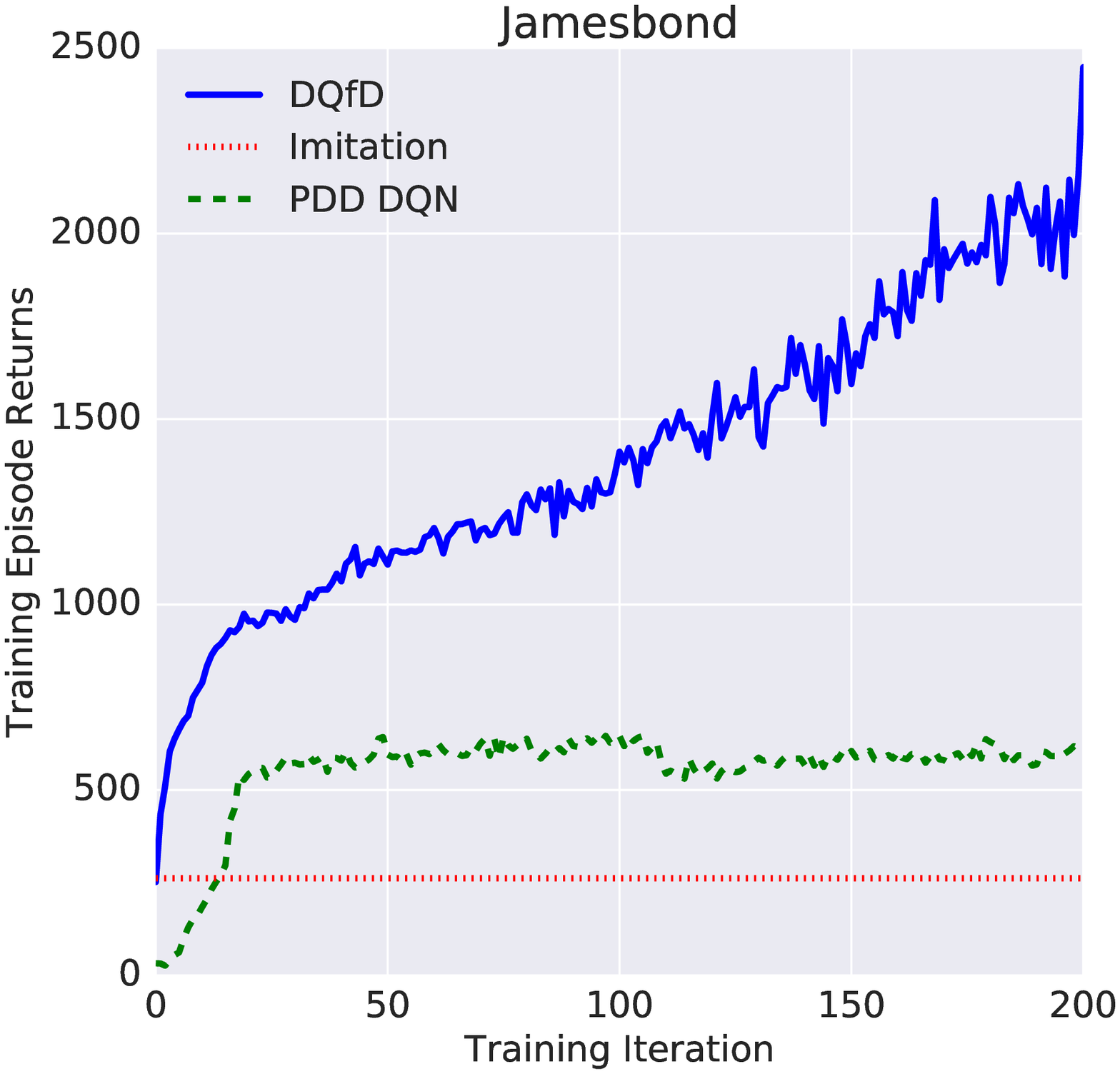}}
  \hspace{-0.02\linewidth}\subfigure{\includegraphics[width=0.177\linewidth]{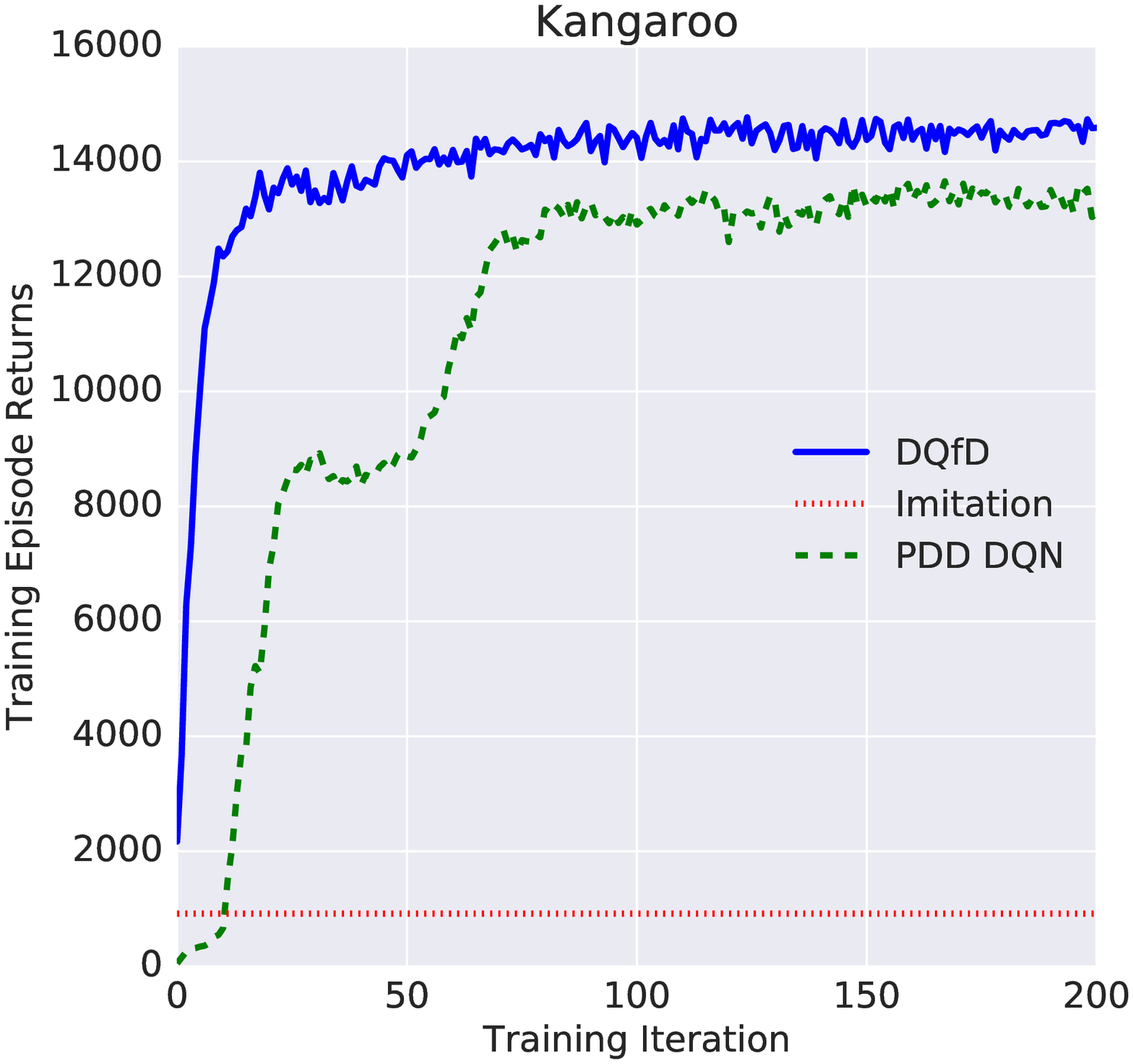}}
  \hspace{-0.02\linewidth}\subfigure{\includegraphics[width=0.177\linewidth]{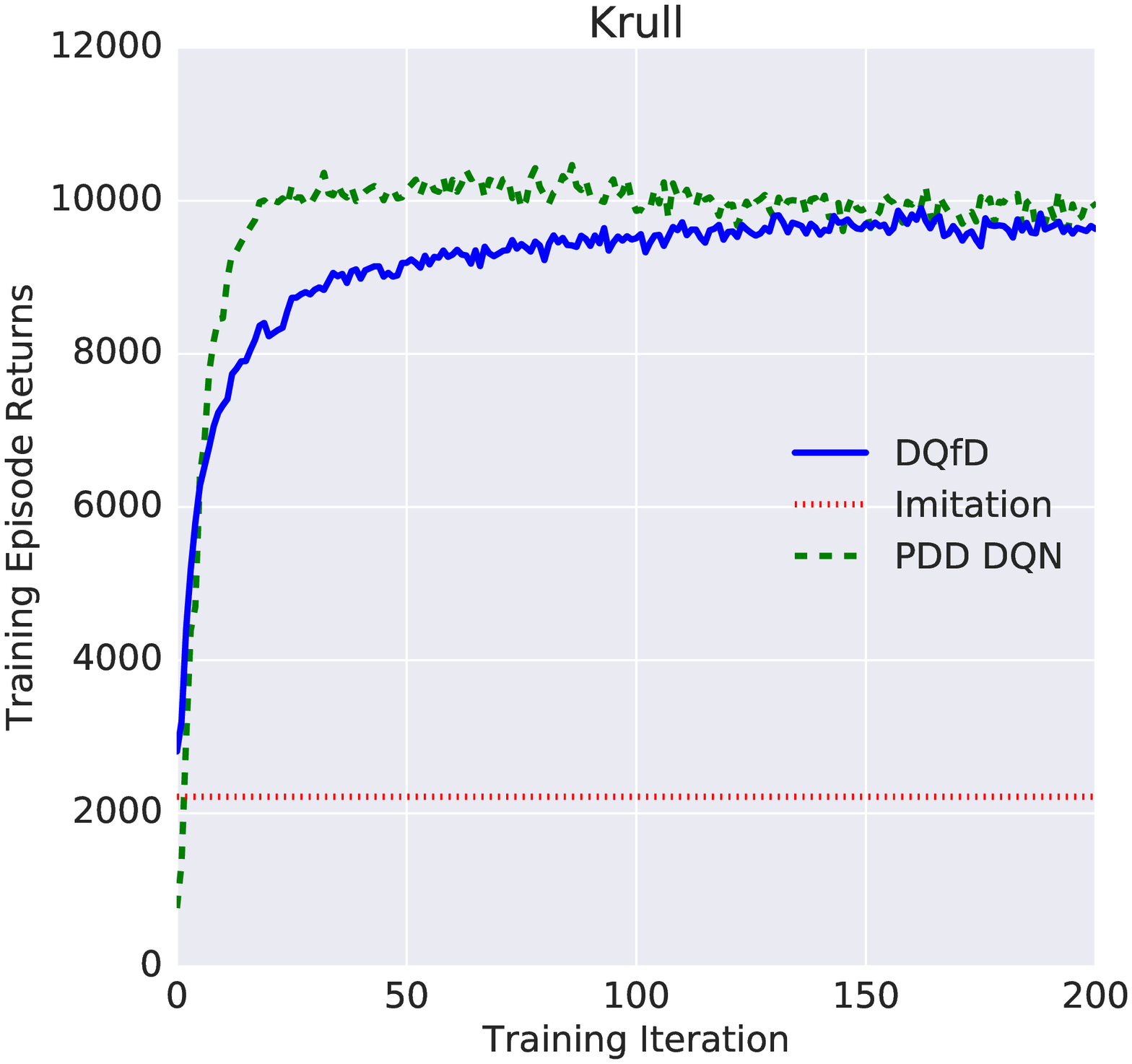}}
  \hspace{-0.02\linewidth}\subfigure{\includegraphics[width=0.177\linewidth]{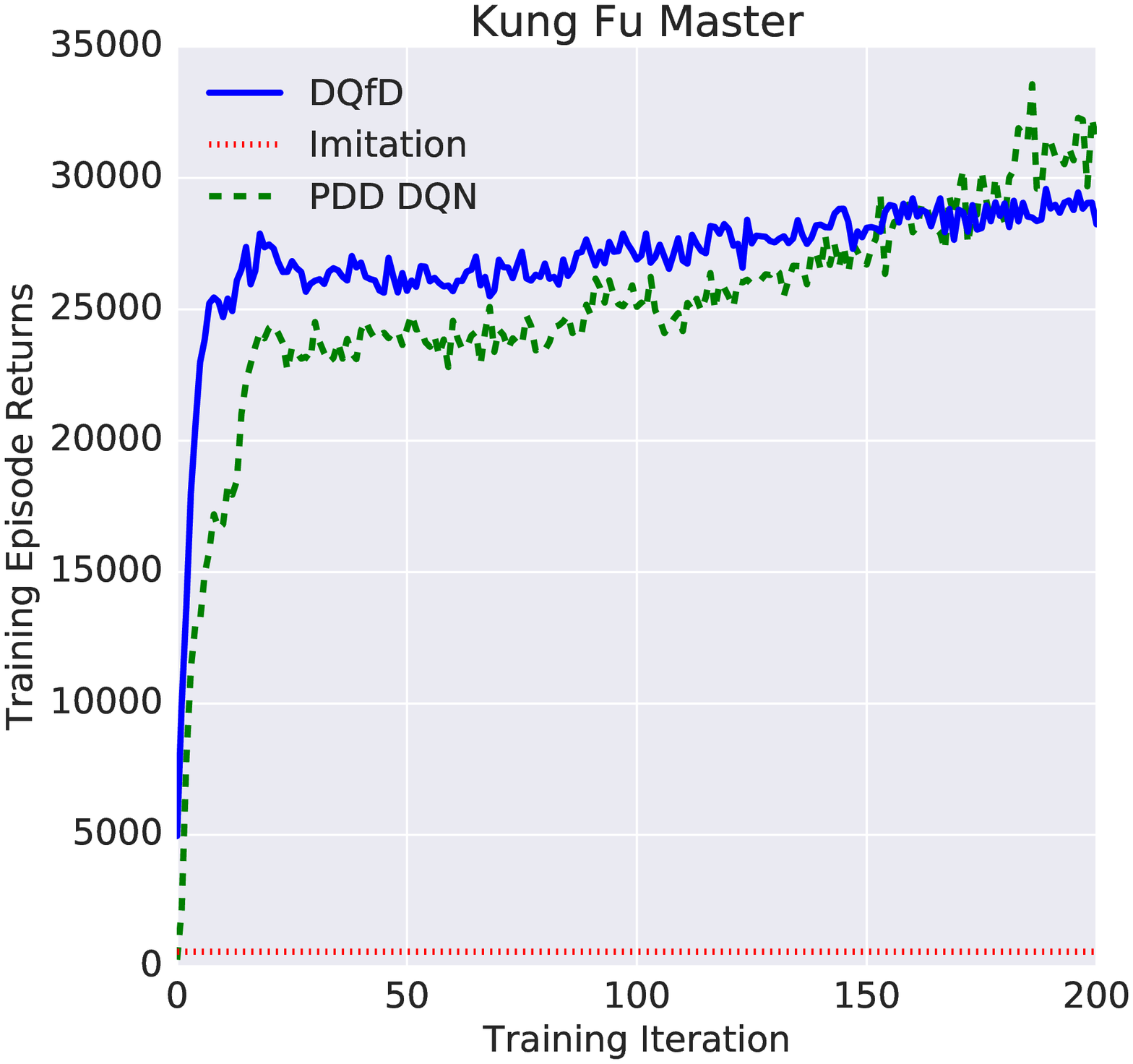}}
  \hspace{-0.02\linewidth}\subfigure{\includegraphics[width=0.177\linewidth]{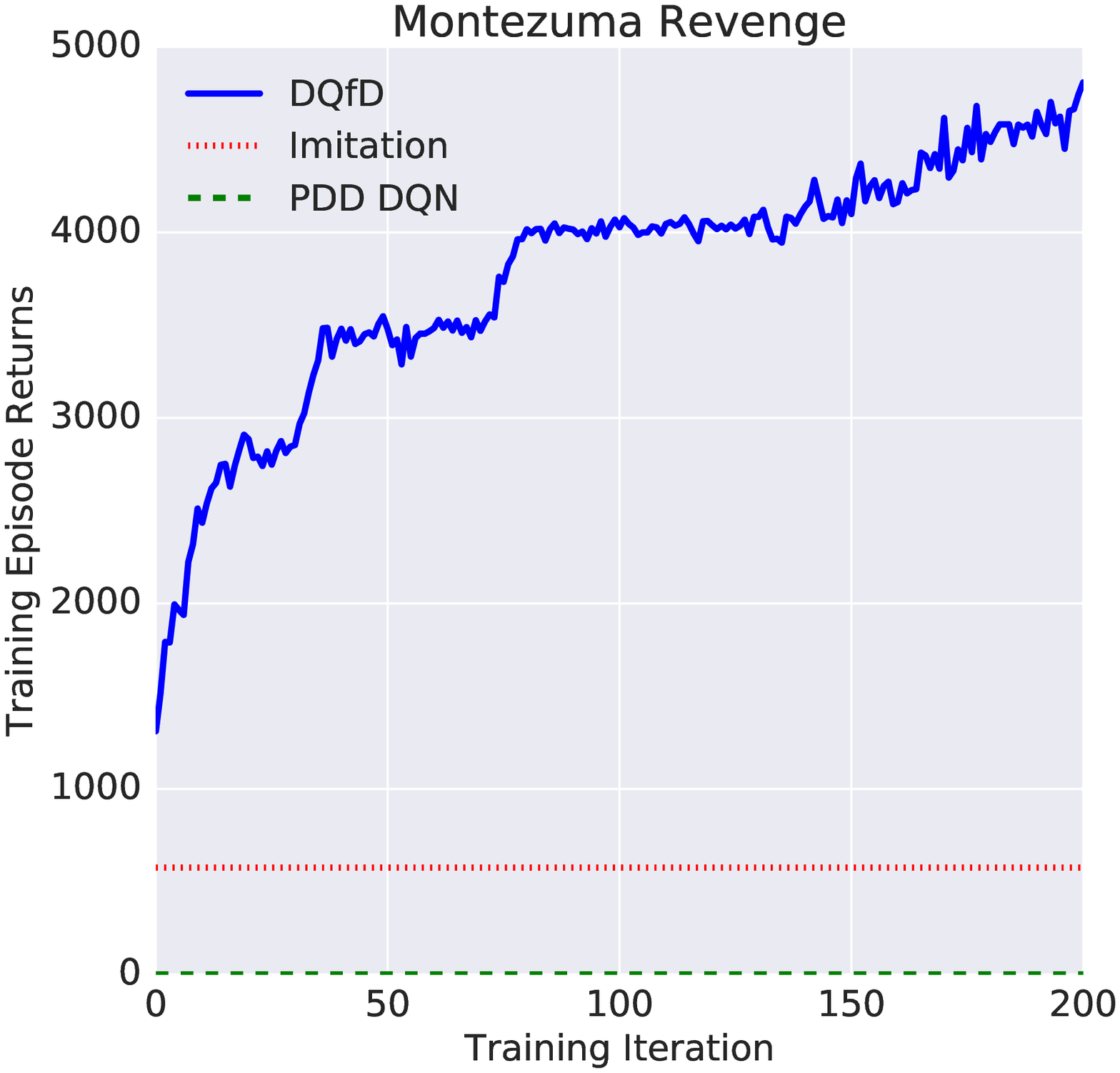}}
  \hspace{-0.02\linewidth}\subfigure{\includegraphics[width=0.177\linewidth]{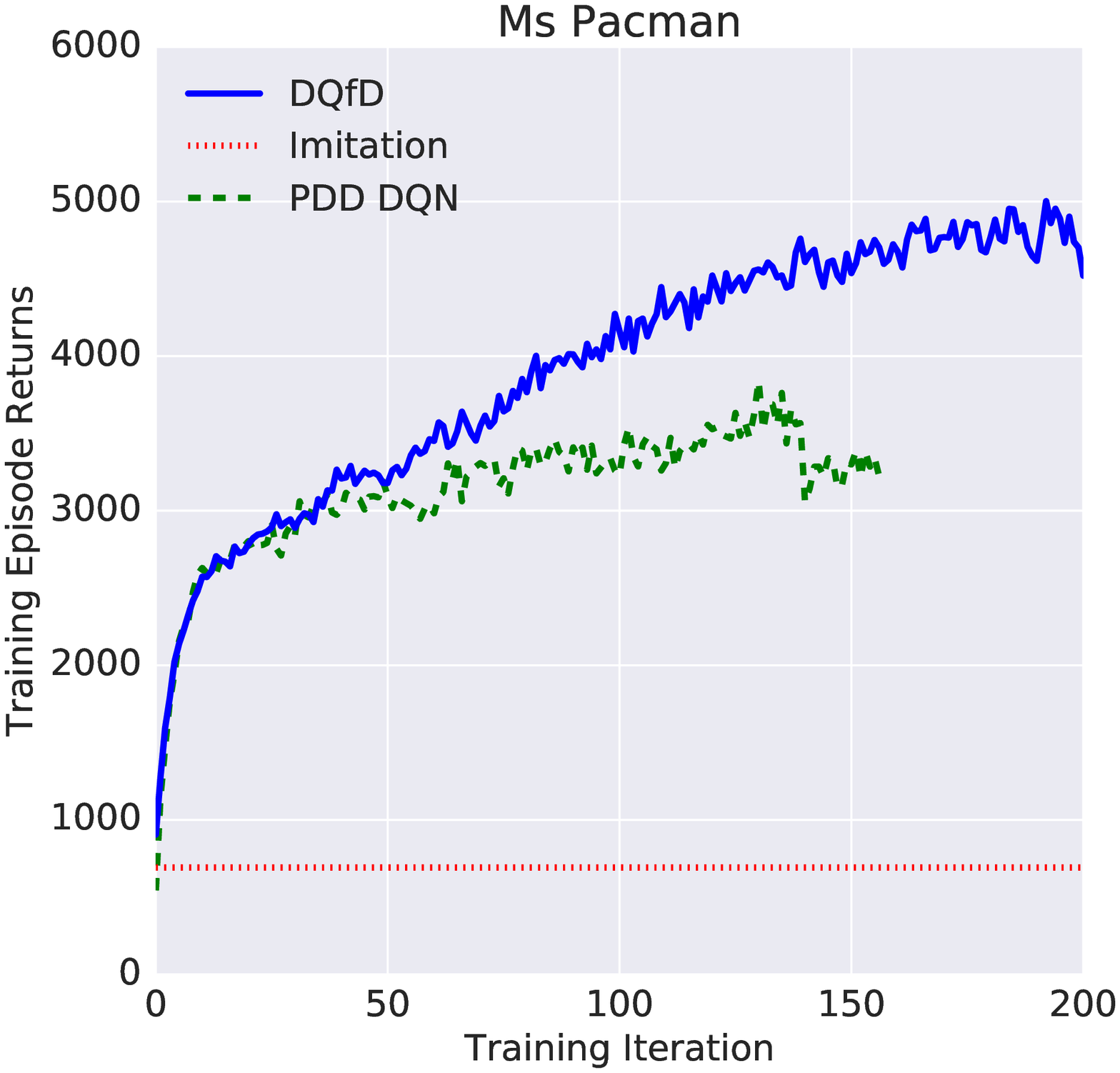}}
  \vspace{-0.25cm}
  \hspace{-0.02\linewidth}\subfigure{\includegraphics[width=0.177\linewidth]{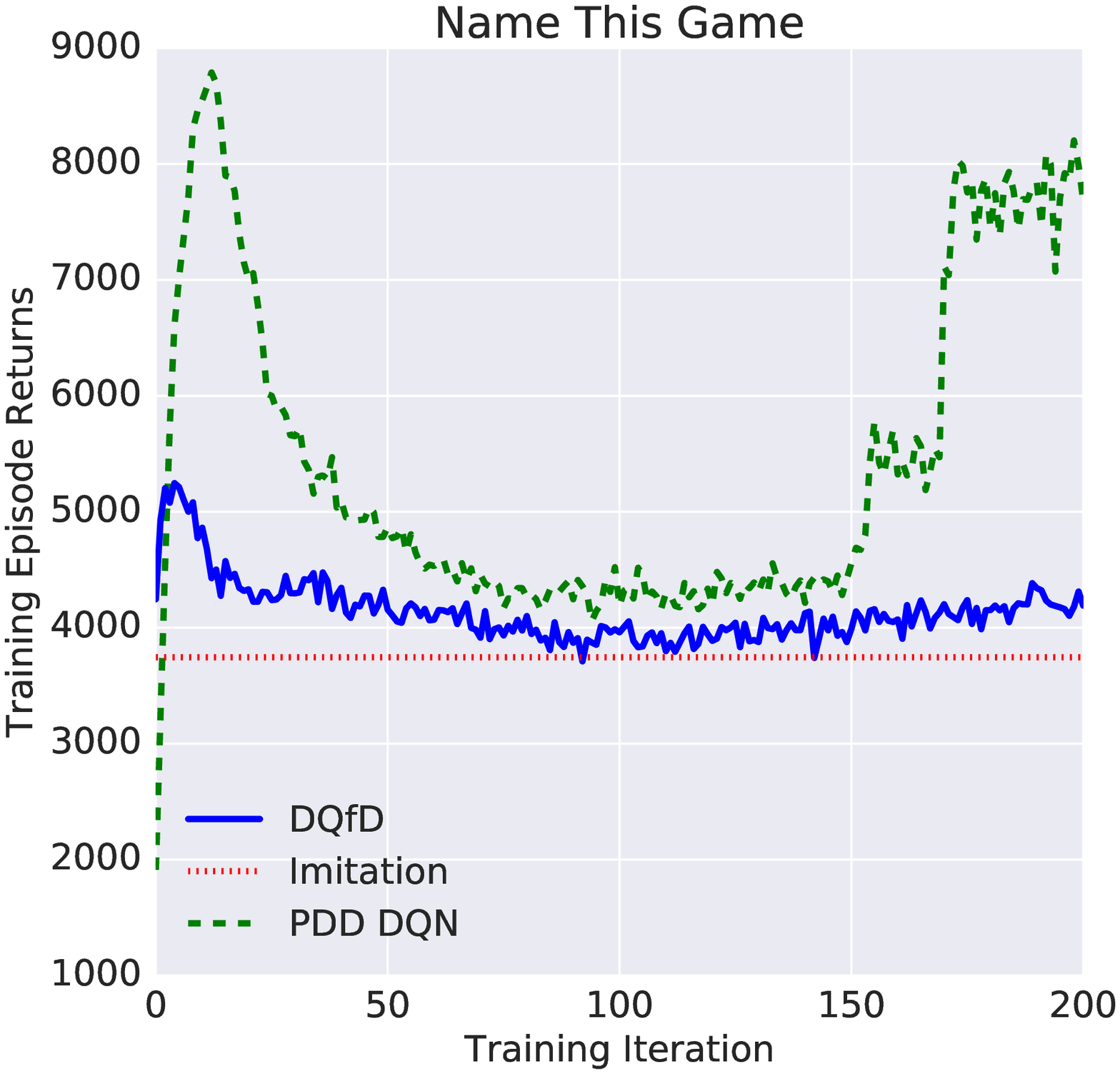}}
  \hspace{-0.02\linewidth}\subfigure{\includegraphics[width=0.177\linewidth]{figures/Pitfall_training_mean_rewards_new2.eps}}
  \hspace{-0.02\linewidth}\subfigure{\includegraphics[width=0.177\linewidth]{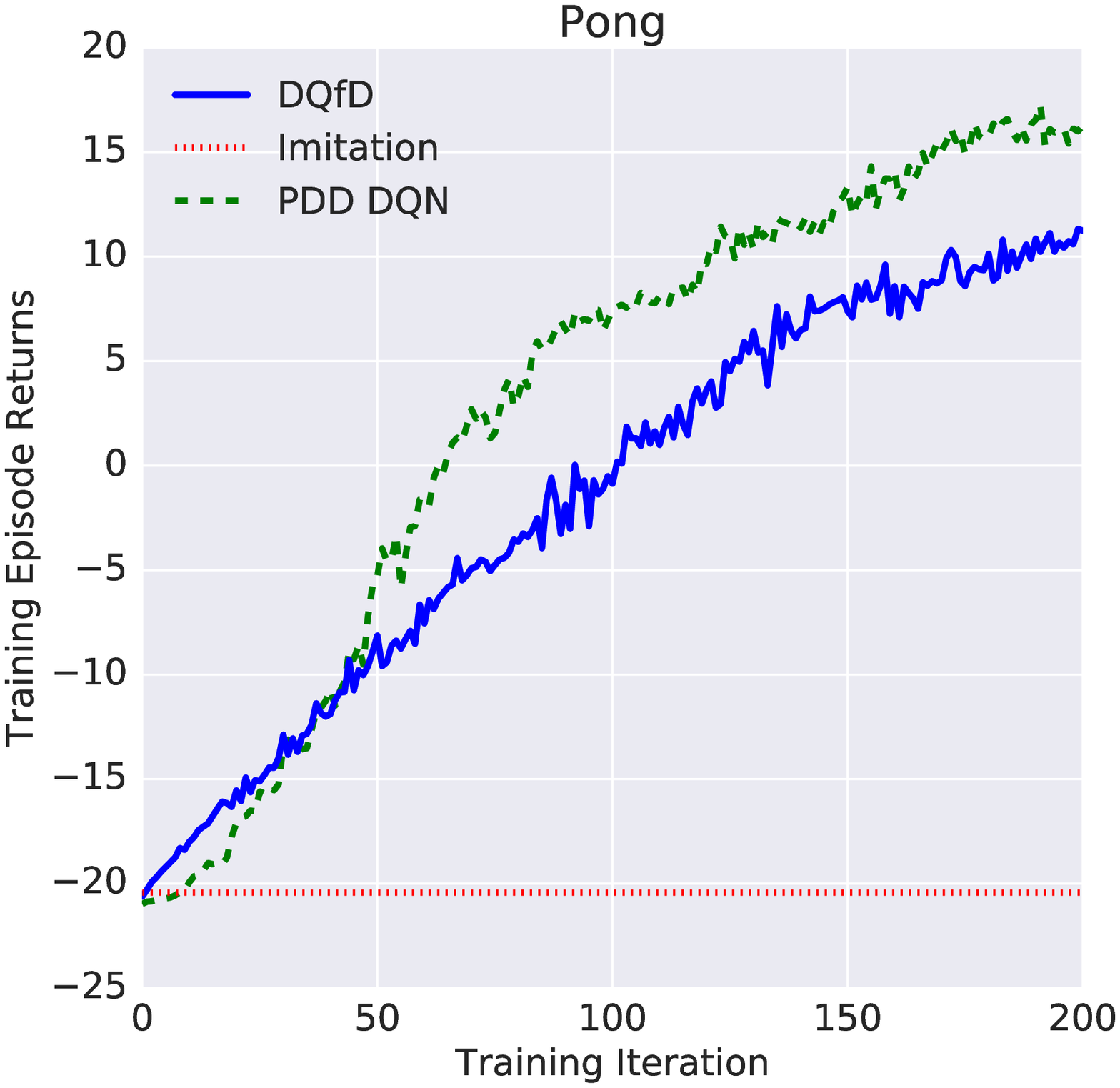}}
  \hspace{-0.02\linewidth}\subfigure{\includegraphics[width=0.177\linewidth]{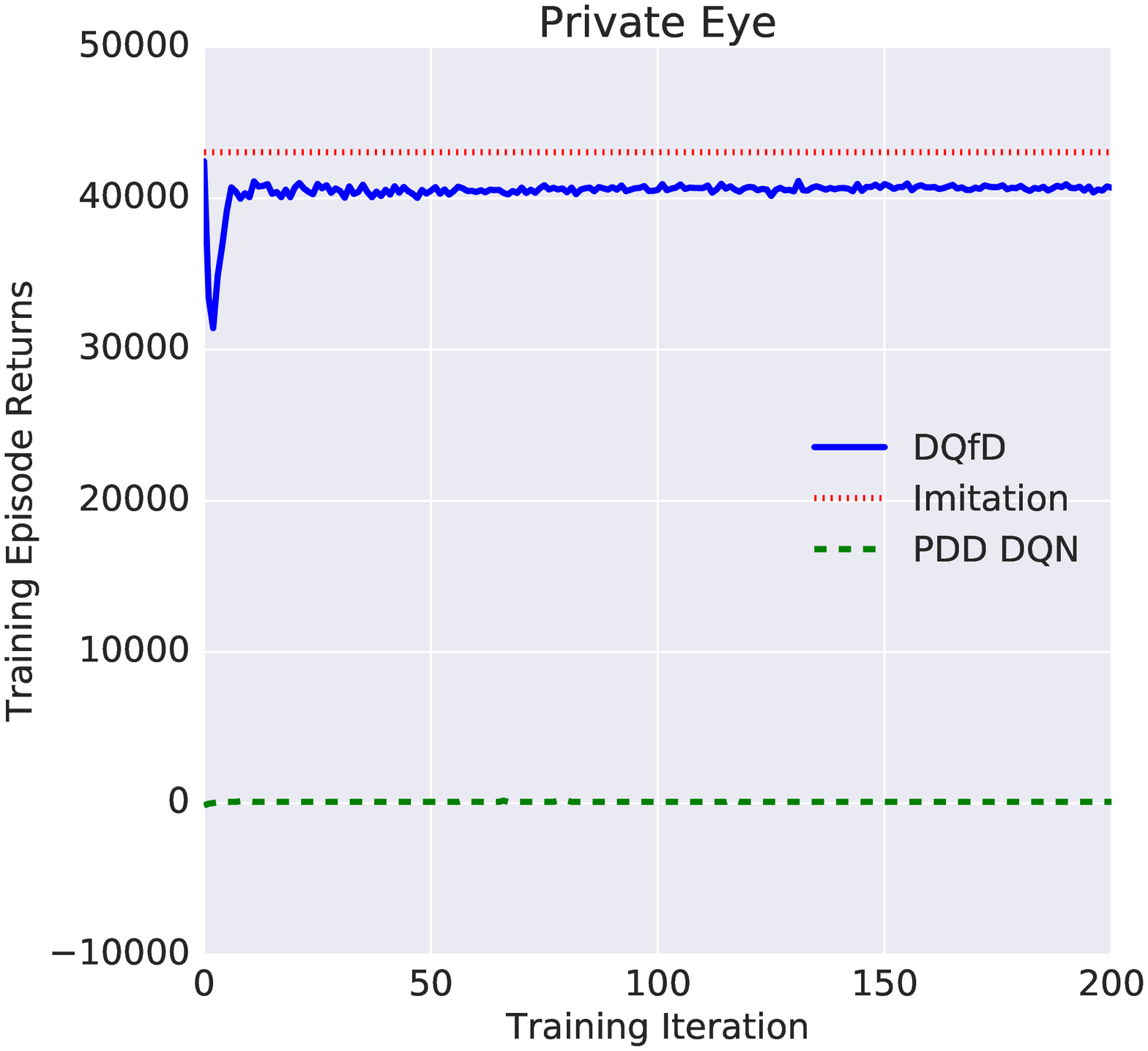}}
  \hspace{-0.02\linewidth}\subfigure{\includegraphics[width=0.177\linewidth]{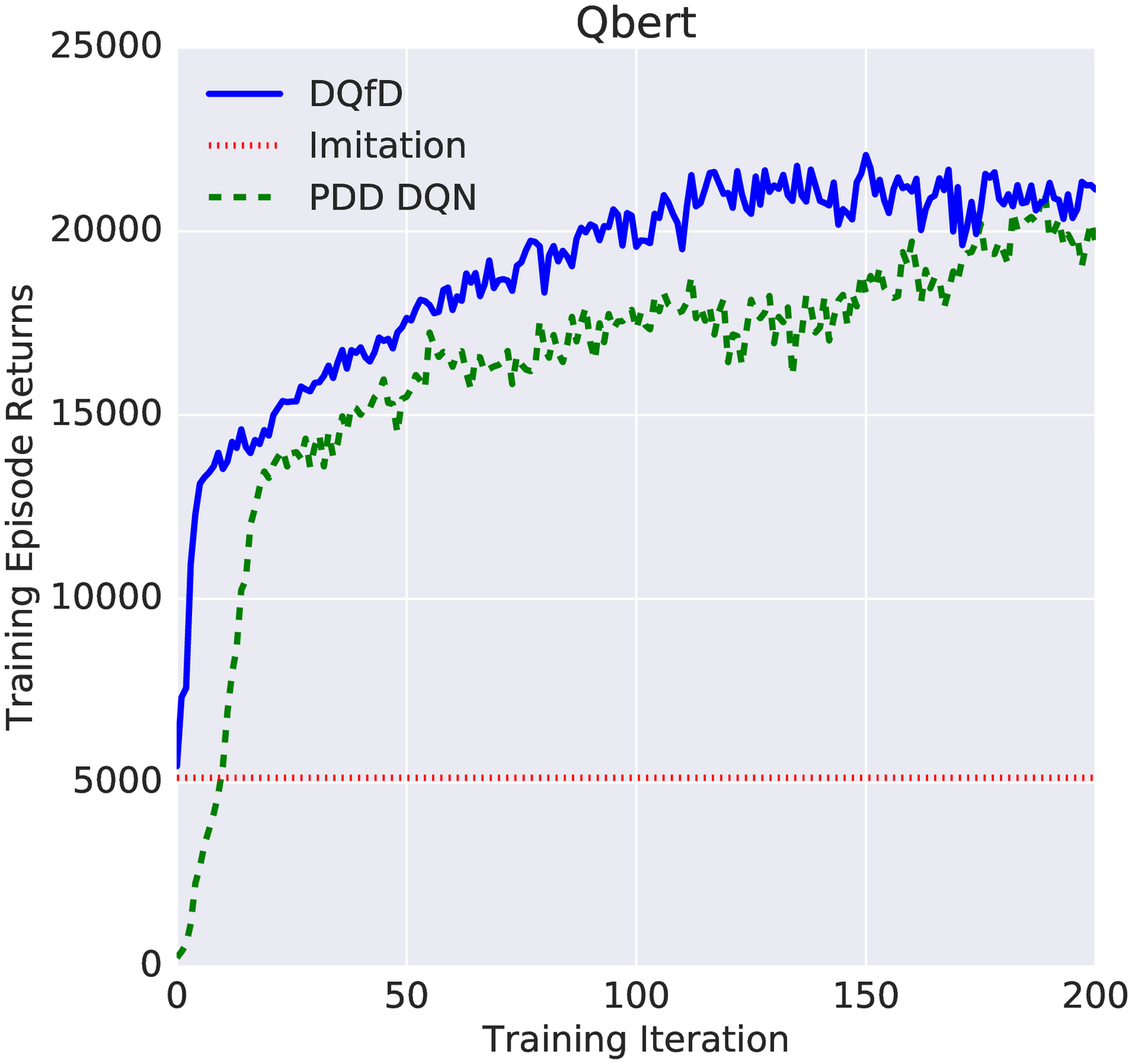}}
  \hspace{-0.02\linewidth}\subfigure{\includegraphics[width=0.177\linewidth]{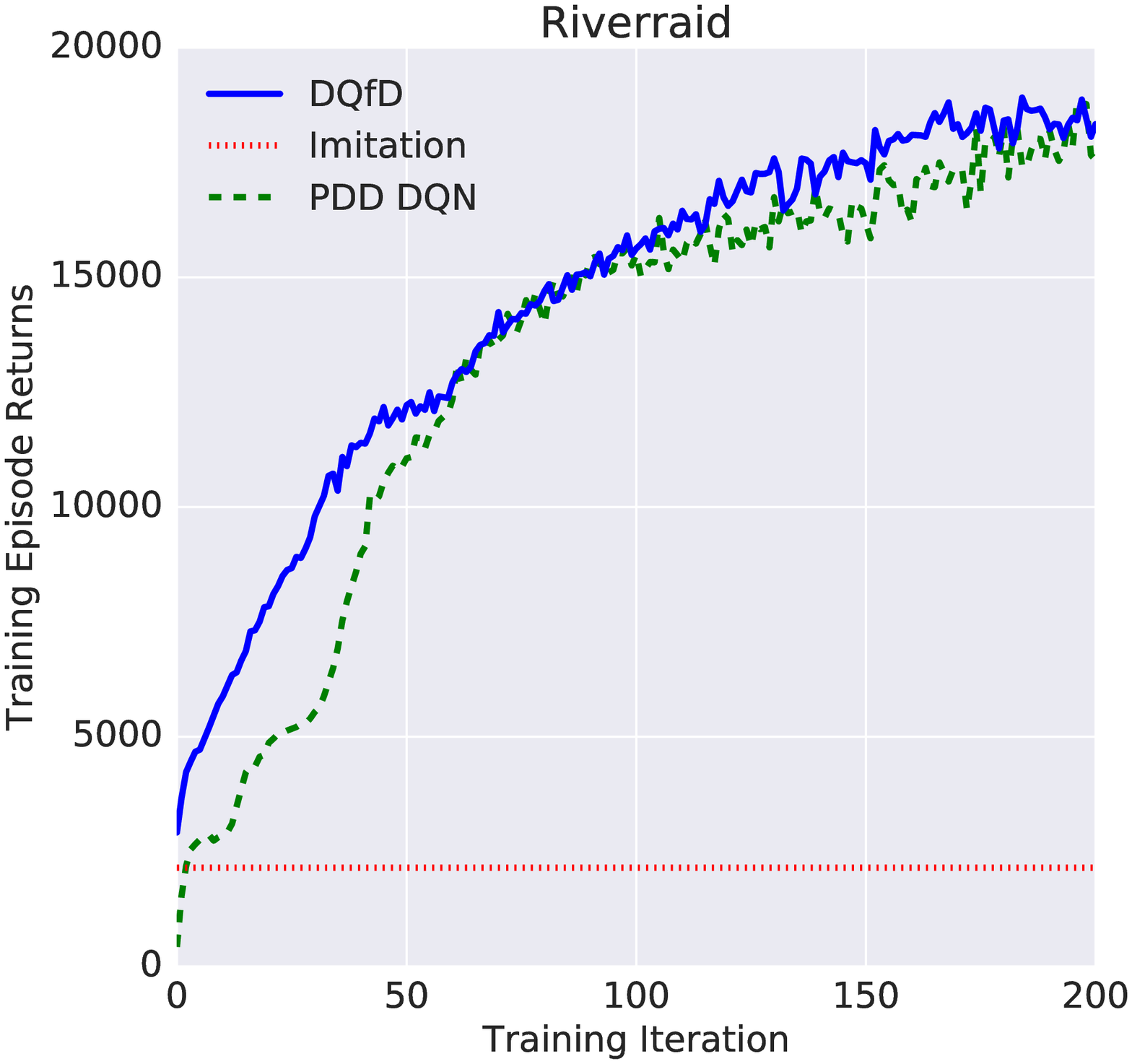}}
  \vspace{-0.25cm}
  \hspace{-0.02\linewidth}\subfigure{\includegraphics[width=0.177\linewidth]{figures/Road_Runner_training_mean_rewards_new2.eps}}
  \hspace{-0.02\linewidth}\subfigure{\includegraphics[width=0.177\linewidth]{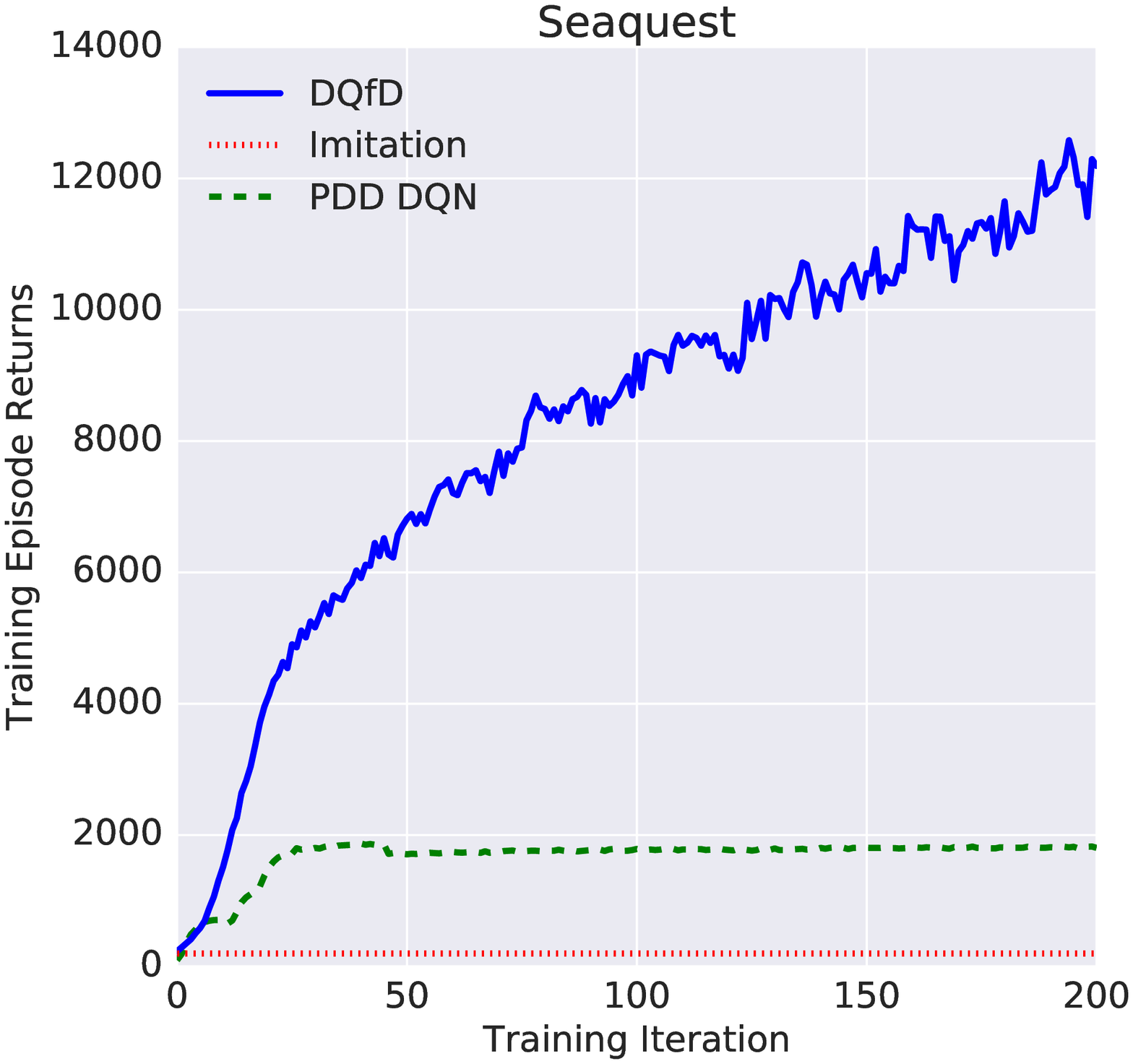}}
  \hspace{-0.02\linewidth}\subfigure{\includegraphics[width=0.177\linewidth]{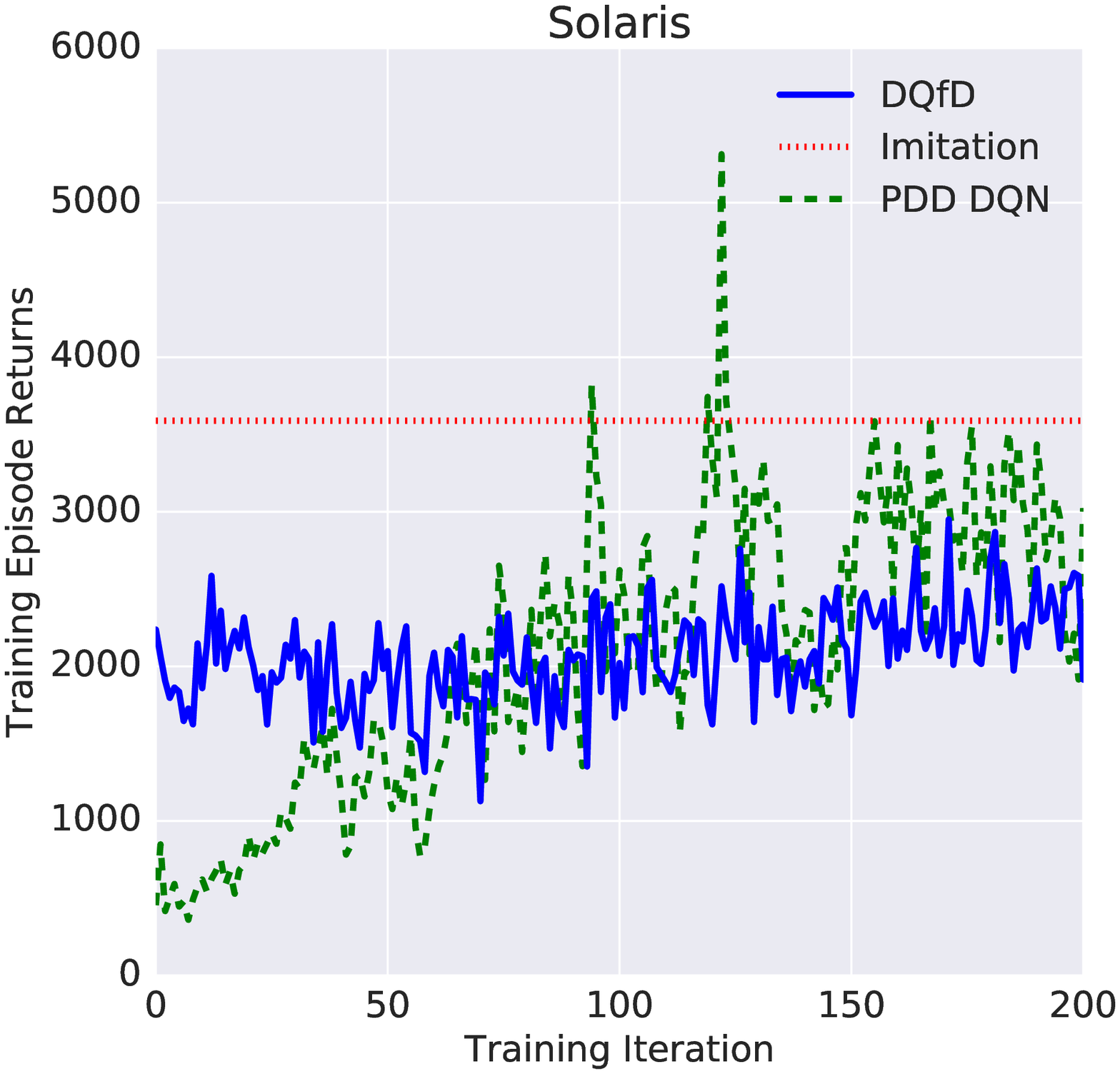}}
  \hspace{-0.02\linewidth}\subfigure{\includegraphics[width=0.177\linewidth]{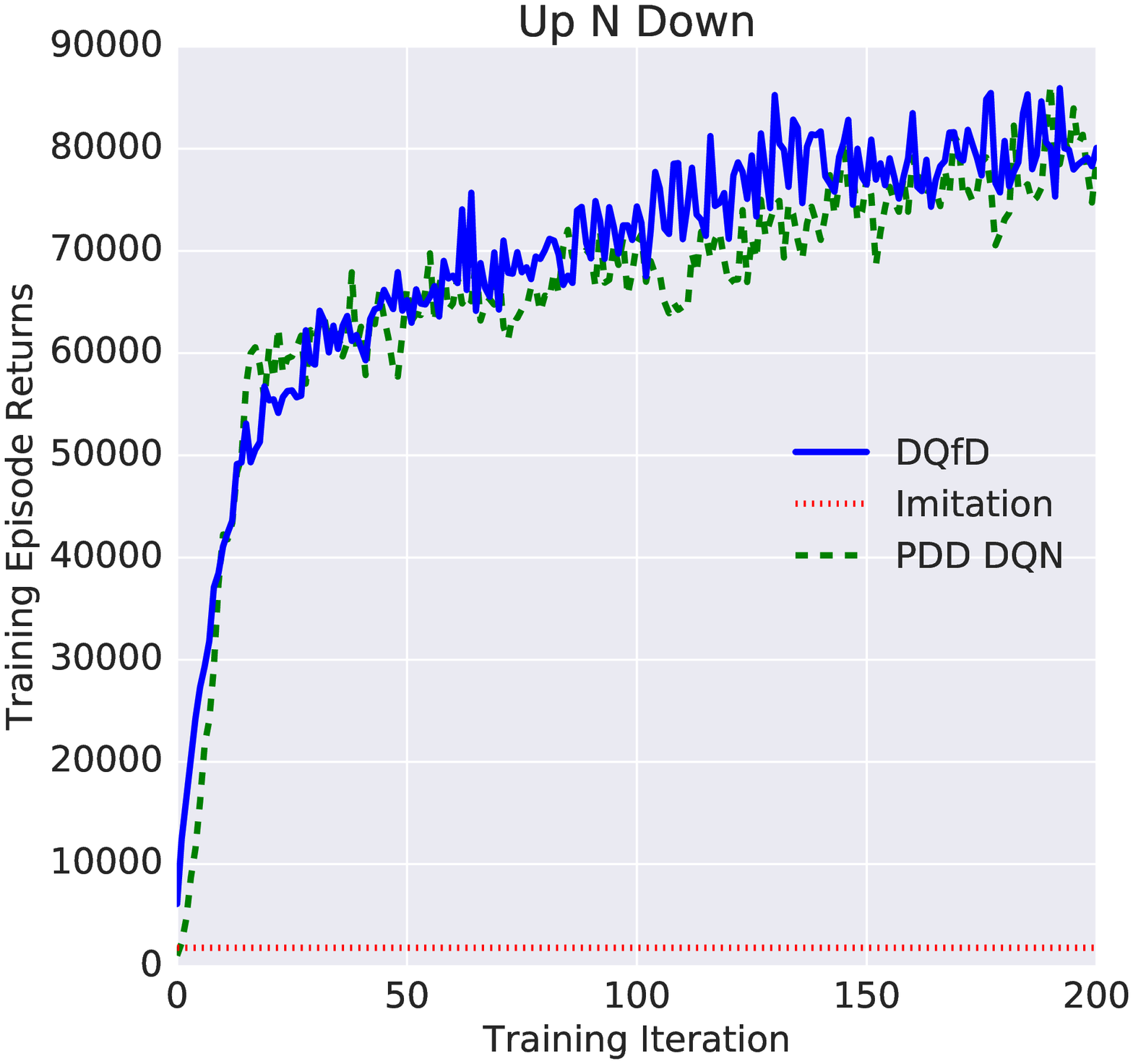}}
  \hspace{-0.02\linewidth}\subfigure{\includegraphics[width=0.177\linewidth]{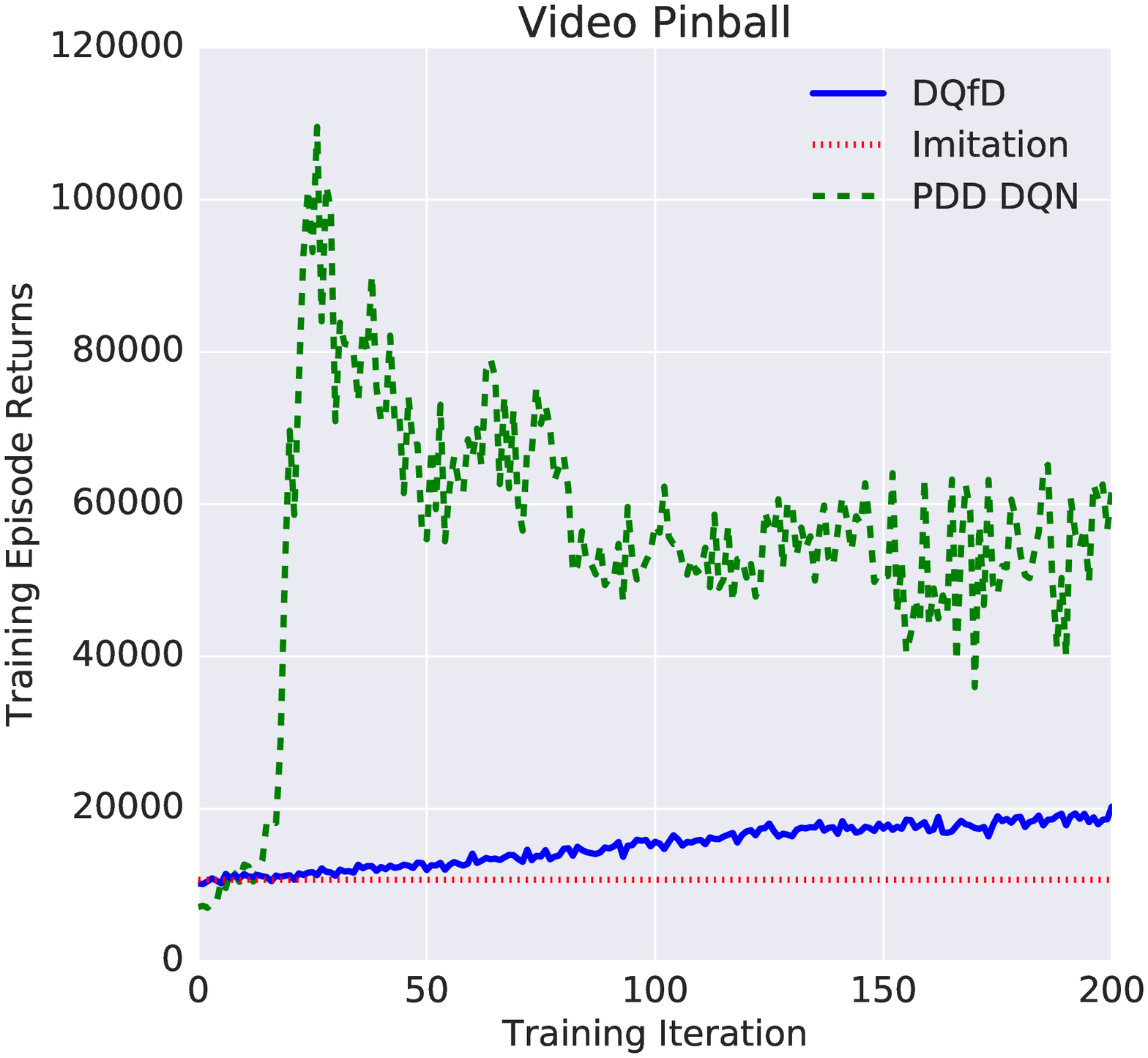}}
  \hspace{-0.02\linewidth}\subfigure{\includegraphics[width=0.177\linewidth]{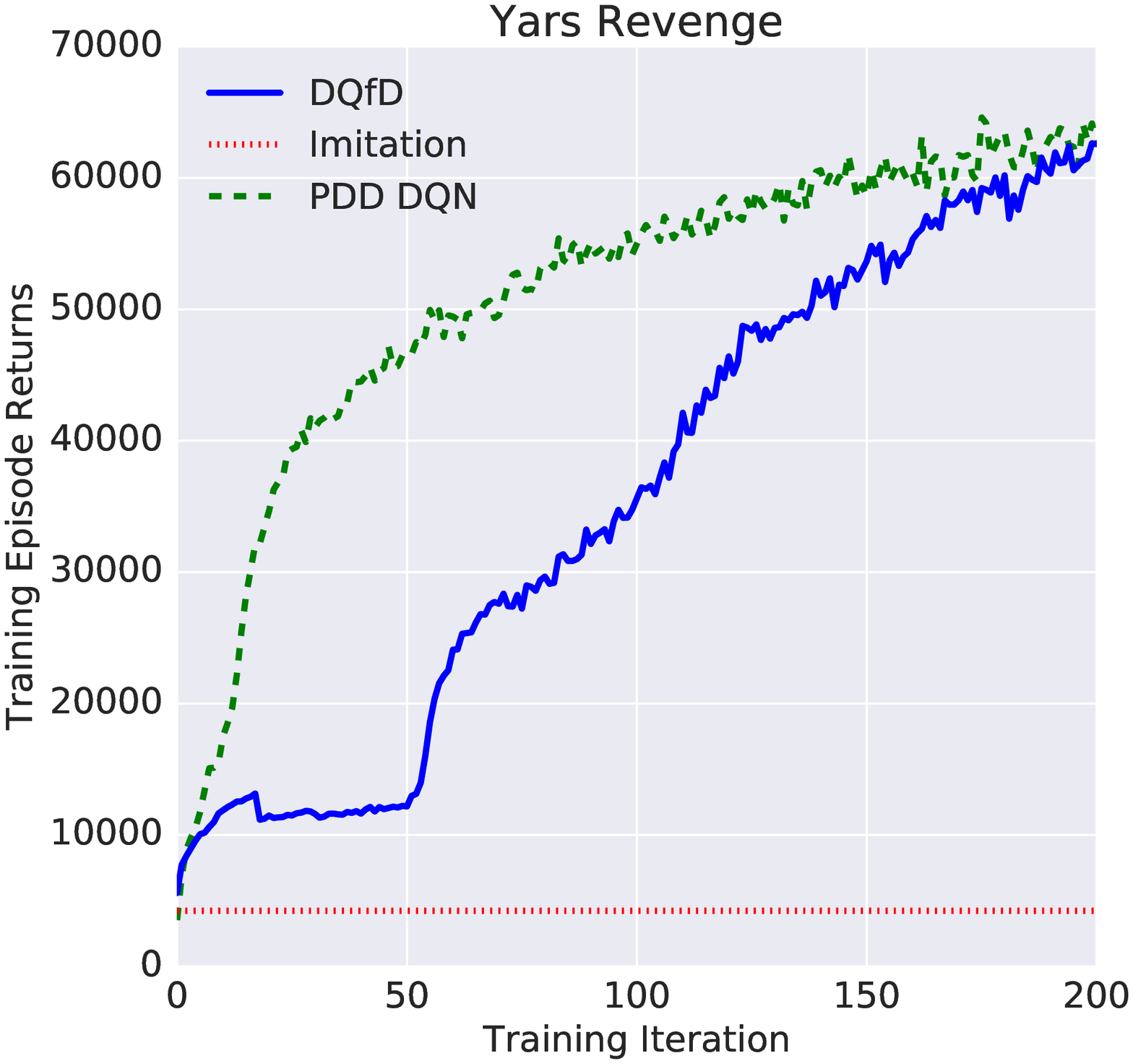}}
  \caption{On-line rewards of the three algorithms on the 42 Atari games, averaged over 4 trials. Each episode is started with up to 30 radom no-op actions. Scores are from the Atari game, regardless of the internal representation of reward used by the agent.}
  \label{fig:allgames}
\end{figure*}

\begin{table*}[p]
  \scriptsize \tabcolsep 3pt \centering
  \begin{tabular}{||l||l|l|l|l||l|l|l||}
    \hline \hline
    Game & Worst Demo & Best Demo & Number & Number & DQfD & PDD DQN & Imitation \\
    & Score & Score & Transitions & Episodes & Mean Score & Mean Score & Mean Score \\
    \hline
    Alien & 9690 & 29160 & 19133 & 5 & 4737.5 & \textbf{6197.1} & 473.9 \\ 
Amidar & 1353 & 2341 & 16790 & 5 & \textbf{2325.0} & 2140.4 & 175.0 \\ 
Assault & 1168 & 2274 & 13224 & 5 & 1755.7 & \textbf{1880.9} & 634.4 \\ 
Asterix & 4500 & 18100 & 9525 & 5 & 5493.6 & \textbf{7566.2} & 279.9 \\ 
Asteroids & 14170 & 18100 & 22801 & 5 & 3796.4 & \textbf{3917.6} & 1267.3 \\ 
Atlantis & 10300 & 22400 & 17516 & 12 & \textbf{920213.9} & 303374.8 & 12736.6 \\ 
Bank Heist & 900 & 7465 & 32389 & 7 & \textbf{1280.2} & 1240.8 & 95.2 \\ 
Battle Zone & 35000 & 60000 & 9075 & 5 & 41708.2 & \textbf{41993.7} & 14402.4 \\ 
Beam Rider & 12594 & 19844 & 38665 & 4 & 5173.3 & \textbf{5401.4} & 365.9 \\ 
Bowling & 89 & 149 & 9991 & 5 & \textbf{97.0} & 71.0 & 92.6 \\ 
Boxing & 0 & 15 & 8438 & 5 & 99.1 & \textbf{99.3} & 7.8 \\ 
Breakout & 17 & 79 & 10475 & 9 & \textbf{308.1} & 275.0 & 3.5 \\ 
Chopper Command & 4700 & 11300 & 7710 & 5 & \textbf{6993.1} & 6973.8 & 2485.7 \\ 
Crazy Climber & 30600 & 61600 & 18937 & 5 & \textbf{151909.5} & 136828.2 & 14051.0 \\ 
Defender & 5150 & 18700 & 6421 & 5 & \textbf{27951.5} & 24558.8 & 3819.1 \\ 
Demon Attack & 1800 & 6190 & 17409 & 5 & \textbf{3848.8} & 3511.6 & 147.5 \\ 
Double Dunk & -22 & -14 & 11855 & 5 & -20.4 & \textbf{-14.3} & -21.4 \\ 
Enduro & 383 & 803 & 42058 & 5 & 1929.8 & \textbf{2199.6} & 134.8 \\ 
Fishing Derby & -10 & 20 & 6388 & 4 & \textbf{38.4} & 28.6 & -74.4 \\ 
Freeway & 30 & 32 & 10239 & 5 & 31.4 & \textbf{31.9} & 22.7 \\ 
Gopher & 2500 & 22520 & 38632 & 5 & 7810.3 & \textbf{12003.4} & 1142.6 \\ 
Gravitar & 2950 & 13400 & 15377 & 5 & 1685.1 & \textbf{2796.1} & 248.0 \\ 
Hero & 35155 & 99320 & 32907 & 5 & \textbf{105929.4} & 22290.1 & 5903.3 \\ 
Ice Hockey & -4 & 1 & 17585 & 5 & -9.6 & \textbf{-2.0} & -13.5 \\ 
James Bond & 400 & 650 & 9050 & 5 & \textbf{2095.0} & 639.6 & 262.1 \\ 
Kangaroo & 12400 & 36300 & 20984 & 5 & \textbf{14681.5} & 13567.3 & 917.3 \\ 
Krull & 8040 & 13730 & 32581 & 5 & 9825.3 & \textbf{10344.4} & 2216.6 \\ 
Kung Fu Master & 8300 & 25920 & 12989 & 5 & 29132.0 & \textbf{32212.3} & 556.7 \\ 
Montezuma's Revenge & 32300 & 34900 & 17949 & 5 & \textbf{4638.4} & 0.1 & 576.3 \\ 
Ms Pacman & 31781 & 55021 & 21896 & 3 & \textbf{4695.7} & 3684.2 & 692.4 \\ 
Name This Game & 11350 & 19380 & 43571 & 5 & 5188.3 & \textbf{8716.8} & 3745.3 \\ 
Pitfall & 3662 & 47821 & 35347 & 5 & 57.3 & 0.0 & \textbf{182.8} \\ 
Pong & -12 & 0 & 17719 & 3 & 10.7 & \textbf{16.7} & -20.4 \\ 
Private Eye & 70375 & 74456 & 10899 & 5 & 42457.2 & 154.3 & \textbf{43047.8} \\ 
Q-Bert & 80700 & 99450 & 75472 & 5 & \textbf{21792.7} & 20693.7 & 5133.8 \\ 
River Raid & 17240 & 39710 & 46233 & 5 & 18735.4 & \textbf{18810.0} & 2148.5 \\ 
Road Runner & 8400 & 20200 & 5574 & 5 & 50199.6 & \textbf{51688.4} & 8794.9 \\ 
Seaquest & 56510 & 101120 & 57453 & 7 & \textbf{12361.6} & 1862.5 & 195.6 \\ 
Solaris & 2840 & 17840 & 28552 & 6 & 2616.8 & \textbf{4157.0} & 3589.6 \\ 
Up N Down & 6580 & 16080 & 10421 & 4 & \textbf{82555.0} & 82138.5 & 1816.7 \\ 
Video Pinball & 8409 & 32420 & 10051 & 5 & 19123.1 & \textbf{101339.6} & 10655.5 \\ 
Yars' Revenge & 48361 & 83523 & 21334 & 4 & 61575.7 & \textbf{63484.4} & 4225.8 \\ 
     \hline \hline
\end{tabular}
\caption{Best and worst human demonstration scores, the number of
  trials and transitions collected, and average score for each algorithm on the best 3 million step window, averaged over four seeds. }
  \label{tab:games}
\vspace{-0.3cm}
\end{table*}

\end{document}